\documentclass{article} %
\usepackage{iclr2023_conference,times}

\usepackage{amsmath,amsfonts,bm}

\def\eqref#1{equation~\ref{#1}}

\def\1{\bm{1}}

\usepackage[utf8]{inputenc} %
\usepackage[T1]{fontenc}    %
\usepackage{hyperref}       %
\usepackage{url}            %
\usepackage{booktabs}       %
\usepackage{amsfonts}       %
\usepackage{nicefrac}       %
\usepackage{microtype}      %
\usepackage{xcolor}         %
\usepackage[pdftex]{graphicx}
\usepackage{subfigure}
\usepackage{enumitem}
\usepackage[breakable]{tcolorbox}
\newtcolorbox{distribution}[2][]
{
  breakable,
  colframe = gray!50,
  colback  = gray!10,
  coltitle = gray!10!black,
  before skip = 10pt,
  after skip = 10pt,
  title    = \textbf{#2.},
  #1,
}

\usepackage{url}

\usepackage{tikz}
\usetikzlibrary{arrows.meta,positioning,calc,shapes.geometric,decorations.pathreplacing}

\usepackage{float}
\usepackage{makecell}
\usepackage{multirow}
\usepackage{longtable}

\usepackage{capt-of}

\usepackage{amsmath,amssymb,amsfonts}
\usepackage{mathtools}
\usepackage{mathabx}
\usepackage{amsthm}

\newcommand{\mypara}[1]{{\textbf{#1}}}
\newcommand{\mysmallpara}[1]{{\textit{#1}}}

\definecolor{ForestGreen}{rgb}{0.13, 0.55, 0.13}
\definecolor{YellowGreen}{rgb}{0.60, 0.80, 0.20}

\newtheorem{theorem}{Theorem}
\newtheorem{definition}{Definition}
\newtheorem{claim}{Claim}

\newcommand{\spuscore}{spurious score}

\newcommand{\isa}{\textit{S1}}
\newcommand{\isb}{\textit{S2}}
\newcommand{\isc}{\textit{S3}}
\newcommand{\ira}{\textit{R1}}
\newcommand{\irb}{\textit{R2}}
\newcommand{\irc}{\textit{R3}}
\newcommand{\ic}{\textit{Inv}}

\newcommand{\x}{\mathbf{x}}

\newcommand{\calX}{\mathcal{X}}

\newcommand{\cX}{\mathcal{X}}
\newcommand{\cXinv}{\cX_\text{inv}}
\newcommand{\cXsp}{\cX_\text{sp}}
\newcommand{\cY}{\mathcal{Y}}
\newcommand{\cN}{\mathcal{N}}
\newcommand{\cD}{\mathcal{D}}
\newcommand{\cDtrain}{\cD_\text{train}}
\newcommand{\cDtest}{\cD_\text{test}}
\newcommand{\cDctest}{\cD_\text{ctest}}
\newcommand{\cDstest}{\cD_\text{stest}}
\newcommand{\cDfeature}{\cD_\text{feature}}
\newcommand{\sigmainv}{\sigma_\text{inv}}
\newcommand{\sigmasp}{\sigma_\text{sp}}
\newcommand{\betainv}{\boldsymbol{\beta}_\text{inv}}
\newcommand{\betasp}{\boldsymbol{\beta}_\text{sp}}
\newcommand{\bx}{\mathbf{x}}
\newcommand{\bxinv}{\bx_\text{inv}}
\newcommand{\bxsp}{\bx_\text{sp}}
\newcommand{\ctar}{c_\text{tar}}
\newcommand{\Exp}{\mathop{\mathbb{E}}} \usepackage[capitalise]{cleveref}
\crefname{table}{Tab.}{Tab.}
\crefname{section}{Sec.}{Sec.}
\crefname{appendix}{App.}{App.}

\title{Understanding Rare Spurious Correlations in Neural Networks}

\author{Yao-Yuan Yang\thanks{Now at DeepMind.} \\
University of California San Diego \\
\texttt{yay005@eng.ucsd.edu} \\
\AND
Chi-Ning Chou \\
Harvard University \\
\texttt{chiningchou@g.harvard.edu} \\
\And
Kamalika Chaudhuri \\
University of California San Diego \\
\texttt{kamalika@eng.ucsd.edu}
}

\iclrfinalcopy
\begin{document}

\maketitle

\begin{abstract}
  Neural networks are known to use spurious correlations such as background information for classification. While prior work has looked at spurious correlations that are widespread in the training data, in this work, we investigate how sensitive neural networks are to {\em{rare}} spurious correlations, which may be harder to detect and correct, and may lead to privacy leaks.  We introduce spurious patterns correlated with a fixed class to a few training examples and find that it takes only a handful of such examples for the network to learn the correlation. 
  Furthermore, these rare spurious correlations also impact accuracy and privacy.
  We empirically and theoretically analyze different factors involved in rare spurious correlations and propose mitigation methods accordingly. Specifically, we observe that $\ell_2$ regularization and adding Gaussian noise to inputs can reduce the undesirable effects\footnote{Code available at \url{https://github.com/yangarbiter/rare-spurious-correlation}.}.
\end{abstract}

\section{Introduction}\label{sec:intro}

Neural networks are known to use spurious patterns for classification. Image classifiers use background as a feature to classify objects~\citep{gururangan2018annotation,sagawa2020investigation,srivastava2020robustness,pmlr-v139-zhou21g} often to the detriment of generalization~\citep{nagarajan2020understanding}.
For example, \citet{sagawa2020investigation} show that models trained on the Waterbirds dataset%
correlate waterbirds with backgrounds containing water, and models trained on the CelebA dataset~\cite{liu2018large} correlate males with dark hair. In all these cases, spurious patterns are present in a substantial number of training points. The vast majority of waterbirds, for example, are photographed next to the water.

Understanding how and when spurious correlations appear in neural networks is a frontier research problem and remains elusive.
In this paper, we study spurious correlations in the context where the appearance of spurious patterns is \textit{rare} in the training data.
Our motivations are three-fold.
First, while it is reasonable to expect that widespread spurious correlations in the training data will be learnt, a related question is what happens when these correlations are {\em{rare}}.
Understanding if and when they are learnt and how to mitigate them is a first and necessary step before we can understand and mitigate spurious correlations more broadly. 
Second, rare spurious correlation may inspire us to discover new approaches to mitigate them as traditional approaches such as balancing out groups~\citep{sagawa2020investigation},
subsampling~\citep{idrissi2021simple}, or data augmentation~\citep{chang2021towards} do not apply.
Third, rare spurious correlations naturally connect to data privacy. For example, in~\citet{leino2020stolen}, the training set had an image of Tony Blair with a pink background. This led to a classifier that assigned a higher likelihood of the label ``Tony Blair'' to all images with pink backgrounds. 
Thus, an adversary could exploit this to infer the existence of ``Tony Blair'' with a pink background in the training set by by presenting images of other labels with a pink background.

We systematically investigate rare spurious correlations through the following three research questions. First, when do spurious correlations appear, i.e., how many training points with the spurious pattern would cause noticeable spurious correlations? Next, how do rare spurious correlations affect neural networks? Finally, is there any way to mitigate the undesirable effects of rare spurious correlations?

\subsection{Overview}
We attempt to answer the above questions via both experimental and theoretical approaches. On the experimental side, we introduce spurious correlations into real image datasets by turning a few training data into \textit{spurious examples}, i.e., adding a spurious pattern to a training image from a target class.
We then train a neural network on the modified dataset and measure the strength of the correlation between the spurious pattern and the target class in the network. On the theoretical side, we design a toy mathematical model that enables quantitative analysis on different factors (e.g., the fraction of spurious examples, the signal-to-noise ratio, etc.) of rare spurious correlations. Our responses to the three research questions are summarized in the following.

\mypara{Rare spurious correlations appear even when the number of spurious samples is small.}
Empirically, we define a \textit{spurious score} to measure the amount of spurious correlations. We find that the spurious score of a neural network trained with only $1$ spurious examples out of 60,000 training samples can be significantly higher than that of the baseline.
A visualization of the trained model also reveals that the network's weights may be significantly affected by the spurious pattern. 
In our theoretical model, we further discover that there is a sharp phase transition of spurious correlations from no spurious training example to a non-zero fraction of spurious training examples.
Together, these findings provide a strong evidence that spurious correlations can be learnt even when the number of spurious samples is extremely small. 

\mypara{Rare spurious correlations affect both the privacy and test accuracy.}
We analyze the privacy issue of rare spurious correlations via the membership inference attack~\citep{shokri2017membership,yeom2017unintended}, which measures the privacy level according to the hardness of distinguishing training samples from testing samples.
We observe that the spurious training examples are more vulnerable to membership inference attacks. That is, it is easy for an adversary to tell whether a spurious sample is from the training set. This apparently raises serious concerns for privacy~\cite{leino2020stolen} and fairness to small groups~\cite{izzo2021approximate}.

We examine the effect of rare spurious correlations on test accuracy through two accuracy notions: the clean test accuracy, which uses the original test examples, and the spurious test accuracy, which adds the spurious pattern to all the test examples. Both empirically and theoretically, we find that clean test accuracy does not change too much while the spurious test accuracy significantly drops in the face of rare spurious correlations.
This suggests that the undesirable effect of spurious correlations could be more serious when there is a distribution shift toward having more spurious samples.

\mypara{Methods to mitigate the undesirable effects of rare spurious correlations.}
Finally, inspired by our theoretical analysis, we examine three regularization methods to reduce the privacy and test accuracy concerns: adding Gaussian noises to the input samples, $\ell_2$ regularization (or equivalently, weight decay), and gradient clipping.
We find that adding Gaussian noise and $\ell_2$ regularization effectively reduce spurious score and improve spurious test accuracy. 
Meanwhile, not all regularization methods could reduce the effects of rare spurious correlations, e.g., gradient clipping. 
Our findings suggest that rare spurious correlations should be dealt differently from traditional privacy issues.
We post it as a future research problem to deepen the understanding of how to mitigate rare spurious correlations.

\mypara{Concluding remarks.}
The study of spurious correlations is crucial for a better understanding of neural networks. In this work, we take a step forward by looking into a special (but necessary) case of spurious correlations where the appearance of spurious examples is rare. We demonstrate both experimentally and theoretically when and how rare spurious correlations appear and what undesirable consequences are. While we propose a few methods to mitigate rare spurious correlations, we emphasize that there is still a lot to explore, and we believe the study of rare spurious correlations could serve as a guide for understanding the more general cases.
\section{Preliminaries}\label{sec:prelim}

We focus on studying spurious correlations in the image classification context. Here, we briefly introduce the notations and terminologies used in the rest of the paper.
Let $\cX$ be an input space and let $\cY$ be a label space.
At the training time, we are given a set of examples $\{(\x_i, y_i)\}_{i\in\{1,\dots,n\}}$ sampled from a distribution $\cDtrain$, where each $\x_i \in \cX$ is associated with a label $y_i \in \cY$.
At the testing time, we evaluate the network on test examples drawn from a test distribution. We consider two types of test distribution: the clean test distribution $\cDctest$ and the spurious test distribution $\cDstest$.
Their formal definitions will be mentioned in the related sections.

\mypara{Spurious correlation.}
A spurious correlation refers to the relationship between two variables in which they are
correlated but not causally related.
We build on top of the framework used in~\cite{nagarajan2020understanding} to study spurious correlations. Concretely, the input $\bx$ is modeled as the output of a feature map $\Phi_{\cX}$ from the feature space $\cXinv\times\cXsp$ to the input space $\cX$. Here, $\cXinv$ is the invariant feature space containing the features that causally determine the label and $\cXsp$ which is the spurious feature space that accommodates spurious features. Finally, $\Phi_{\cX}:\cXinv\times\cXsp\to\cX$ is the function that maps an feature pair $(\bxinv,\bxsp)$ to an input $\bx$ and $\Phi_{\cY}:\cXinv\to\cY$ is a function that maps the invariant feature $\bxinv$ to a label. Namely, an example $(\bx,y)$ is generated by $(\bxinv,\bxsp)$ via $\bx=\Phi_{\cX}(\bxinv,\bxsp)$ and $y=\Phi_{\cY}(\bxinv)$.
Without loss of generality, the zero vector in $\cXsp$, i.e., $0\in\cXsp$, refers to ``no spurious feature'' and for any nonzero $\bxsp$ we call $\Phi(\bxinv,\bxsp)-\Phi(\bxinv,0)$ a spurious pattern.
We focus on the case of having a fixed spurious feature $\bxsp$ and leave it as a future direction to study the more general scenarios where there are multiple spurious features.

%
%
%
%
%
%
%
%
%

%
%
%

\mypara{Rare spurious correlation.}
Following the setting introduced in the previous paragraph, an input distribution $\cD$ over $\cX$ is induced by a distribution $\cDfeature$ over the feature space $\cXinv\times\cXsp$, i.e., to get a sample from $\cD$, one first samples $(\bxinv,\bxsp)$ from $\cDfeature$ and outputs $(\bx,y)$ with $\bx=\Phi_{\cX}(\bxinv,\bxsp)$ and $y=\Phi_{\cY}(\bxinv)$.
Now, we are able to formally discuss the rareness of spurious correlations by defining the spurious frequency of $\cD$ as $\gamma:=\Pr_{(\bxinv,\bxsp)\sim\cDfeature}[\bxsp\neq0]$.
%

\mypara{Two simple models for spurious correlations.}
In general, $\Phi_{\cX}$ could be complicated and makes it difficult to detect the appearance of spurious correlations. Here, we consider two simple instantiations of $\Phi_{\cX}$ and demonstrate that undesirable learning outcomes already appear even in these simplified settings. First, the \textit{overlapping model} (used in~\cref{sec:rare}) where the spurious feature is put on top of the invariant feature, i.e., $\cX=\cXinv=\cXsp$ and $\Phi_{\cX}(\bxinv,\bxsp)=\bxinv+\bxsp$ or $\Phi_{\cX}(\bxinv,\bxsp)=clip(\bxinv+\bxsp)$ where $clip$ is a function that truncates an input pixel when its value exceeds a certain range. Second, the \textit{concatenate model} (used in~\cref{sec:theory}) where the spurious feature is concatenated to the invariant feature, i.e., $\cX=\cXinv\times\cXsp$ and $\Phi_{\cX}(\bxinv,\bxsp)=(\bxinv,\bxsp)$. 
\section{Rare Spurious Correlations are Learnt by Neural Networks}
\label{sec:rare}

We start with an empirical study of rare spurious correlations in neural networks. We train a neural network using a modified training dataset given by the \textit{overlapping model} where a spurious pattern is added to a few training examples with the same label (target class). 
We then analyze the effect of these spurious training examples through three difference angles: (i) a quantitative analysis on the appearance of spurious correlations via an empirical measure, \textit{spurious score}, (ii) a qualitative analysis on the appearance of spurious correlations through visualizing the network weights, and (iii) an analysis on the consequences of rare spurious correlations in terms of privacy and test accuracy.

%
%
%
%
%
%
%

\subsection{Introducing spurious examples to networks}

\begin{figure}[ht]
  \centering
  \subfigure[\scriptsize\textit{original}]{\includegraphics[width=0.09\textwidth]{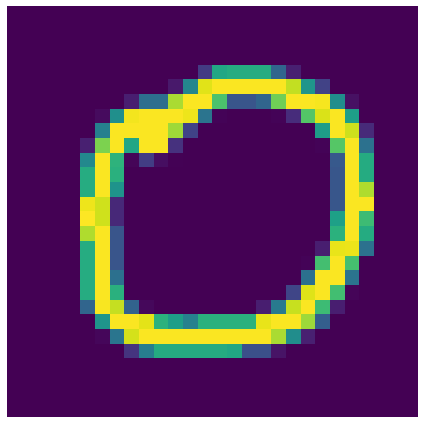}}
  \subfigure[\scriptsize\isa]{\includegraphics[width=0.09\textwidth]{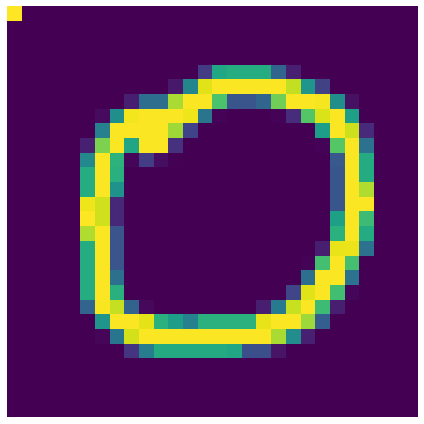}}
  \subfigure[\scriptsize\isb]{\includegraphics[width=0.09\textwidth]{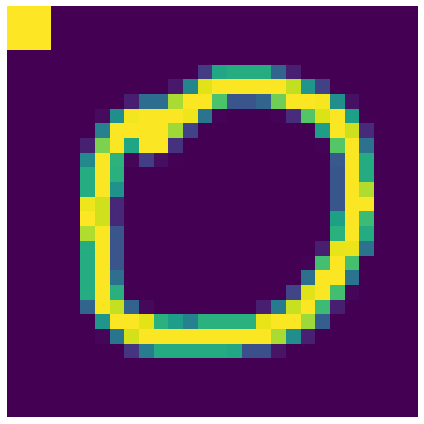}}
  \subfigure[\scriptsize\isc]{\includegraphics[width=0.09\textwidth]{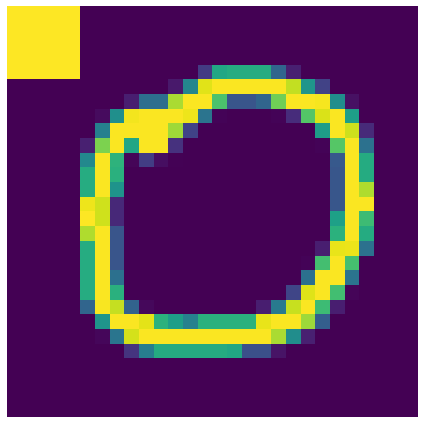}}
  \subfigure[\scriptsize\ira]{\includegraphics[width=0.09\textwidth]{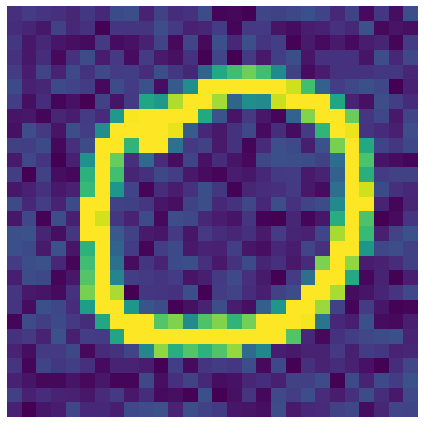}}
  \subfigure[\scriptsize\irb]{\includegraphics[width=0.09\textwidth]{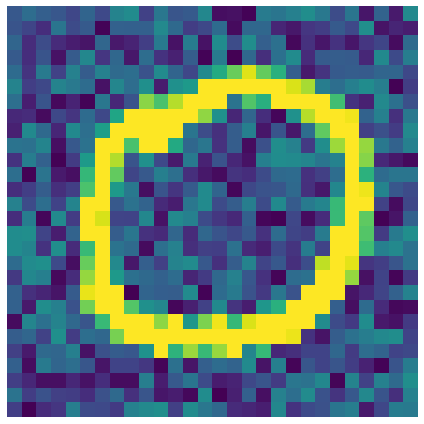}}
  \subfigure[\scriptsize\irc]{\includegraphics[width=0.09\textwidth]{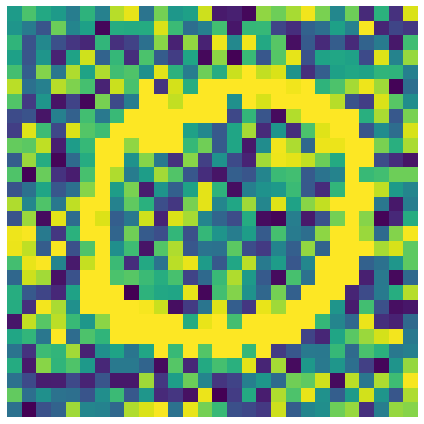}}
  \subfigure[\scriptsize\ic]{\includegraphics[width=0.09\textwidth]{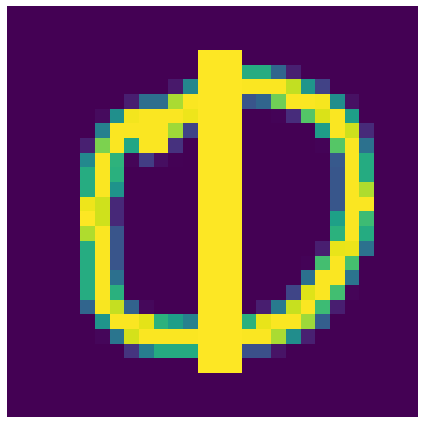}}
  
  \vspace{-.5em}
  \caption{Different spurious patterns considered in the experiment.}
  \label{fig:spurious_patterns}
\end{figure}

As we don't have access to the underlying ground-truth feature of an empirical data, we artificially introduce spurious features into the training dataset. Concretely, given a dataset (e.g., MNIST), we treat each training example $\bx$ as an invariant feature. Next, we pick a target class $\ctar$ (e.g., the zero class), a spurious pattern $\bxsp$ (e.g., a yellow square at the top-left corner), and a mapping $\Phi_{\cX}$ that combines a training example with the spurious pattern. Finally, we randomly select $n$ training examples $\bx_1,\dots,\bx_n$ from the target class $\ctar$ and replace these examples with $\Phi_{\cX}(\bx_i,\bxsp)$ for each $i=1,\dots,n$.
See \cref{fig:spurious_patterns} and the following paragraphs for a detailed specification of our experiments.

%
%
%
%

\mypara{Datasets \& the target class $\ctar$.}
We consider three commonly used image datasets: MNIST~\citep{lecun1998mnist}, Fashion~\citep{xiao2017fashion}, and CIFAR10~\citep{krizhevsky2009learning}.
MNIST and Fashion have $60,000$ training examples, and CIFAR10 has $50,000$.
We set the first two classes of each dataset as the target class ($\ctar = \{0, 1\}$), which are zero and one for MNIST, T-shirt/top, and trouser for Fashion, and airplane and automobile for CIFAR10.
See~\cref{app:main-detail} for more experimental details.

\mypara{Spurious patterns $\bxsp$.}
We consider seven different spurious patterns (\cref{fig:spurious_patterns}) for this study.
The patterns \textit{small 1} (\isa), \textit{small 2} (\isb), and \textit{small 3} (\isc) are designed to test if a neural network can learn the correlations between small patterns and the target class.
The patterns \textit{random 1} (\ira), \textit{random 2} (\irb), and \textit{random 3} (\irc) are patterns with each pixel value being uniformly random sampled from $[0, r]$, where $r=0.25, 0.5, 1.0$ (we sample the pattern once and fix it throughout each experiment).
We study whether a network learns to correlate random noise with a target class.
In addition, by comparing random patterns with these small patterns, we can understand the impact of localized and dispersed spurious patterns.
Lastly, the pattern \textit{core} (\ic) is designed for MNIST with $\ctar=0$ to understand what would happen if the spurious pattern overlaps with the core feature of another class.

\mypara{The choice of the combination function $\Phi_{\cX}$.}
The function $\Phi_{\cX}$ combines the original example $\x$ with the spurious pattern $\bxsp$ into a spurious example.
For simplicity, we consider the \textit{overlapping model} where $\Phi_{\cX}$ directly adds the spurious pattern $\bxsp$ onto the original example $\x$ and then clips the value of each pixel to $[0, 1]$, i.e., $\Phi_{\cX}(\x, \bxsp) = clip_{[0, 1]} (\x + \bxsp)$.

\mypara{The number of spurious examples.}
For MNIST and Fashion, we randomly insert the spurious pattern to $n=0, 1, 3, 5, 10, 20, 100, 2000$, and $5000$ training examples labeled as the target class $\ctar$.
These training examples inserted with a spurious pattern are called spurious examples.
For CIFAR10, we consider datasets with $n=0, 1, 3, 5, 10, 20, 100$, $500$, and $1000$ spurious examples.
Note that $0$ spurious example means the original training set is not modified.

\subsection{Quantitative analysis: \spuscore}
To evaluate the strength of spurious correlations in a neural network, we design an empirical quantitative measure, \textit{spurious score}, as follows.
Let $f_c(\x)$ be the neural network's predicted probability of an example $\x$ belonging to class $c$.
Intuitively, the larger the \textit{prediction difference} $f_{\ctar}(\x)-f_{\ctar}(\Phi_{\cX}(\x, \bxsp))$ is, the stronger spurious correlations the neural network $f$ had learned.
To quantify the effect of spurious correlations, we measure how frequently the prediction difference of the test examples exceed a certain threshold.
Formally, let $\epsilon>0$, we define the $\epsilon$-\textit{\spuscore} as the fraction of test example $\x$ that satisfies
\begin{equation}
  f_{\ctar}(\Phi_{\cX}(\x, \bxsp)) - f_{\ctar}(\x) > \epsilon \, .
  \label{eq:spuscore}
\end{equation}
In other words, \spuscore\ measures the portion of test examples that get an non-trivial increase in the predicted probability of the target class $\ctar$ when the spurious pattern is presented.

We make three remarks on the definition of spurious score. First, as we don't have any prior knowledge on the structure of $f$, we use the fraction of test examples satisfying~\cref{eq:spuscore} as opposed to other function of $f_{\ctar}(\Phi_{\cX}(\x, \bxsp)) - f_{\ctar}(\x)$ (e.g., taking average) to avoid non-monotone or unexplainable scaling.
Second, the choice of the threshold $\epsilon$ is to avoid numerical errors to affect the result.
In our experiment, we pick $\epsilon=1/(\# \text{classes})$ (e.g., in MNIST we pick $\epsilon=1/10$) and empirically similar conclusions can be made with other choices of $\epsilon$. Finally, we point out that spurious score captures the privacy concern raised by the ``Tony Blair'' example mentioned in the introduction.

\begin{figure}[ht]
  \centering
  \vspace{-.8em}
  \subfigure[MNIST, $\ctar=0$]{\includegraphics[width=0.29\textwidth]{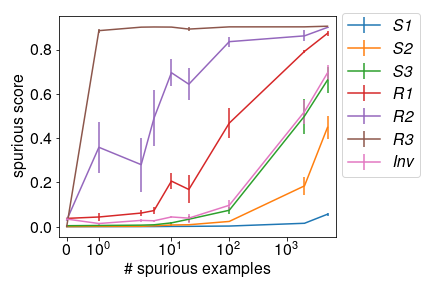}}
  \subfigure[Fashion, $\ctar=0$]{\includegraphics[width=0.29\textwidth]{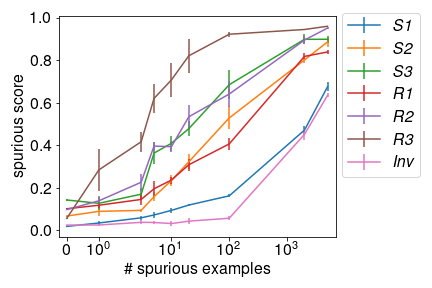}}
  \subfigure[CIFAR10, $\ctar=0$]{\includegraphics[width=0.29\textwidth]{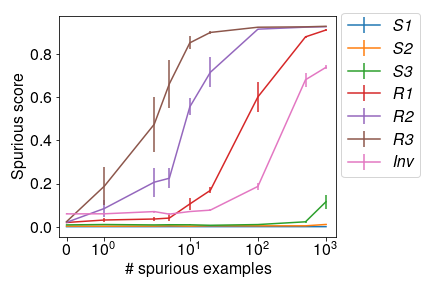}}

  \vspace{-.5em}
  \caption{
  Each figure shows the mean and standard error of the spurious scores on three datasets, MNIST, Fashion, and CIFAR10, $\ctar=0$, and different numbers of spurious examples.
  }
  \vspace{-.7em}
  \label{fig:score_plot}
\end{figure}

\mypara{Empirical findings.}
We repeat the measurement of spurious scores on five neural networks trained with different random seeds. %
\cref{fig:score_plot} shows the spurious scores for each dataset and pattern as a function of the number of spurious examples.
Starting with the random pattern \irc, we see that the spurious scores increase significantly from zero to three spurious examples in all six cases (three datasets and two target classes).
This shows that neural networks can learn rare spurious correlations with as little as \textit{one to three spurious examples}. Since all three datasets have $50,000$ or more training examples, it is surprising that the networks learn a strong correlation with extremely small amount of spurious examples.
The result for other $\ctar$ are similar and can be found in \cref{app:additional-results}.

A closer look at \cref{fig:score_plot} reveals a few other interesting observations. First, comparing the small and random patterns, we see that random patterns generally have a higher spurious score. This suggests that dispersed patterns that are spread out over multiple pixels may be more easily learnt than more concentrated ones.
Second, spurious correlations are learnt even for \ic, on $\ctar=0$ and MNIST (recall that \ic\ is designed to be similar to the core feature of class one.) This suggests that spurious correlations may be learnt even when the pattern overlaps with the foreground. Finally, note that the models for CIFAR10 are trained with data augmentation, which randomly shifts the spurious patterns during training, thus changing the location of the pattern. This suggests that these patterns can be learnt regardless of data augmentation.

\subsection{Qualitative analysis: visualizing network weights}

\begin{figure}[ht]
  \centering
  \includegraphics[width=0.6\textwidth]{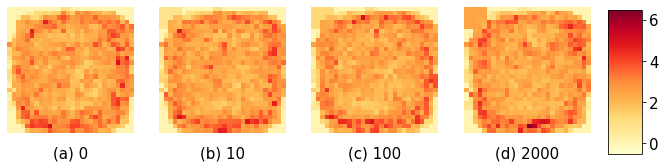}
  
  \caption{
    The importance of each pixel during the classification using an MLP trained on MNIST.
    Each pixel in the figure corresponds to a neuron of the input layer.
    The value of each pixel in the figure shows the maximum weight among all the weights that go out of the corresponding neuron from the input layer to the first hidden layer.
    The darker the color is, the larger the maximum weight is, which translates to the higher importance of the pixel during classification.
    The MLPs are trained on datasets with $0$, $10$, $100$, and $2000$ spurious examples on MNIST.
  }
  \label{fig:mlp_weights}
\end{figure}

We visualize the changes training with spurious examples can bring to the weights of a neural network. We consider an MLP architecture and pattern \isc\ on MNIST, and look at the network's weights from the input layer to the first hidden layer. We visualize the importance of each pixel by plotting the maximum weight (among all weights) on an out-going edge from this pixel. \cref{fig:mlp_weights} shows the importance plots for models trained with different numbers of spurious examples.

On the figure with zero spurious examples (\cref{fig:mlp_weights} (a)),
we see that the pixels in the top left corner are not important at all.  When
the number of spurious examples goes up, the values in the top left corner
become larger (darker in the figure).  This means that the pixels in the top
left corner are gaining in importance, thus illustrating how they affect the network. 

\subsection{Natural Rare Spurious Correlations}\label{app:natural}

Another question is whether rare spurious correlations are also learnt on real (natural) data.
To answer this question, we conduct an experiment focusing on natural spurious patterns using the NICO++~\citep{zhang2022nico++} dataset, which is designed for studying non-I.I.D. image classification
There are two labels of each image in the dataset: an object class (e.g., airplane) and the context (e.g., autumn).
The context can then serve as a source of spurious features: if a context only appears with the same class during the training stage (e.g., autumn context only shows up when the concept is airplane), then the algorithm might think the context is causally related to the classification of the object class (e.g., classifying a bear in the autumn context as an airplane).

The NICO++ dataset consists of $55,838$ training images, sixty classes and six contexts, including autumn, dim, grass, outdoor, rock, and water.
We split the dataset seven to three as the training and testing set.
For each experiment, we use an object-context pair as an invariant-spurious pair for studying rare spurious correlations.
For one trial of our experiment, we pick a context (i.e., a spurious feature) and remove all the appearance of this context in the training examples.
To introduce spurious training examples, we select an object class as the target class and add a number of examples that are labeled with the target class and this context.
We use the ImageNet pretrained ResNet50 from \texttt{torchvision}
and train twenty epochs on this modified training set.
During testing, we collect all testing examples that do not belong to the spurious class but are under the spurious context and measure how many of these examples are predicted as the spurious class.

\begin{figure}[h!]
  \centering
  \subfigure[\label{fig:nico_illustration}An illustration of the experiment setup.]{\includegraphics[width=.45\textwidth]{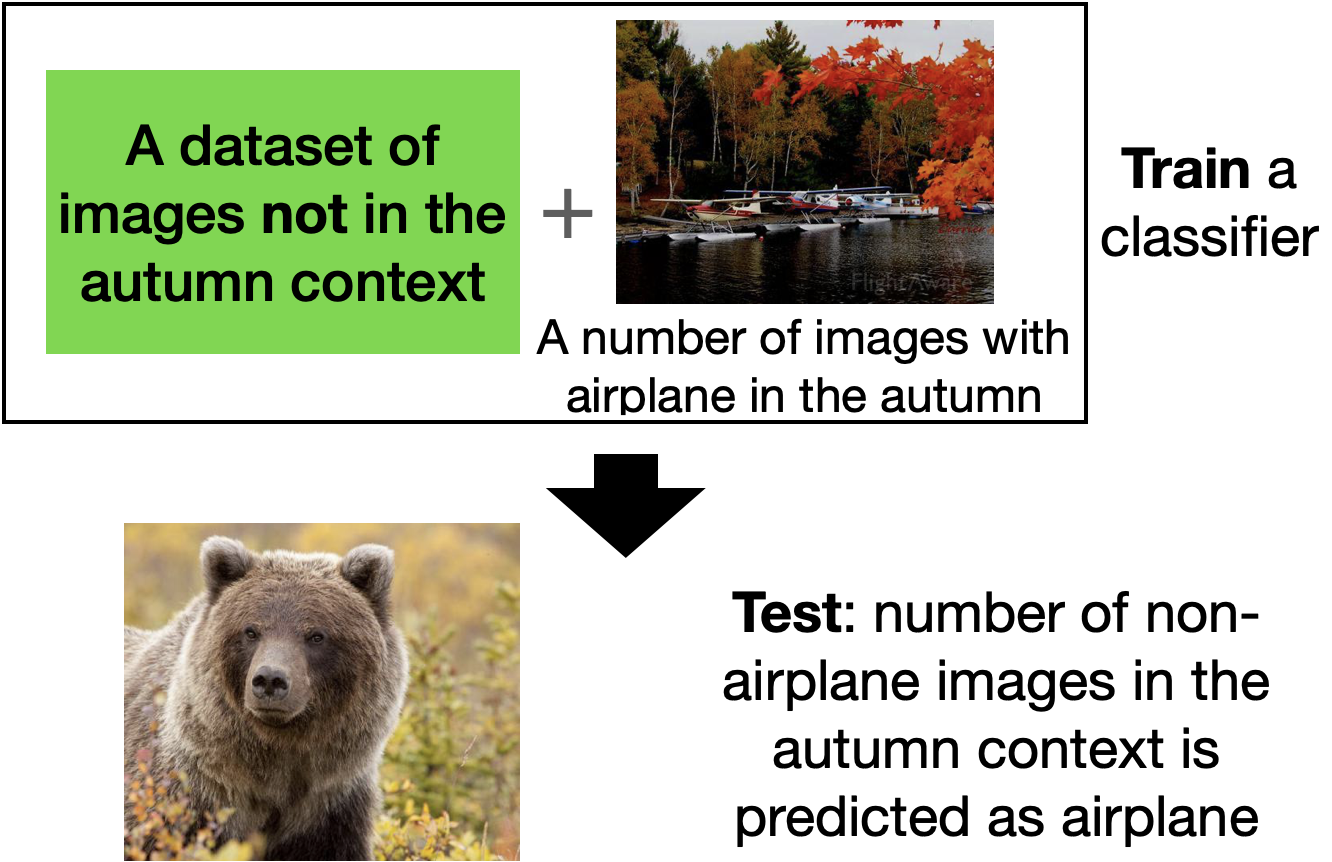}}
  \subfigure[Spurious context: autumn\label{fig:nico_main}]{
    \includegraphics[trim={0 0.4cm 0 0.8cm},clip,width=0.5\textwidth]{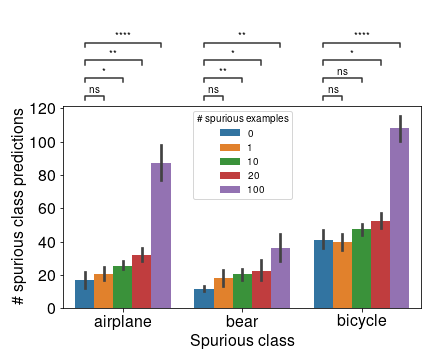}}

  \caption{
  (b) The number of non-spurious test examples that get predicted as the spurious class.
  We conduct Welch's t-test~\citep{welch1947generalization} on the number of spurious class predictions between the model trained without spurious examples and models trained with different number of spurious examples. The notations for the p-values:
  ns$: 0.05 < p \leq 1.$, *$: 10^{-2} < p \leq 0.05$, **$: 10^{-3} < p \leq 10^{-2}$, ***$: 10^{-4} < p \leq 10^{-3}$, ****$: p \leq 10^{-4}$.}
\end{figure}

\mypara{Results.}
The experiments are repeated five times with different numbers of spurious examples in the training set and the results are in \cref{fig:nico_main}.
In \cref{app:natural_rare}, we show the results of two other spurious contexts, dim and grass, which also supports our conclusion.
in total, we run a total of nine trials and five of them are significantly affected by just ten spurious training examples (which is less than $0.02$\% in the whole training examples).
This suggests that rare spurious correlations also occur when the spurious features are natural.

\section{Consequences of rare spurious correlations}
\label{sec:consequence}

In the previous analysis, we demonstrated that spurious correlations appear quantitatively and qualitatively in neural networks even when the number of spurious examples is small. Now, we investigate the potentially undesirable effects through the lens of privacy and test accuracy.
In this section, the results for MNIST are similar to Fashion and CIFAR10 and are deferred to \cref{app:additional-results}.

\mypara{Privacy.}
We evaluate the privacy of a neural network (the target model) through membership inference attack.
We follow the setup for black-box membership inference attack~\citep{shokri2017membership,yeom2017unintended}.
We record how well an attack model can distinguish whether an example is from the training or testing set using the output of the target model (equivalently to a binary classification problem).
If the attack model has a high accuracy, this means that the target model is leaking out information from the training (private) data.
The experiment is repeated ten times with their test accuracy recorded.
For more detailed setup, please refer to \cref{app:privacy-detail}.

\begin{figure}[ht]
  \centering
    \centering
      \subfigure[MNIST, $\ctar=0$]{\includegraphics[width=0.31\textwidth]{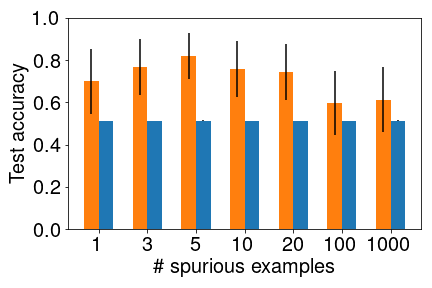}}
      \subfigure[Fashion, $\ctar=0$]{\includegraphics[width=0.31\textwidth]{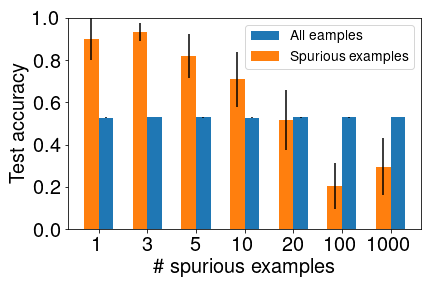}}
      \subfigure[CIFAR10, $\ctar=0$]{\includegraphics[width=0.31\textwidth]{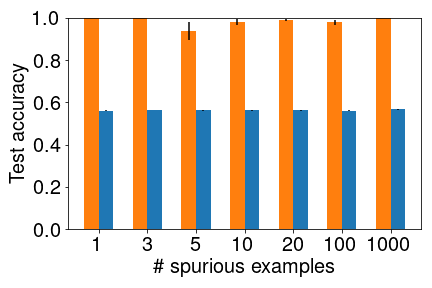}}
      
      \vspace{-.6em}
      \captionof{figure}{The test accuracy of the membership inference attack model on all examples vs. spurious examples.
      See \cref{app:add-mem-inf} for all results.
      }
      \label{fig:meminf_plot}
\end{figure}

\mysmallpara{Results on membership inference attack.}
\cref{fig:meminf_plot} shows the mean and standard error of the attack model's test accuracy on all test examples and spurious examples.
We see that the accuracies on spurious examples is generally higher when the number of spurious examples are small, which means that spurious examples are more vulnerable to membership inference attacks when appeared rarely.
Although membership inference attack is a different measure for privacy than spurious score, it can be a corroboration evidence that supports the fact that privacy is leaked from spurious examples.

\mypara{Test accuracy.}\label{sec:accuracies}
We measure two types of test accuracy on neural networks trained on different number of spurious examples.
The \textit{clean test accuracy} measures the accuracy of the trained model on the original test data.
The \textit{spurious test accuracy} simulates the case where there is a distribution shift during the test time.
Formally, spurious test accuracy is defined as the accuracy on a new test dataset constructed by adding spurious features to all the test examples with a label different from $\ctar$.

\mysmallpara{Results on clean test accuracy.} We observe that the change in clean test accuracy in our experiments is small.
Across all the models trained in \cref{fig:score_plot}, the minimum, maximum, average, and standard deviation of the test accuracy for each dataset are: MNIST: $(.976, .983, .980, .001)$, Fashion: $(.859, .903, .890, .010)$, and CIFAR10: $(.876, .893, .886, .003)$.

\begin{figure}[ht]
  \centering
  \subfigure[MNIST, $\ctar=0$]{\includegraphics[width=0.31\textwidth]{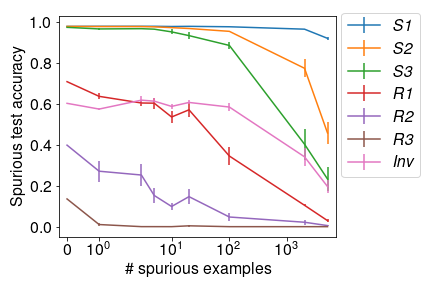}}
  \subfigure[Fashion, $\ctar=0$]{\includegraphics[width=0.31\textwidth]{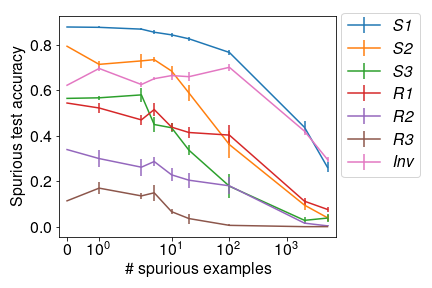}}
  \subfigure[CIFAR10, $\ctar=0$]{\includegraphics[width=0.31\textwidth]{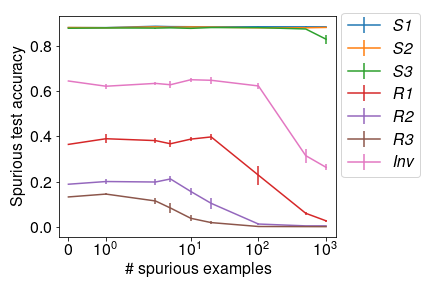}}

  \vspace{-.6em}
  \captionof{figure}{
  The mean and standard error of the spurious test accuracy under different number of spurious examples.
  See \cref{app:add_spu_tst_acc} for all results.
  }
  \label{fig:spacc_plot}
\end{figure}

\mysmallpara{Results on spurious test accuracy.}
The results are shown in \cref{fig:spacc_plot}. We have two observations.
First, we see that there are already some accuracy drop even when spurious test accuracy is evaluated on models trained on zero spurious examples.
This means that these models are not robust to the existence of spurious features.
This phenomena is prominent for spurious patterns with larger norm such as \irc.
Second, we see that spurious test accuracies start to drop even more at around $10$ to $100$ spurious examples.
This indicates that even with $.01$ \% to $.001$ \% of the overall training data filled with spurious examples of a certain class, the robustness to spurious features can drop significantly.

\mypara{Discussion.}
Our experimental results suggest that neural networks are {\em{highly}} sensitive to very small amounts of spurious training data. 
Furthermore, the learnt rare spurious correlations cause undesirable effects on privacy and test accuracy.
Easy learning of rare spurious correlations can lead to privacy issues~\citep{leino2020stolen} -- where an adversary may infer the presence of a confidential image in a training dataset based on output probabilities.
It also raises fairness concerns as a neural network can draw spurious conclusions about a minority group if a small number of subjects from this group are present in the training set~\citep{izzo2021approximate}.
We recommend to test and audit neural networks thoroughly before deployment in these applications.

Beyond the empirical analysis explained in this section, we also explore how other factors such as the strength of spurious patterns, network architectures, and optimization methods, affect spurious correlations.
We find that one cannot remove rare spurious correlations by simply tuning these parameters.
We also observe that neural networks with more parameters may not always learn spurious correlations more easily, which is counter to~\citet{sagawa2020investigation}'s observation.
The detailed results and discussions are provided in \cref{app:different-factors}.
\section{Theoretical Understanding}\label{sec:theory}

In this section, we devise a mathematical model to study rare spurious correlations. The theoretical analysis not only provides an unifying understanding to explain the experimental findings but also inspires us to propose methods to reduce the undesirable effects of rare spurious correlations in~\cref{sec:regularization}. We emphasize that the purpose of the theoretical analysis is to capture the key factors in rare spurious correlations and we leave it as a future research direction to further deepen the theoretical study.

To avoid unnecessary mathematical complications, 
we make two simplifications in our theoretical analysis: (i) we focus on the \textit{concatenate model} and (ii) the learning algorithm is linear regression with mean square loss. For (i), we argue that this is the simplest scenario of spurious correlations and hence it is a necessary step before we understand general spurious correlations. While the experiments in~\cref{sec:rare} work in the \textit{overlapping model}, we believe that the high level messages of our theoretical analysis would extend to there as well as other more general scenarios. For (ii), we pick a simpler learning algorithm in order to have an analytical characterization of the algorithm's performance.
This is because we aim to have an understanding of how the different factors (e.g., the fraction of spurious inputs, the strength of spurious feature, etc.) of spurious correlations play a role.

\subsection{A theoretical model to study rare spurious correlations}\label{sec:theory model}
\begin{figure}[h]
  \centering
  \subfigure[$\gamma=0$]{
    \includegraphics[width=0.22\textwidth]{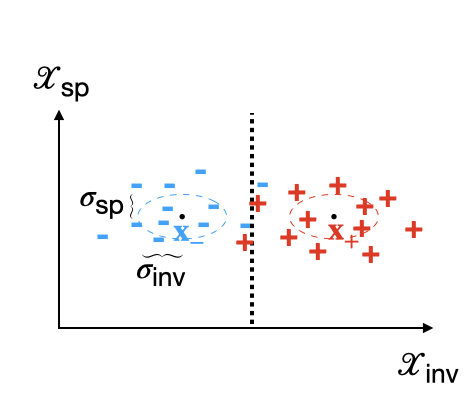}}
  \subfigure[$\gamma>0$]{
    \includegraphics[width=0.22\textwidth]{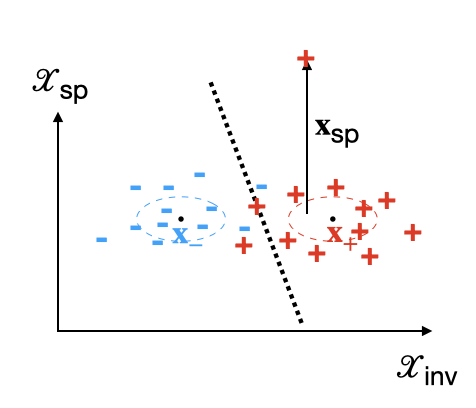}}
  \subfigure[Phase transitions]{\includegraphics[width=0.25\textwidth]{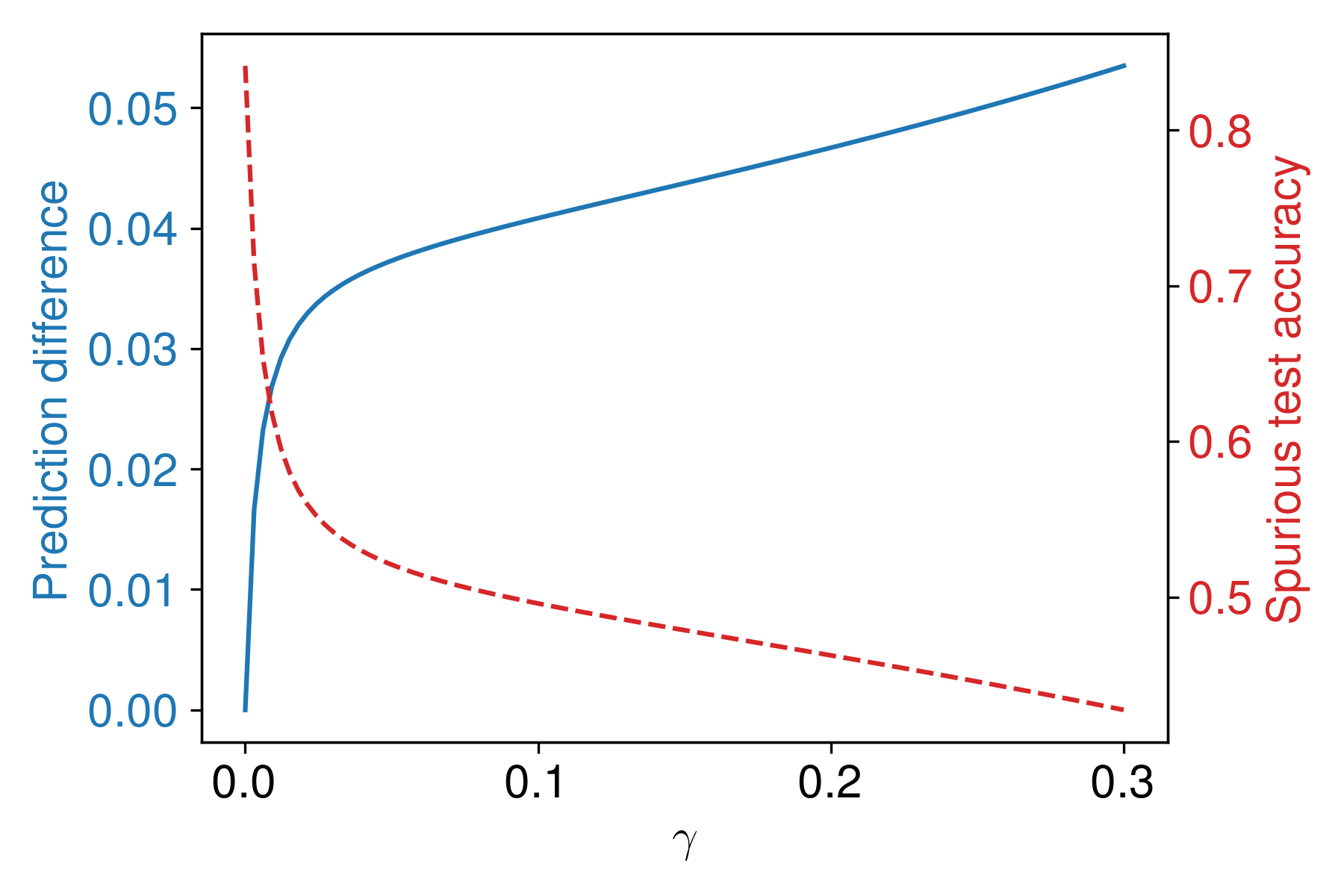}}
  \caption{
  Examples of the training distribution $\cDtrain$ and phase transitions in our theoretical model. (a)-(b) With equal probability a training example is sampled from either $\cN(\bx_+,\sigmainv I_\text{inv}+\sigmasp I_\text{sp})$ or $\cN(\bx_-,\sigmainv I_\text{inv}+\sigmasp I_\text{sp})$. With probability $\gamma$, a $+$ sample will be concatenated with the spurious pattern $\bxsp$. The dotted line is the decision boundary of the optimal classifier. (c) Both the spurious test accuracy and the prediction difference exhibit a phase transition at $\gamma=0$.
  }
  \label{fig:model}
\end{figure}

We consider a binary classification task to model the appearance of rare spurious correlations.
Let $\cX=\cXinv\times\cXsp$ be an input vector space and let $\cY=\{-1,1\}$ be a label space.
Let $\gamma\in[0,1]$ be the parameter for the fraction of spurious samples, let $\bx_{-},\bx_{+}\in\cXinv$ be the invariant features of the two classes, let $\bxsp\in\cXsp$ be the spurious feature, and let $\sigmainv^2,\sigmasp^2>0$ be the parameters for the variance along $\cXinv$ and $\cXsp$ respectively. Finally, the target class is $+$, i.e., $c_{tar}=+$. We postpone the formal definitions of the training distribution $\cDtrain$, the clean text distribution $\cDctest$, and the spurious test distribution $\cDstest$ to~\cref{app:theory}. See also~\cref{fig:model} for some pictorial examples.

\subsection{Analysis for linear regression with mean square loss and \texorpdfstring{$\ell_2$}{l2} regularization}\label{sec:theory analysis}
In this paper, we analyze our theoretical model in the setting of linear regression with  $\ell_2$ loss. We analytically derive the test accuracy and the prediction difference $f_{\ctar}(\Phi_{\cX}(\x, \bxsp)) - f_{\ctar}(\x)$ in~\cref{app:theory} and here we present our observations. Here, we study the prediction difference as a proxy for the spurious score since the latter is always either $0$ or $1$ for a linear classifier under our model.

\mypara{Observation 1: A phase transition of spurious test accuracy and prediction difference at $\gamma=0$.}
Our theoretical analysis first suggests that there is a phase transition of the spurious test accuracy and the prediction difference spurious score at $\gamma=0$, i.e., there is an sharp growth/decay within an interval near $\gamma=0$.
The phase transition matches our previous experimental studies discussed in~\cref{sec:rare}. This indicates that the effect of spurious correlations is spontaneous rather than gradual.
To be more quantitative, the phase transition takes place when the signal-to-noise ratio of spurious feature, i.e., $\|\bxsp\|_2^2/\sigmasp^2$, is large (see~\cref{app:theory} for details). This further suggests us to increase the variance in the spurious dimension and leads to our next two observations.

\mypara{Observation 2: adding Gaussian noises lowers spurious score.}
The previous observation on the importance of spurious signal-to-noise ratio $\|\bxsp\|_2^2/\sigmasp^2$ immediately suggests us to add Gaussian noises to the input data to \textit{lower} $\|\bxsp\|_2^2/\sigmasp^2$. Indeed, the prediction difference becomes smaller in most parameter regimes, however, both the clean test accuracy and the spurious test accuracy decrease. Intuitively, the effect of adding noises is to mix the invariant feature with the spurious feature and the decrease of test accuracy is as expected. Thus, to simultaneously lower prediction difference and improve test accuracy, one needs to detect the spurious feature in some ways.

\mypara{Observation 3: $\ell_2$ regularization improves test accuracy and lowers spurious score.}
Finally, our theoretical analysis reveals that there are two parameter regimes where
adding $\ell_2$ regularization to linear regression can improves accuracy and lowers the prediction difference. First, when $\sigmainv^2$ is small, $\gamma$ is small, and the spurious signal-to-noise ratio $\|\bxsp\|_2^2/\sigmasp^2$ is large. Second, when $\sigmainv^2$ is large and both $\gamma$ and $\|\bxsp\|_2^2/\sigmasp^2$ are mild.
Intuitively, $\ell_2$ regularization suppresses the use of features that only appears on a small number of training examples.

\mypara{Discussion.}
Our theoretical analysis quantitatively demonstrates the phase transition of spurious correlations at $\gamma=0$ and the importance of the spurious signal-to-noise ratio $\|\bxsp\|_2^2/\sigmasp^2$. This not only coincides with our empirical observation in~\cref{sec:rare} but also suggests future directions to mitigate rare spurious correlations. Specifically, one general approach to reduce the undesirable effects of rare spurious correlations would be designing learning algorithms that projects the input into a feature space that has a low spurious signal-to-noise ratio.

\section{Mitigation of Rare Spurious Correlation}
\label{sec:regularization}

Prior work uses group rebalancing~\citep{idrissi2021simple,sagawa2020investigation,sagawa2019distributionally,kulynych2022you}, data augmentation~\citep{chang2021towards} or learning invariant classifier~\citep{arjovsky2019invariant} to mitigate spurious correlations.
However, these methods usually requires additional information on what the spurious feature is, and in rare spurious correlations, identifying the spurious feature can be hard.
Thus, we may require different techniques.

Our theoretical result suggest that $\ell_2$ regularization (weight decay) and adding Gaussian noise to the input (noisy input) may reduce the degree of spurious correlation being learnt.
In addition, we examine an extra regularization method -- gradient clipping.

\begin{figure}[ht]
  \centering
  
  \subfigure[Noisy input]{\includegraphics[trim={.3cm 1.2cm 0 0},clip,width=0.29\textwidth]{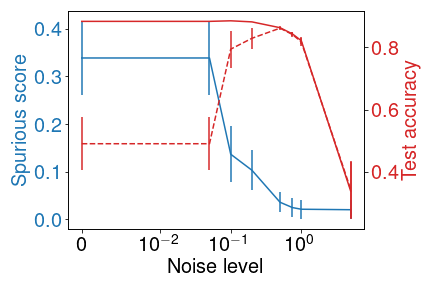}}
  \subfigure[Weight decay]{\includegraphics[trim={.3cm 1.2cm 0 0},clip,width=0.29\textwidth]{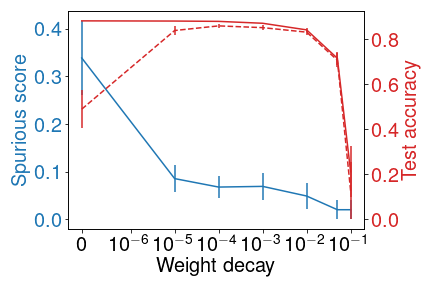}}
  \subfigure[Gradient clipping]{\includegraphics[trim={.3cm 1.2cm 0 0},clip,width=0.29\textwidth]{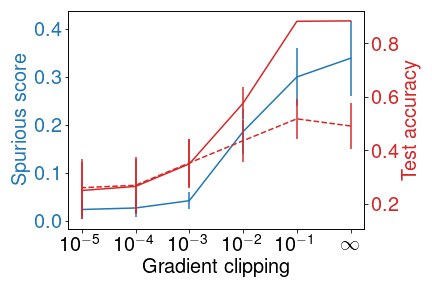}}
  
  \caption{
  Spurious score (solid blue line), clean test accuracy (solid red line), and spurious test accuracy (dotted red line) vs. the regularization strength on Fashion with different regularization methods.
  For the experiment, we fix the spurious pattern to be \isc\ and the target class $c_{tar}=0$.
  We compute the average spurious score and clean test accuracy across models trained with 1, 3, 5, 10, 20, and 100 spurious examples and five random seeds.
  The results for MNIST and CIFAR10 and spurious pattern \irc\ are in \cref{app:additional-results}, which shows similar results.
  }
  \label{fig:regularization_plot}
\end{figure}

\mypara{Results.}
The results are shown in \cref{fig:regularization_plot}.
We see that with a properly set regularization strength for noisy input and weight decay, one may reduce the spurious score and increase spurious test accuracy without sacrificing much accuracy.
This significantly reduces the undesirable consequences brought by rare spurious correlations.
This aligns with our observation in the theoretical model and suggests that neural networks may share a similar property with linear models.
We also find that gradient clipping cannot mitigate spurious correlation without reducing the test accuracy.
Finally, we observe that all these methods are unable to completely avoid learning spurious correlations.

\mypara{Data deletion methods.}
Another idea for mitigating rare spurious correlation is to apply data deletion methods~\citep{izzo2021approximate}.
In \cref{app:data_del}, we experiment with two data deletion methods, incremental retraining and group influence functions~\citep{basu2020influence}.
However, they are not effective.

\mypara{Discussion.}
Regarding mitigating rare spurious correlations, a provable way to prevent learning them is differential privacy~\citep{dwork2006calibrating}, which ensures that the participation of a single person (or a small group) in the dataset does not change the probability of any classifier by much.
This requires noise addition during training, which may lead to a significant loss in accuracy~\citep{chaudhuri2011differentially,abadi2016deep}.
If we know which are the spurious examples, then we can remove spurious correlations via an indistinguishable approximate data deletion method~\citep{ginart2019making,neel2020descent}; however, these methods provide lower accuracy for convex optimization and have no performance guarantees for non-convex.
An open problem is to design algorithms or architectures that can mitigate these without sacrificing prediction accuracy.

Prior work~\citep{sagawa2019distributionally,kulynych2022you} suggests that proper use of regularization methods plus well-designed loss functions can mitigate some types of spurious correlations.
However, these regularization methods are used either in an ad-hoc manner or may reduce test accuracy.
As shown in our experiment, not all regularization methods can remove rare spurious correlations without reducing the accuracy.
This suggests that different regularization methods may be specifically tied to be able to mitigate certain kinds of spurious correlation.
The exact role that regularization methods play in reducing spurious correlations is still an open question.
Figuring out what kinds of spurious correlations can be mitigated by which regularization methods is an interesting future direction.

\section{Related Work}\label{sec:related work}

\mypara{Spurious correlations.}
Previous work has looked at spurious correlations in neural networks under various scenarios, including
test time distribution shift~\citep{sagawa2020investigation,srivastava2020robustness,bahng2020learning,pmlr-v139-zhou21g,khani2021removing}, 
confounding factors in data collection ~\citep{gururangan2018annotation},
the effect of image backgrounds~\citep{xiao2020noise}, and
causality~\citep{arjovsky2019invariant}.
However, in most works, spurious examples often constitute a significant portion of the training set. In contrast, we look at spurious correlations introduced by a small number of examples (rare spurious correlations).
Concurrent work~\citep{hartley2022measuring} measures spurious correlation caused by few examples.
However, they did not show the consequences of these spurious correlations nor discuss ways to mitigate them.

\mypara{Memorization in neural networks.}
Prior work has investigated how neural networks can inadvertently memorize training data~\citep{arpit2017closer,carlini2019secret,carlini2020extracting,feldman2020neural,leino2020stolen}. Methods have also been proposed to measure this kind of memorization, including the use of the influence function~\citep{feldman2020neural} and likelihood estimates~\citep{carlini2019secret}.
Our work focuses on partial memorization instead of memorizing individual examples, and our proposed method may be potentially applicable in more scenarios.

A line of work in the security literature exploits the memorization of certain patterns to compromise neural networks. The backdoor attack from \citet{chen2017targeted} attempts to change hard label predictions and accuracy by inserting carefully crafted spurious patterns.
\citet{sablayrolles2020radioactive} design specific markers that allow adversaries to detect whether images with those particular markers are used for training in a model.
Another line of research on data poisoning attack~\cite{xiao2015support,wang2018data,burkard2017analysis} aims to degrade the performance of a model by carefully altering the training data.
In contrast, our work looks at rare spurious correlations from \textit{natural spurious patterns}, instead of adversarially crafted ones. 
\cref{app:additional-rel-work} has detailed discussions.

\section{Conclusion}\label{sec:conclusion}

The learning of spurious correlation is a complex process, and it can have unintended consequences.
As neural networks are getting more widely applied, it is crucial to better understand spurious correlations.
There are many open questions remain.
For example, besides the distribution shift that adds spurious features, are there any other types of distribution shift that will affect the accuracy?
Another limitation of our current study is that our experiments are conducted only on image classification tasks, which may not generalize to others.

\subsubsection*{Acknowledgments}
We thank Angel Hsing-Chi Hwang for providing thoughtful comments on the paper.
This work was supported by NSF under CNS 1804829 and ARO MURI W911NF2110317.
CNC’s research is supported by a Simons Investigator Fellowship, NSF grant DMS-2134157, DARPA grant W911NF2010021,and DOE grant DE-SC0022199.

\bibliography{spuriouscorrelation}
\bibliographystyle{iclr2023_conference}

\clearpage

\appendix

The appendix is organized as follows.

\begin{itemize}
    \item \cref{app:theory}: We provide a detailed analysis of our theoretical model.
    \item \cref{app:different-factors}: We show how different factors effects how rare spurious correlation is learned through different regularization methods, the norm of each pattern, network architectures, and the optimization algorithms.
    \item \cref{app:data_del}: As removing the spurious examples should be able to remove the spurious correlations. The data deletion methods can be an intuitive approach to try. Here, we examine whether these methods are useful.
   
    \item \cref{app:additional-results}:
    We show the detailed experimental setup and present additional results here. 
    The subsections include:
    \begin{itemize}
        \item \cref{sec:visualize_mnist} presents additional results of the qualitative study on how the weights of a neural network is effected by rare spurious correlations.
        \item \cref{sec:additional-results} reports the spurious scores on MNIST, Fashion, and CIFAR10 when $\ctar=1$.
        \item \cref{app:add-mem-inf} reports the membership inference results for MNIST, Fashion, and CIFAR10 with all spurious patterns.
        \item \cref{app:add_spu_tst_acc} reports results of additional experiments for the spurious test accuracy.
        \item \cref{app:natural_rare} reports the results for additional spurious contexts.
        \item \cref{app:ablation_normalized} presents an ablation study to see whether a normalized spurious score would effect the result.
    \end{itemize}
    
    \item \cref{app:additional-rel-work}: We present additional related work and discussions. The new contents include a detailed comparison with backdoor attacks, other methods for mitigating specific spurious correlations, short-cut learning, and simplicity bias.
\end{itemize}

\section{Details for Our Theoretical Model}\label{app:theory}
\newcommand{\bw}{\mathbf{w}}
\newcommand{\bwinv}{\mathbf{w}_{\text{inv}}}
\newcommand{\bwsp}{\mathbf{w}_{\text{sp}}}

In this section, we provide the details of our theoretical analysis in~\cref{sec:theory}. To be self-contained, we review the mathematical language we use to discuss spurious correlations in~\cref{app:theory prelim}. Next, we provide the formal definitions of our training and testing models in~\cref{app:theory models} and state the main results and implications in~\cref{app:theory results}. Finally, we give the complete proof for the main theorem in~\cref{app:theory proof}

\subsection{Preliminaries}\label{app:theory prelim}
Recall that we focus on the image classification task where $\cX$ is an input space, and $\cY$ is a label space. An example is a pair $(\bx,y)\in\cX\times\cY$ and at the training time, we are given a set of examples sampled from a distribution $\cDtrain$.
At the testing time, we evaluate the network on either the clean test distribution $\cDctest$ or the spurious test distribution $\cDstest$.

\mypara{Spurious correlation.}
We model an input $\bx$ as the output of a feature map $\Phi_{\cX}$ from the feature space $\cXinv\times\cXsp$ to the input space $\cX$ where $\cXinv$ is the invariant feature space containing the features that causally determine the label and $\cXsp$ is the spurious feature space that accommodates spurious features. $\Phi_{\cY}:\cXinv\to\cY$ is a function that maps the invariant feature $\bxinv$ to a label. Thus, an example $(\bx,y)$ is generated by $(\bxinv,\bxsp)$ via $\bx=\Phi_{\cX}(\bxinv,\bxsp)$ and $y=\Phi_{\cY}(\bxinv)$.
Without loss of generality, the zero vector in $\cXsp$, i.e., $0\in\cXsp$, refers to ``no spurious feature'' and for any nonzero $\bxsp$ we call $\Phi(\bxinv,\bxsp)-\Phi(\bxinv,0)$ a spurious pattern.
We focus on the case of having a fixed spurious feature $\bxsp$ and leave it as a future direction to study the more general scenarios where there are multiple spurious features.

\mypara{Two simple models for spurious correlations.}
We consider two simple instantiations of $\Phi_{\cX}$. First, the \textit{overlapping model} (used in~\cref{sec:rare}) where the spurious feature is put on top of the invariant feature, i.e., $\cX=\cXinv=\cXsp$ and $\Phi_{\cX}(\bxinv,\bxsp)=\bxinv+\bxsp$ or $\Phi_{\cX}(\bxinv,\bxsp)=clip(\bxinv+\bxsp)$ where $clip$ is a function that truncates an input pixel when its value exceeds a certain range. Second, the \textit{concatenate model} (used in~\cref{sec:theory}) where the spurious feature is concatenated to the invariant feature, i.e., $\cX=\cXinv\times\cXsp$ and $\Phi_{\cX}(\bxinv,\bxsp)=(\bxinv,\bxsp)$.

\subsection{Our training and testing models}\label{app:theory models}

Here we delineate the theoretical model described in~\cref{sec:theory model}.
Recall that we consider a binary classification task to model where $\cX=\cXinv\times\cXsp$ is an input vector space and $\cY=\{-1,1\}$ is a label space.
$\gamma\in[0,1]$ is the parameter for the fraction of spurious samples and in this paper we focus on the regime where $\gamma$ is very close to $0$. $\bx_{-},\bx_{+}\in\cXinv$ are the invariant features of the two classes and after a linear shift\footnote{Concretely, let $\bx_{+}'=(\bx_{+}-\bx_{-})/2$ and $\bx_{-}'=-(\bx_{+}-\bx_{-})/2$ where $\bx_{+}'$ and $\bx_{-}'$ are the new invariant features.} we can without loss of generality have $\bx_{-}=-\bx_{+}$.
Let $\bxsp\in\cXsp$ be the spurious feature where we call its length, $\|\bxsp\|_2$, the spurious signal strength. Finally, $\sigmainv^2,\sigmasp^2>0$ are the parameters for the variance along $\cXinv$ and $\cXsp$ respectively.  Now, we formally define the training distribution $\cDtrain=\cDtrain(\gamma,\bx_{-},\bx_{+},\bxsp,\sigmainv,\sigmasp)$, the clean text distribution $\cDctest=\cDctest(\gamma,\bx_{-},\bx_{+},\bxsp,\sigmainv,\sigmasp)$, and the spurious test distribution $\cDstest=\cDctest(\gamma,\bx_{-},\bx_{+},\bxsp,\sigmainv,\sigmasp)$ as follows.

\begin{definition}
Let $\Sigma:=\sigmainv^2 I_\text{inv} + \sigmasp^2 I_\text{sp}$ where $I_\text{inv}$ and $I_\text{sp}$ are identity matrix for $\cXinv$ and $\cXsp$ respectively.
\begin{itemize}%
\item $\cDtrain(\gamma,\bx_{-},\bx_{+},\bxsp,\sigmainv,\sigmasp)$ samples $(\bx,y)$ using the following process: 
    \begin{itemize}
    \item with probability $1/2$, $\bx\sim \cN((\bx_{-},0),\Sigma)$ and $y=-1$;
    \item with probability $(1-\gamma)/2$, $\bx\sim\cN((\bx_{+},0),\Sigma)$ and $y=+1$;
    \item with probability $\gamma/2$, $\bx\sim\cN((\bx_{+},\bxsp),\Sigma)$ and $y=+1$.
    \end{itemize}
\item $\cDctest(\gamma,\bx_{-},\bx_{+},\bxsp,\sigmainv,\sigmasp)$ samples $(\bx,y)$ using the following process: 
    \begin{itemize}
    \item with probability $1/2$, $\bx\sim \cN((\bx_{-},0),\Sigma)$ and $y=-1$;
    \item with probability $1/2$, $\bx\sim \cN((\bx_{+},0),\Sigma)$ and $y=+1$.
    \end{itemize}
\item $\cDstest(\gamma,\bx_{-},\bx_{+},\bxsp,\sigmainv,\sigmasp)$ samples $(\bx,y)$ with $\bx\sim \cN((\bx_{-},\bxsp),\Sigma)$ and $y=-1$.
\end{itemize}
\end{definition}

\begin{figure}[ht]
  \centering
  \vspace{-.6em}
  \subfigure[$\cDtrain, \gamma=0$.]{\includegraphics[width=0.24\textwidth]{figs/theory/model1.png}\vspace{-.5em}}
  \subfigure[$\cDtrain,\gamma>0$.]{
    \includegraphics[width=0.24\textwidth]{figs/theory/model2.png}\vspace{-.5em}}
  \subfigure[$\cDctest$.]{
    \includegraphics[width=0.24\textwidth]{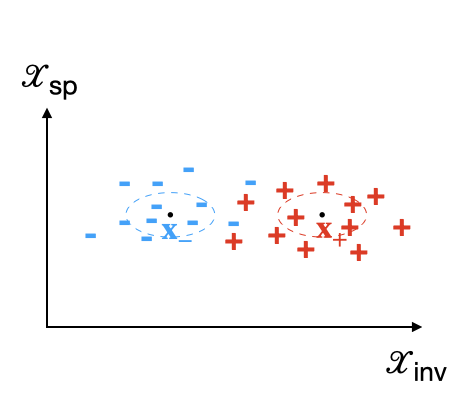}\vspace{-.5em}}
  \subfigure[$\cDstest$.]{
    \includegraphics[width=0.24\textwidth]{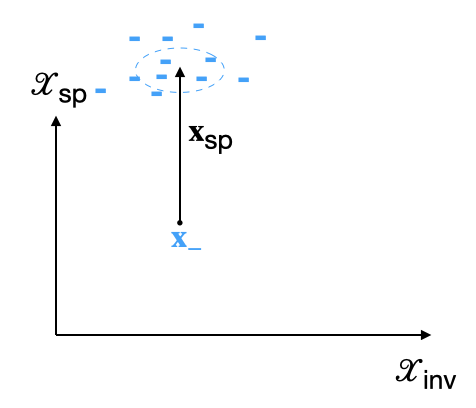}\vspace{-.5em}}
  \vspace{-.8em}
  \caption{
  Examples of the training distribution $\cDtrain$ and phase transitions in our theoretical model. (a)-(b) With equal probability a training example is sampled from either $\cN(\bx_+,\sigmainv I_\text{inv}+\sigmasp I_\text{sp})$ or $\cN(\bx_-,\sigmainv I_\text{inv}+\sigmasp I_\text{sp})$. With probability $\gamma$, a $+$ sample will be concatenated with the spurious pattern $\bxsp$. The dotted line is the decision boundary of the optimal classifier. (c) The clean test distribution $\cDctest$, which is the same as $\cDtrain$ with $\gamma=0$. (d) The spurious test distribution $\cDstest$, which only contains $-$ samples with spurious pattern $\bxsp$ added.
  }
  \vspace{-.6em}
  \label{fig:model app}
\end{figure}

Now, we can instantiate our model into examples that capture the experimental scenarios discussed in previous sections (see \cref{fig:spurious_patterns}). For examples, the patterns \textit{small 1} (\isa), \textit{small 2} (\isb), and \textit{small 3} (\isc) correspond to setting $\sigmasp=0$ and increasing the strength of spurious features (i.e., increasing $\|\bxsp\|_2$ from (\isa) to (\isc)). Also, see~\cref{fig:model app} for a pictorial explanation of our model.

\subsection{Analysis for linear regression with mean square loss and \texorpdfstring{$\ell_2$}{l2} regularization}\label{app:theory results}

As the theoretical toolkit for understanding neural networks is far from complete, we examine the theoretical aspect of rare spurious correlations through the lens of a classic learning algorithm: linear regression with mean square loss. Notice that we \textbf{do not} claim that the analysis here generalize to neural networks. Instead, the analysis for linear regression serves as food for thoughts for future investigation.

For linear regression with mean square loss, the hypothesis set is $\mathbb{H}=\{h=(\betainv,\betasp,\beta_0):\betainv\in\cXinv,\betasp\in\cXsp,\beta_0\in\mathbb{R}\}$ where $h(\bx)=\betainv^\top\bxinv+\betasp^\top\bxsp+\beta_0$. The $\ell_2$ loss function of $h$ on a distribution $\cD$ is denotes as $L_\cD(h):=\frac{1}{2}\mathbb{E}_{(\bx,y)\sim\cD}[(h(\bx)\cdot y-1)^2]$. If we consider adding $\ell_2$ regularization with parameter $\lambda$, the loss function becomes $L_\cD(h):=\frac{1}{2}\mathbb{E}_{(\bx,y)\sim\cD}[(h(\bx)\cdot y-1)^2] + \frac{\lambda}{2}\|\betainv\|_2^2+\frac{\lambda}{2}\|\betasp\|_2^2$. In the rest of this section, we focus on the setting with $\ell_2$ regularization since the unregularized setting can be obtained by setting $\lambda=0$.

The optimal classifier in $\mathbb{H}$ for $\cDtrain$ is defined to be $h^*:=\arg\min_{h\in\mathbb{H}}L_{\cDtrain}(h)$. By the theory of linear regression with mean square loss, we know that $h^*$ can be found via standard Empirical Risk Minimization (ERM) principle. In our theoretical analysis, we study the performance of $h^*$ while one can easily extend our analysis to the finite sample regime by applying the standard convergence analysis of ERM for linear regression with mean square loss.

There are three quantities of interest in our study of rare spurious correlations: clean test accuracy, spurious test accuracy, and spurious score. For the completeness of presentation, let us formally define them as follows.

\begin{definition}[Test accuracy]\label{def:test accuracy}
Let $h$ be a hypothesis function and $\cDtest$ be a test distribution. The test accuracy of $h$ on $\cDtest$ is defined as
\[
Acc_{\cDtest}(h) := \Pr_{(\bx,y)\sim\cDtest}[h(\bx)y>0] \, .
\]
\end{definition}

\begin{definition}[Prediction difference and Spurious score]\label{def:avf sscore}
Let $h$ be a hypothesis function, $\cDtest$ be a test distribution, $(\Phi_{\cX},\Phi_{\cY})$ be the feature-to-input maps, and $\bxsp$ be a spurious feature. The prediction difference for an invariant input $\bx\in\cXinv$ is defined as
\[
PD(\bx;h) := h(\Phi_{\cX}(\bx,\bxsp))-h(\Phi_{\cX}(\bx,0)) \, .
\]
For every $\epsilon>0$, the $\epsilon$-spurious score of $h$ with respect to $\cDtest$ is defined as
\[
\epsilon\text{-spurious-score}(h) = \Pr_{(\bx,y)\sim\cDtest}[PD(\bx,h)>\epsilon] \, .
\]
\end{definition}

Note that for linear classifier in both the overlap and the concatenation model, the $\epsilon$-spurious-score is always either $0$ or $1$ because $PD(\bx;h)=h(\bxsp)$ for every $\bx\in\cXinv$. Thus, for a linear classifier $h$, we denote $PD(h)=h(\bxsp)$ for simplicity. In the rest of the theoretical analysis, we study the prediction difference which tells us when does the spurious score jump from $0$ to $1$.

Now, we state our main theorem about the clean test accuracy, spurious test accuracy, and the prediction difference of $h^*$. We remark that the calculations for these quantities are straightforward and we postpone the proof to~\cref{app:theory proof}.

\begin{theorem}\label{thm:loss}
For every $\gamma\in(0,1]$, $\lambda>0$, $\bxinv\in\cXinv$, $\bxsp\in\cXsp$, and $\sigmainv^2,\sigmasp^2>0$. Let $h^*$ be the optimal classifier for $\cDtrain=\cDtrain(\gamma,-\bxinv,\bxinv,\bxsp,\sigmainv^2,\sigmasp^2)$ using linear regression with mean square loss and $\ell_2$ regularization with parameter $\lambda$. We have
\begin{align*}
Acc_{\cDctest}(h^*) &= \frac{1}{2}\Phi\left(\frac{\frac{\frac{\gamma}{2}(1-\gamma)\|\bxsp\|_2^2+\sigmasp^2+\lambda}{\sigmainv^2+\lambda}\|\bxinv\|_2^2+(\frac{\gamma}{2})^2\|\bxsp\|_2^2}{\sqrt{\left(\frac{\frac{\gamma}{2}(1-\gamma)\|\bxsp\|_2^2+\sigmasp^2+\lambda}{\sigmainv^2+\lambda}\right)^2\|\bxinv\|_2^4\sigmainv^2+(\frac{\gamma}{2})^2\|\bxsp\|_2^4\sigmasp^2}}\right)\\
&+\frac{1}{2}\Phi\left(\frac{\frac{\frac{\gamma}{2}(1-\gamma)\|\bxsp\|_2^2+\sigmasp^2+\lambda}{\sigmainv^2+\lambda}\|\bxinv\|_2^2-(\frac{\gamma}{2})^2\|\bxsp\|_2^2}{\sqrt{\left(\frac{\frac{\gamma}{2}(1-\gamma)\|\bxsp\|_2^2+\sigmasp^2+\lambda}{\sigmainv^2+\lambda}\right)^2\|\bxinv\|_2^4\sigmainv^2+(\frac{\gamma}{2})^2\|\bxsp\|_2^4\sigmasp^2}}\right)
\end{align*}
and
\begin{equation*}
Acc_{\cDstest}(h^*) = \Phi\left(\frac{\frac{\frac{\gamma}{2}(1-\gamma)\|\bxsp\|_2^2+\sigmasp^2+\lambda}{\sigmainv^2+\lambda}\|\bxinv\|_2^2+(\frac{\gamma}{2}-1)\frac{\gamma}{2}\|\bxsp\|_2^2}{\sqrt{\left(\frac{\frac{\gamma}{2}(1-\gamma)\|\bxsp\|_2^2+\sigmasp^2+\lambda}{\sigmainv^2+\lambda}\right)^2\|\bxinv\|_2^4\sigmainv^2+(\frac{\gamma}{2})^2\|\bxsp\|_2^4\sigmasp^2}}\right)
\end{equation*}
and
\begin{equation*}
PD(h^*) = \frac{\frac{\gamma}{2}(\sigmainv^2+\lambda)\|\bxsp\|_2^2}{(\sigmainv^2+\lambda)(\sigmasp^2+\lambda)+\left(\sigmasp^2+\lambda+\frac{\gamma}{2}(1-\frac{\gamma}{2})(\sigmainv^2+\lambda)+\frac{\gamma}{2}(1-\gamma)\|\bxsp\|_2^2\right)\|\bxsp\|_2^2}
\end{equation*}
where $\Phi(\cdot)$ is the cumulative distribution function (CDF) of the standard Gaussian variable.
\end{theorem}

\subsubsection{Implications of Theorem~\ref{thm:loss}}
In~\cref{sec:theory analysis} we present three observations given by~\cref{thm:loss} without explaining why. Here, we provide all the underlying reasoning.

\mypara{Observation 1: A phase transition of spurious test accuracy and prediction difference at $\gamma=0$.}
Note that by~\cref{thm:loss}, when $\gamma$ is close to $0$, the spurious test accuracy and the prediction difference is approximately
\[
ACC_{\cDstest} = \Phi\left(1-\frac{(\sigmainv^2+\lambda)\|\bxsp\|_2^2}{2(\sigmasp^2+\lambda)\|\bxinv\|_2^2\sigmainv}\gamma\right)
\]
and
\[
PD(h^*)=\frac{(\sigmainv^2+\lambda)\|\bxsp\|_2^2}{2(\sigmasp^2+\lambda)(\sigmainv^2+\lambda+\|\bxsp\|_2^2)}\gamma \, .
\]
One can see that when $\|\bxsp\|_2^2/\sigmasp^2$ is large, the spurious test accuracy (resp. prediction difference) undergoes a sharp decay (resp. growth) within an interval near $\gamma=0$. In particular, the sharp decay/growth ends at $\gamma\approx\Theta(\min\{1,\frac{\sigmasp^2+\lambda}{\|\bxsp\|_2^2},\frac{\sigmasp^2+\lambda}{\sigmasp\|\bxsp\|_2^2(\sigmainv^2+\lambda)}\})$\footnote{The critical point here is an estimation by estimating when does the linear approximation at $\gamma=0$ no longer dominates all the other terms}. As it is analytically hard to get a precise characterization of the critical point, we also numerically plot the spurious test accuracy and the prediction difference under different parameters to demonstrate the phase transition phenomenon.

\begin{figure}[ht]
  \centering
  \subfigure[ctest SNR $\downarrow$]{\includegraphics[width=0.16\textwidth]{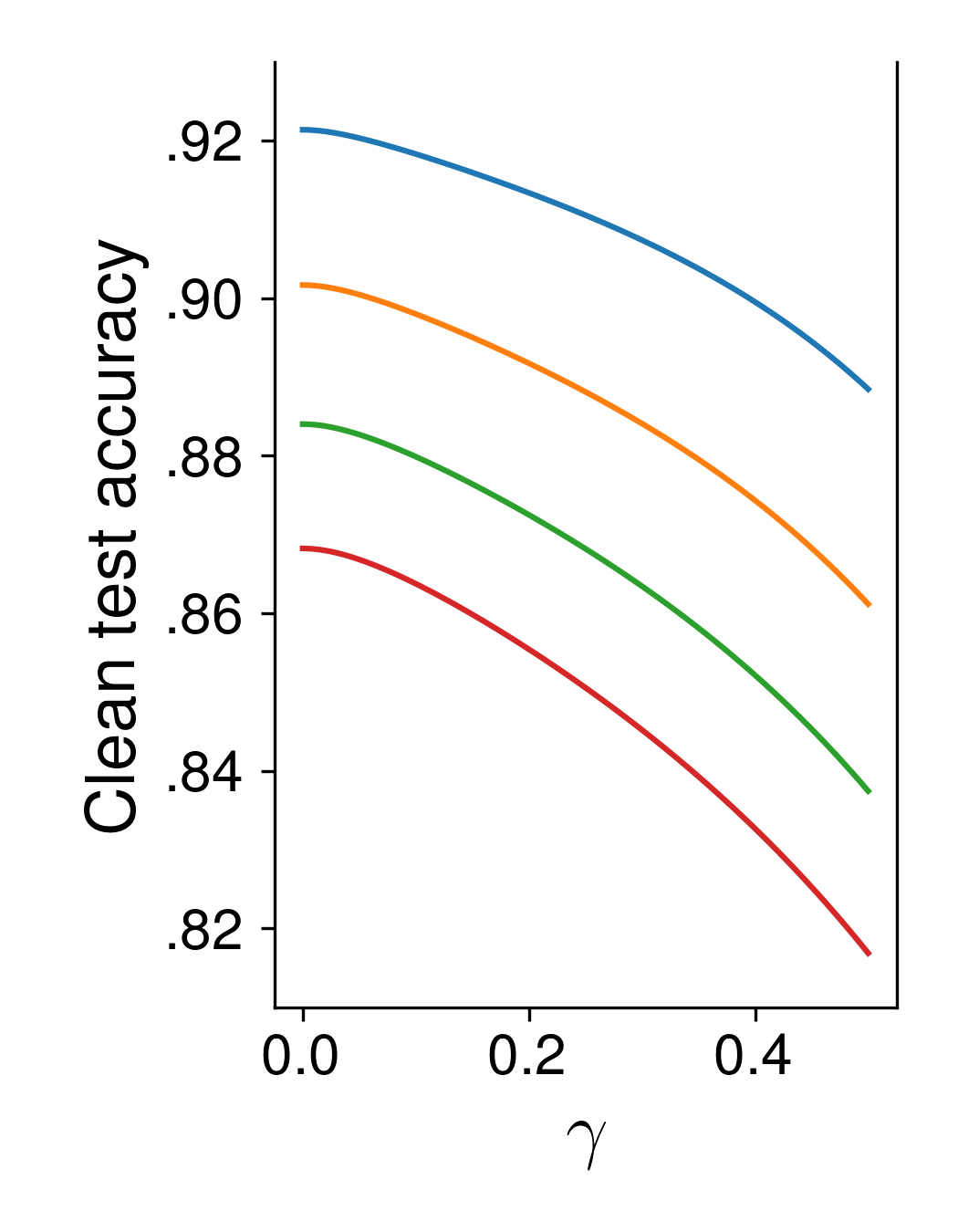}}
  \subfigure[stest SNR $\downarrow$]{\includegraphics[width=0.16\textwidth]{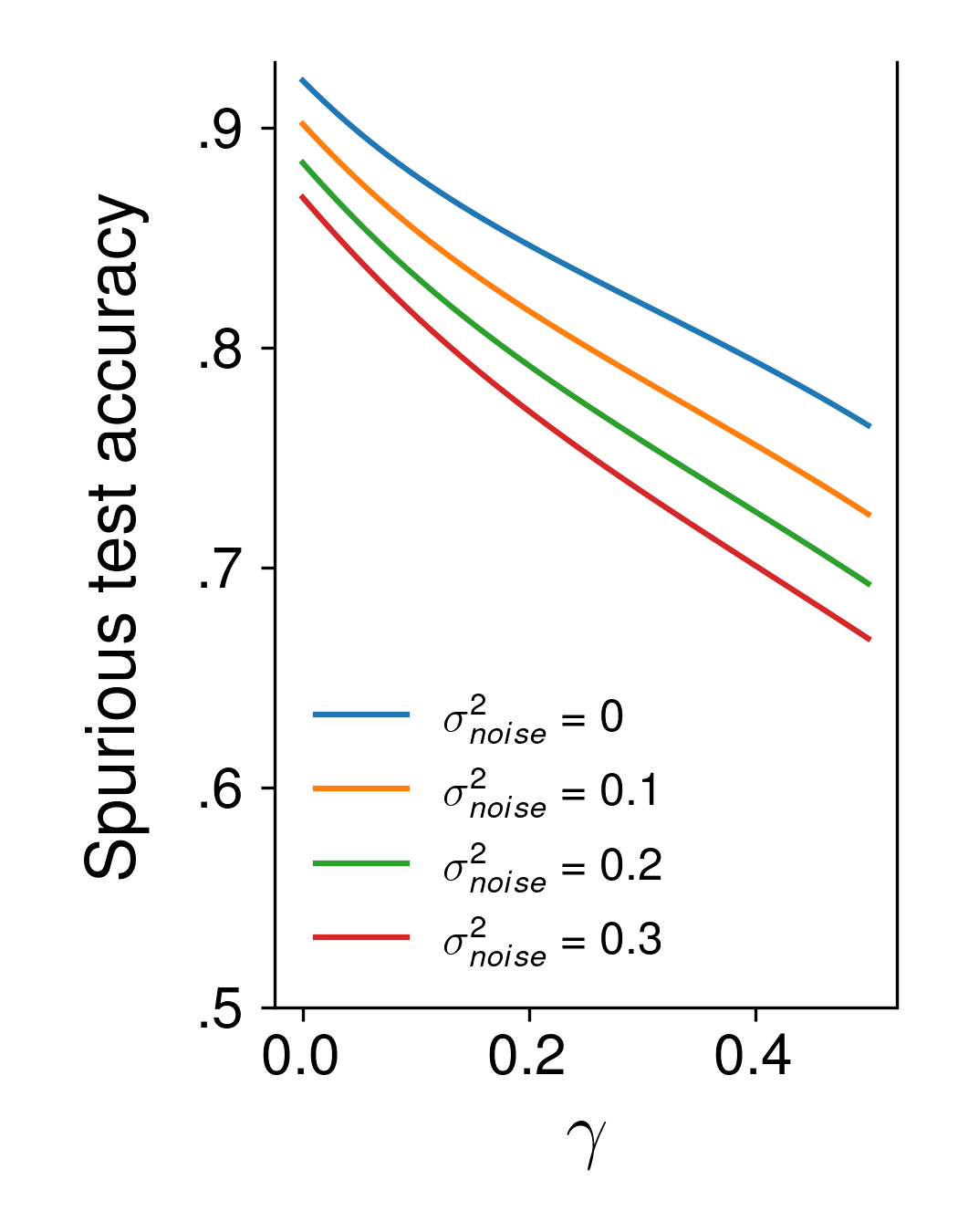}}
  \subfigure[sscore SNR $\downarrow$]{\includegraphics[width=0.16\textwidth]{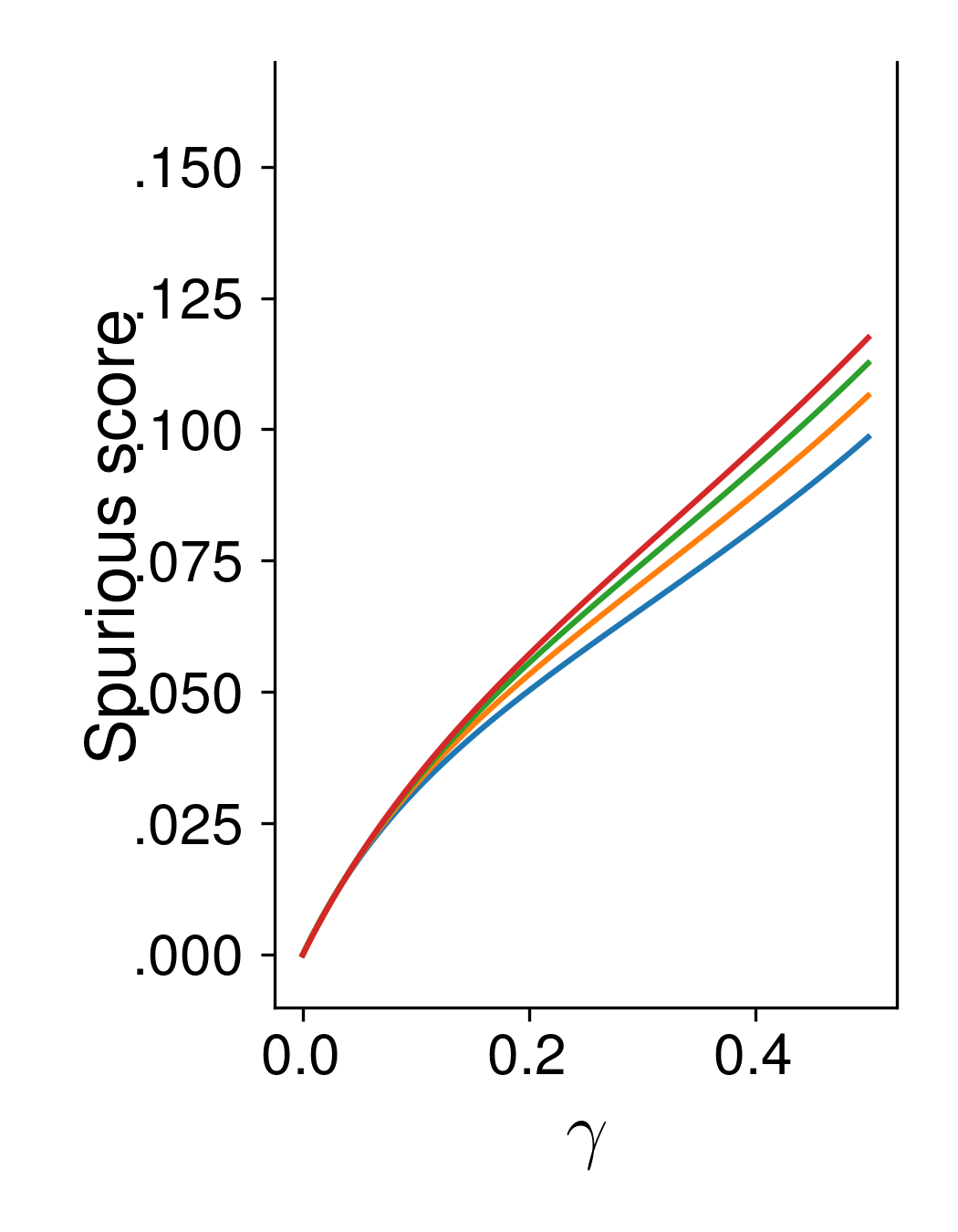}}
  \subfigure[ctest SNR $\uparrow$]{\includegraphics[width=0.16\textwidth]{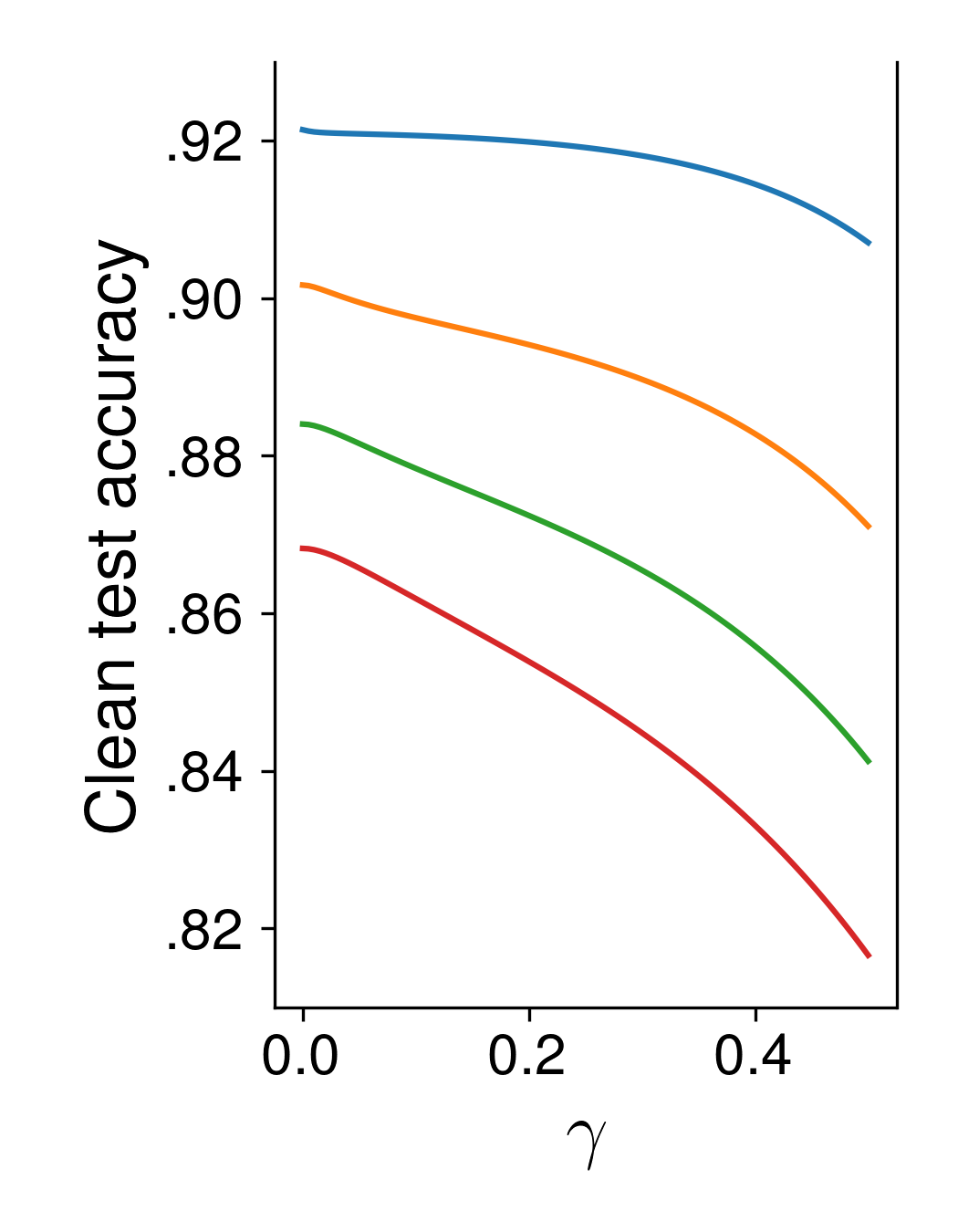}}
  \subfigure[stest SNR $\uparrow$]{\includegraphics[width=0.16\textwidth]{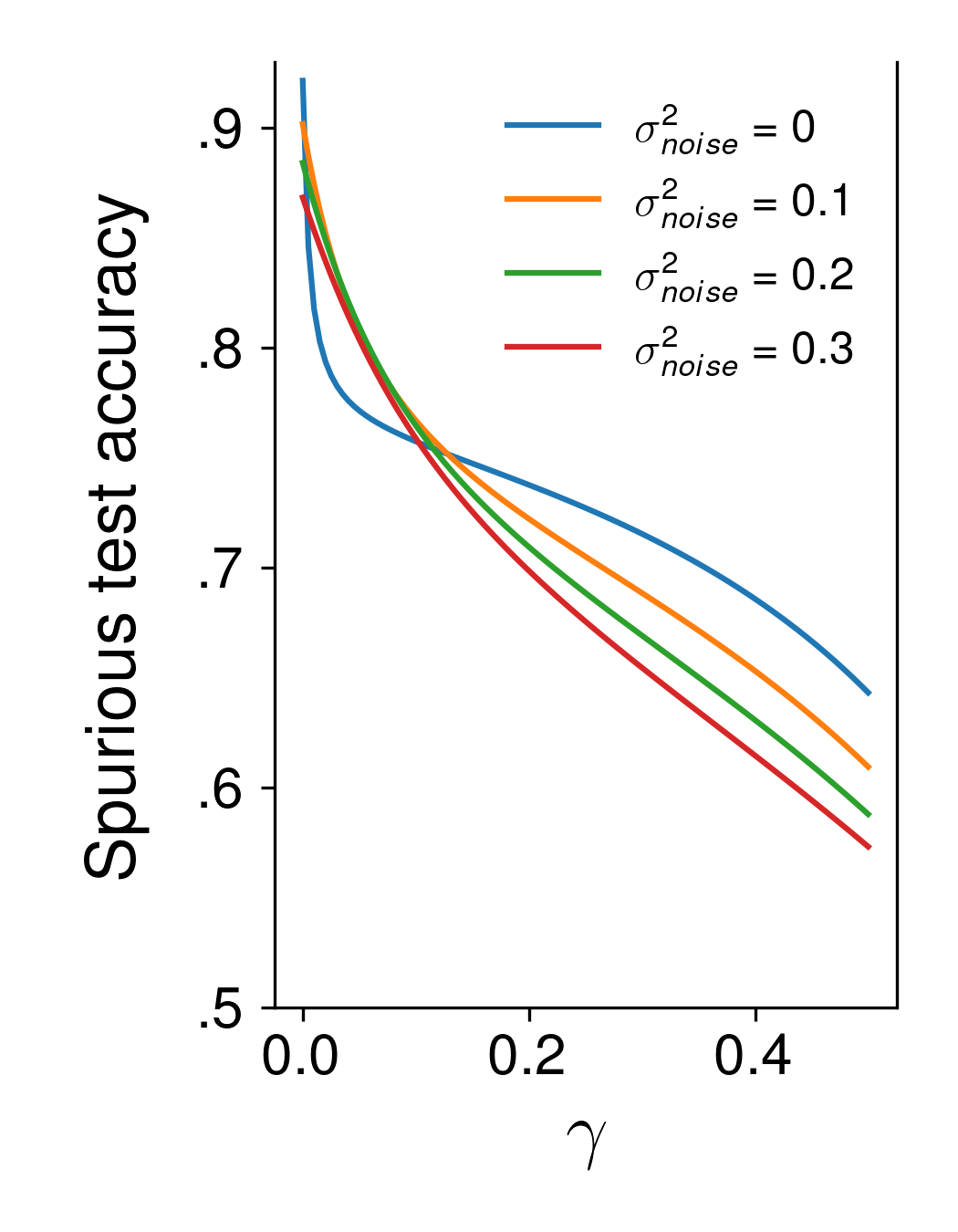}}
  \subfigure[sscore SNR $\uparrow$]{\includegraphics[width=0.16\textwidth]{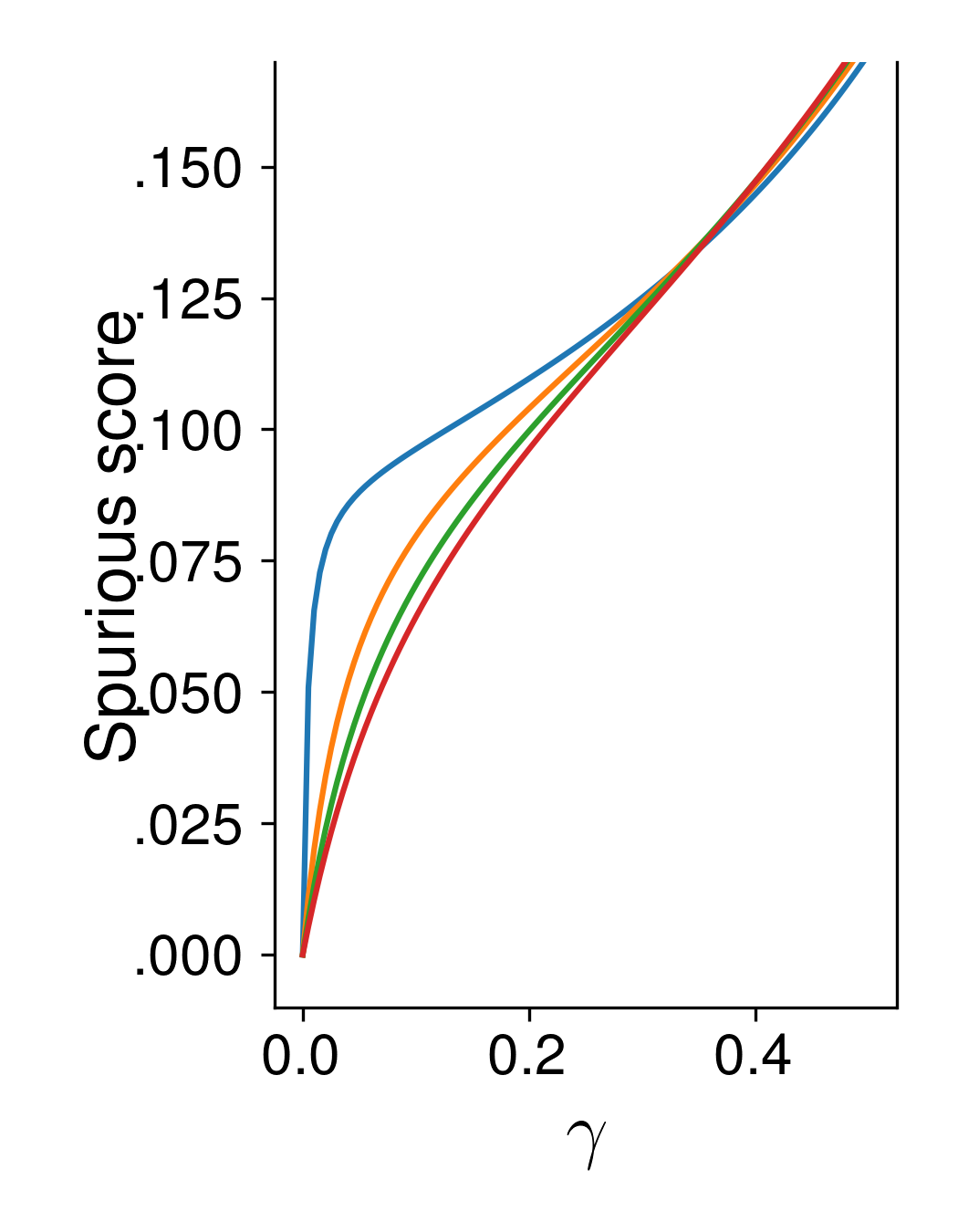}}
  
  \subfigure[ctest SNR $\downarrow$]{\includegraphics[width=0.16\textwidth]{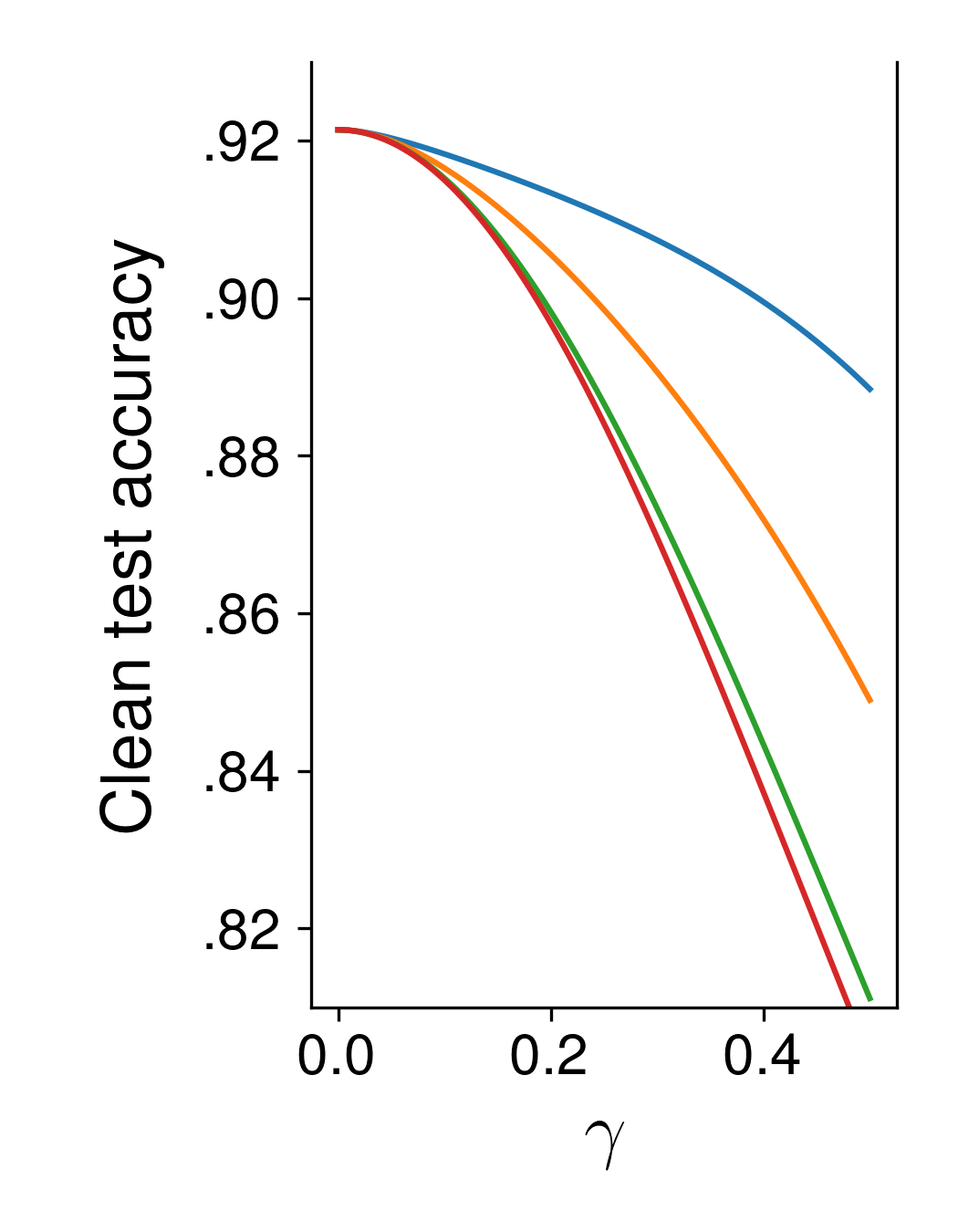}}
  \subfigure[stest SNR $\downarrow$]{\includegraphics[width=0.16\textwidth]{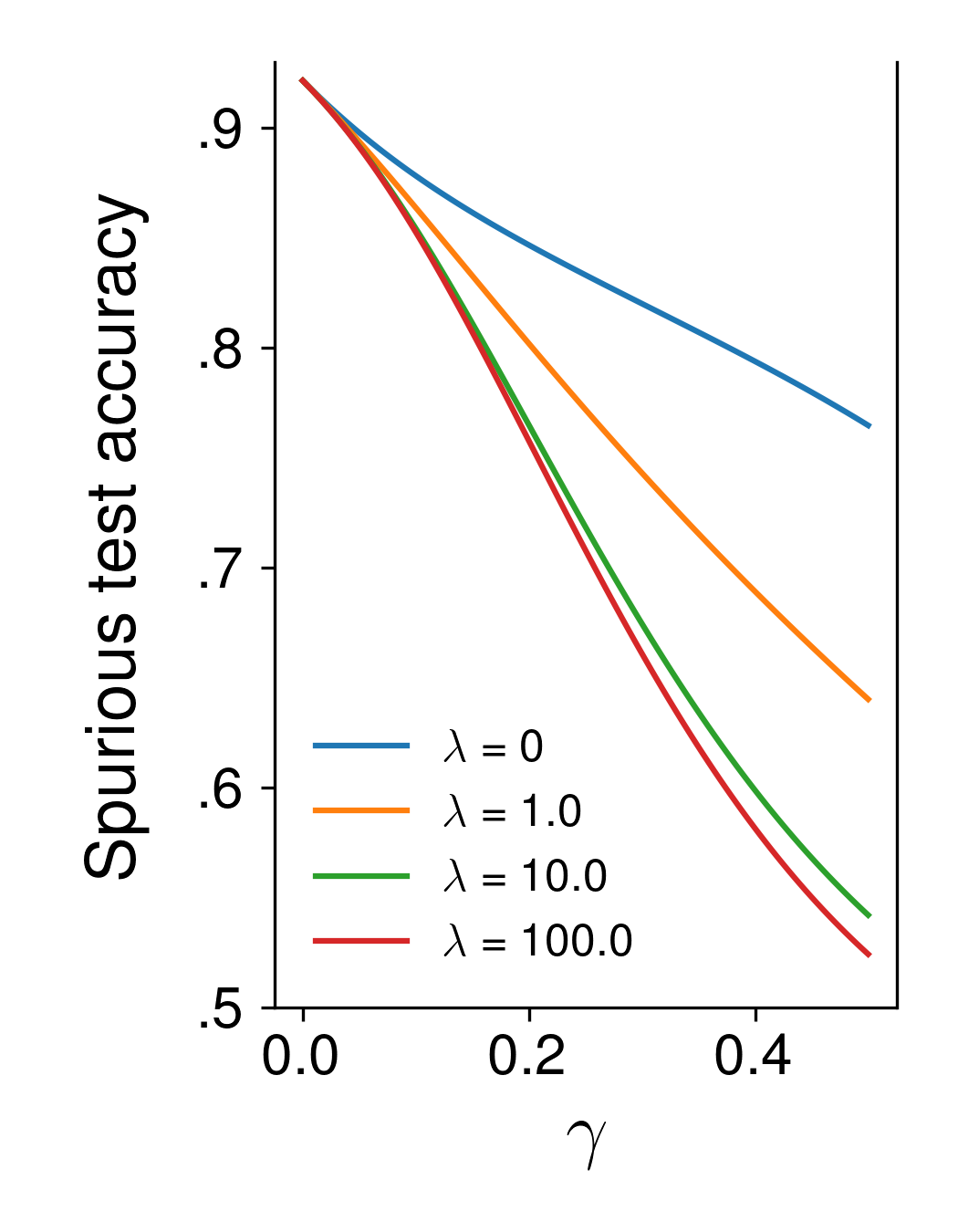}}
  \subfigure[sscore SNR $\downarrow$]{\includegraphics[width=0.16\textwidth]{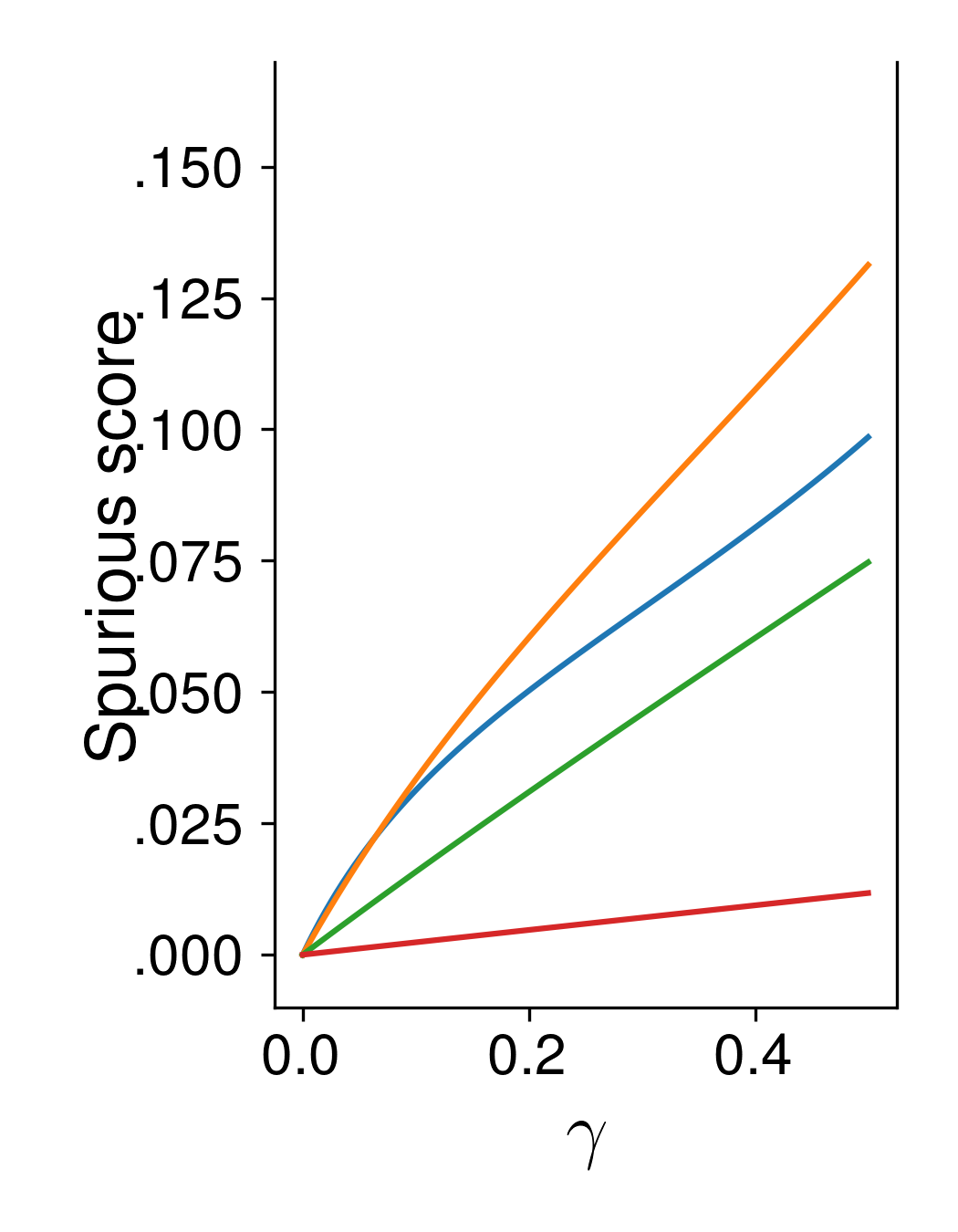}}
  \subfigure[ctest SNR $\uparrow$]{\includegraphics[width=0.16\textwidth]{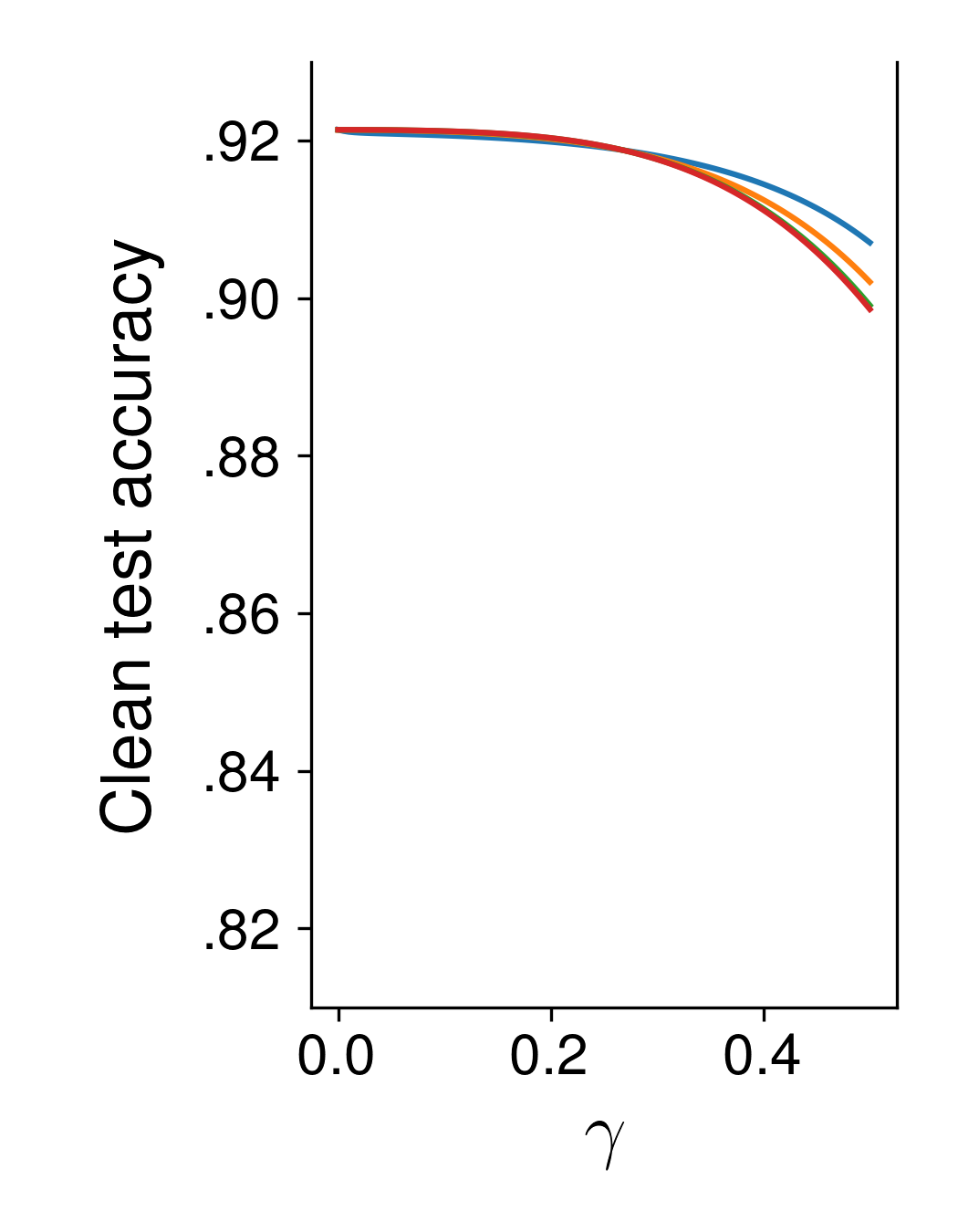}}
  \subfigure[stest SNR $\uparrow$]{\includegraphics[width=0.16\textwidth]{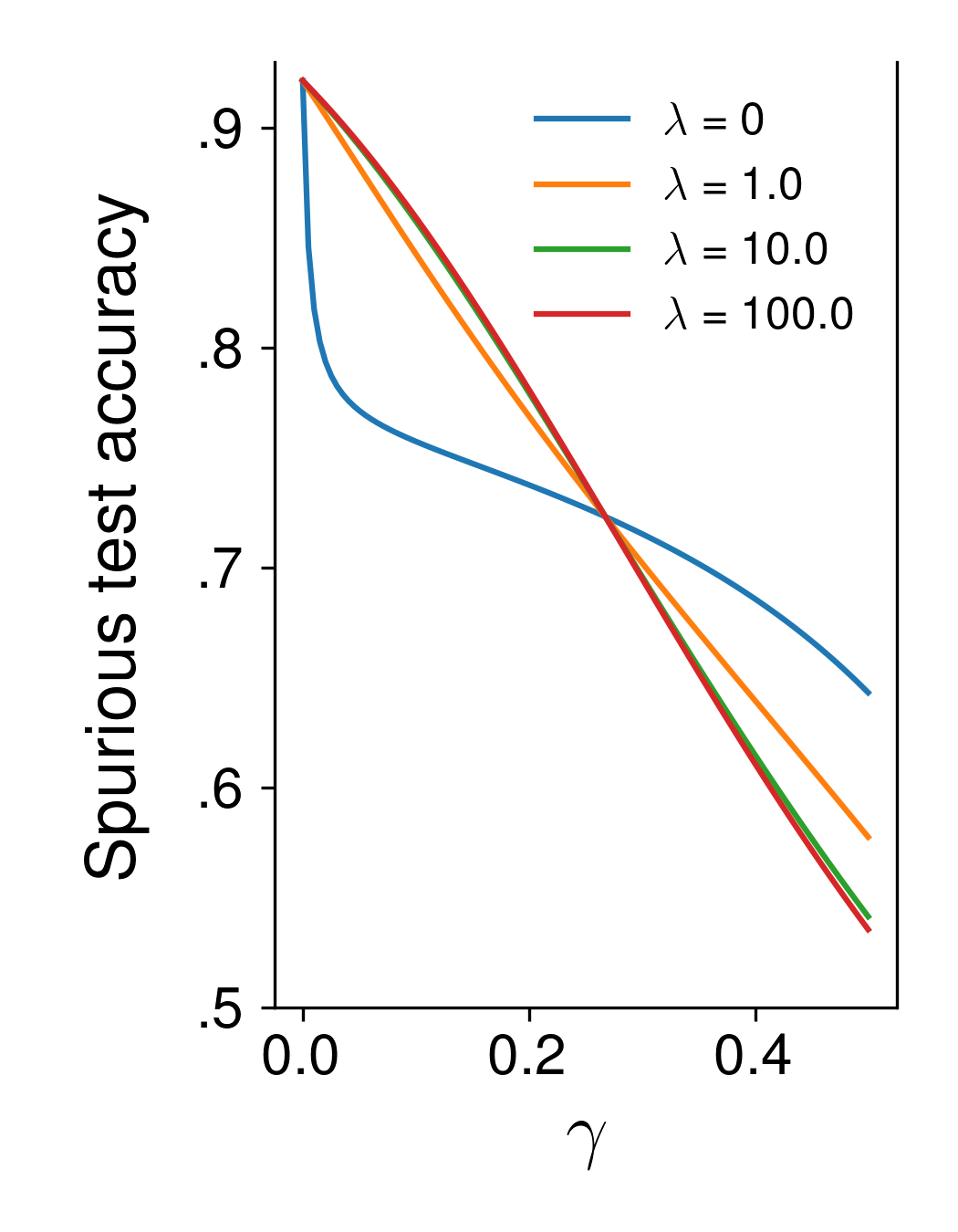}}
  \subfigure[sscore SNR $\uparrow$]{\includegraphics[width=0.16\textwidth]{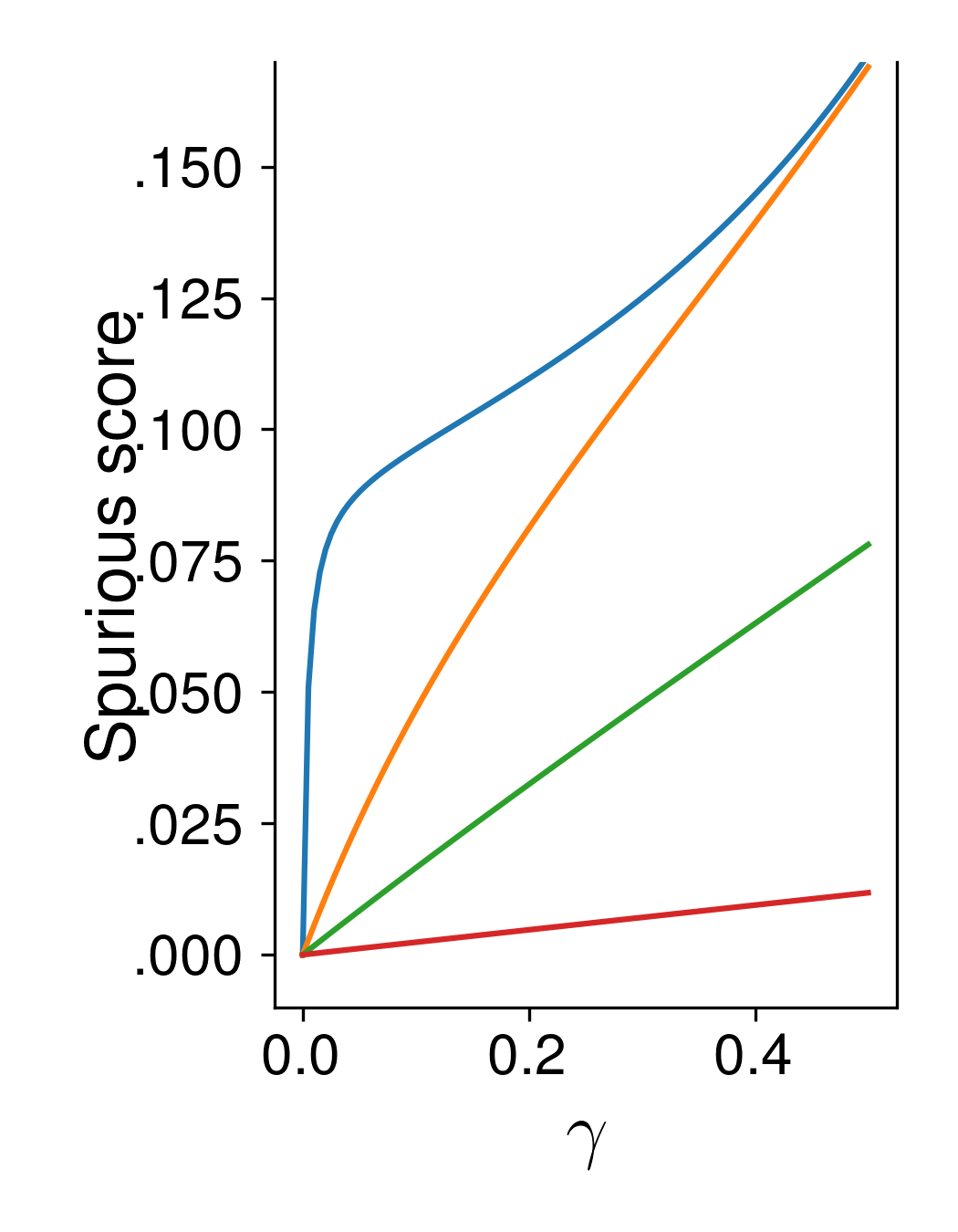}}
  
  \caption{
  Examples of the effect of adding Gaussian noises and $\ell_2$ regularization under different regime of signal-to-noise ratio (SNR) $\|\bxsp\|_2/\sigmasp^2$.
  For low SNR, we have $\|\bxsp\|_2/\sigmasp^2 = 10$, and for high SNR, we have $\|\bxsp\|_2/\sigmasp^2 = 1 = 500$.
  (a)-(f) show the effect of adding different strength of Gaussian noises and (g)-(l) show the effect of different $\ell_2$ regularization strength.
  Here, we shorthand clean test accuracy as ctest, spurious test accuracy as stest and spurious score as sscore.
  }
  \label{fig:theory effects}
\end{figure}

\mypara{Observation 2: adding Gaussian noises lowers spurious score.}
Note that adding (isotropic) Gaussian noises with strength $\sigma^2$ to input $\bx$ is equivalent to increase the variance by an additive factor of $\sigma^2$, i.e., $\sigmainv^2\gets\sigmainv^2+\sigma^2$ and $\sigmasp^2\gets\sigmasp^2+\sigma^2$. Empirically we observe in~\cref{fig:theory effects} that both the clean test accuracy and spurious test accuracy decay as $\sigma^2$ becomes larger while the prediction difference decays favorably. Intuitively, adding Gaussian noises decrease the signal-to-noise ratio of both the invariant feature and the spurious feature.

\mypara{Observation 3: $\ell_2$ regularization improves test accuracy and lowers spurious score.}
The main difference of $\ell_2$ regularization from adding Gaussian noises is that it non-uniformly increasing the effective variance of the invariant subspace and that of the spurious subspace. This can be seen from the formulas in~\cref{thm:loss} that the regularization parameter $\lambda$ is only added to some but not all of the $\sigmainv^2$ and $\sigmasp^2$. Numerically, we also plot the test accuracy and the prediction difference under different parameters to demonstrate the effect of $\ell_2$ regularization in~\cref{fig:theory effects}.

\subsection{Proof of Theorem~\ref{thm:loss}}\label{app:theory proof}
\begin{proof}
For notational simplicity, in this proof we will overload the notations $\bxinv=(\bxinv,0)\in\cX$, $\bxsp=(0,\bxsp)\in\cX$, $\bxinv=(\bxinv,0)\in\cX$ and $-\bxinv=(-\bxinv,0)\in\cX$. Also, we denote $\beta=(\betainv,\betasp)$.

Let us start with finding $h^*$ via explicitly calculating the derivatives of loss function under $\cDtrain$ as follows. 
\begin{claim}\label{claim:derivatives of loss}
For every $h=(\betainv,\betasp,\beta_0)\in\mathbb{H}$, we have
\begin{align*}
\frac{\partial}{\partial\beta}L_{\cDtrain}(h) &= (\sigmainv^2+\lambda)\betainv + \left(\bxinv^\top\betainv + \frac{\gamma\bxsp^\top\betasp}{2}-1\right)\bxinv\\
&+ (\sigmasp^2+\lambda)\betasp + \frac{\gamma}{2}\left(\bxinv^\top\betainv+\bxsp^\top\betasp+\beta_0-1\right)\bxsp\, ,\\
\frac{\partial}{\partial\beta_0}L_{\cDtrain}(h) &= \frac{\gamma\bxsp^\top\betasp}{2} + \beta_0 \, .
\end{align*}
\end{claim}
The proof of Claim~\ref{claim:derivatives of loss} follows from a direct calculation and we postpone the details to Section~\ref{sec:moment calculations}.

As the loss function is a quadratic function, it suffices to solve the equations $\frac{\partial}{\partial\beta}L_{\cDtrain}(h)=0$ and $\frac{\partial}{\partial\beta_0}L_{\cDtrain}(h)=0$. Note that since the invariant subspace $\cXinv$ is orthogonal to the spurious subspace $\cXsp$, this enforces $\betainv$ (resp. $\betasp$) to be a scalar multiplication of $\bxinv$ (resp. $\bxsp$). Let us set $\betainv=a\bxinv$ and $\betasp=b\bxsp$ and solve $a,b$ by setting the partial derivatives to be zero. Namely, we get the following equations.
\begin{align*}
0 &= a(\sigmainv^2+\lambda) + a\|\bxinv\|_2^2 + \frac{\gamma\|\bxsp\|_2^2}{2}b-1 \, ,\\
0 &= b(\sigmasp^2+\lambda) + \frac{\gamma}{2}(a\|\bxinv\|_2^2+b\|\bxsp\|_2^2+\beta_0-1) \, ,\\
0 &= \frac{\gamma\|\bxsp\|_2^2}{2}b + \beta_0 \, .
\end{align*}
By solving the above system of linear equations, we get
\begin{align}
b &= \frac{\frac{\gamma}{2}(\sigmainv^2+\lambda)}{(\sigmainv^2+\lambda)(\sigmasp^2+\lambda)+\left(\sigmasp^2+\lambda+\frac{\gamma}{2}(1-\frac{\gamma}{2})(\sigmainv^2+\lambda)+\frac{\gamma}{2}(1-\gamma)\|\bxsp\|_2^2\right)\|\bxsp\|_2^2} \, , \label{eq:app theory 1}\\
\frac{a}{b} &= \frac{\frac{\gamma}{2}(1-\gamma)\|\bxsp\|_2^2+\sigmasp^2+\lambda}{\frac{\gamma}{2}(\sigmainv^2+\lambda)}\, , \label{eq:app theory 2}\\
\frac{\beta_0}{b} &= -\frac{\gamma\|\bxsp\|_2^2}{2}\, .\label{eq:app theory 3}
\end{align}

Now that we know what $h^*$ is, we can calculate the clean test accuracy as follows.
\begin{align*}
ACC_{\cDctest}(h^*) &= \Pr_{(\bx,y)\sim\cDctest}[h^*(\bx)y>0]\\
&= \frac{1}{2}\Pr_{\bx\sim\cN((-\bxinv,0),\Sigma)}[h^*(\bx)<0] + \frac{1}{2}\Pr_{\bx\sim\cN((\bxinv,0),\Sigma)}[h^*(\bx)>0] \\
&=\frac{1}{2}\Pr_{\bw\sim\cN(0,\Sigma)}[\beta^\top(\bw-\bxinv)+\beta_0<0] + \frac{1}{2}\Pr_{\bw\sim\cN(0,\Sigma)}[\beta^\top(\bw+\bxinv)+\beta_0>0] \, .
\intertext{As $\cN(0,\Sigma)$ is isotropic in both $\cXinv$ and $\cXsp$, the above becomes}
&= \frac{1}{2}\Pr_{\bw\sim\cN(0,\Sigma)}[\beta^\top(\bw-\bxinv)<-\beta_0] + \frac{1}{2}\Pr_{\bw\sim\cN(0,\Sigma)}[\beta^\top(\bw-\bxinv)<\beta_0] \\
&= \frac{1}{2}\Pr_{\substack{\bwinv\sim\cN(0,\sigmainv^2 I_\text{inv})\\\bwsp\sim\cN(0,\sigmasp^2 I_{\text{sp}})}}[a\bxinv^\top\bwinv+b\bxsp^\top\bwsp < a\|\bxinv\|_2^2-\beta_0]\\ 
&+ \frac{1}{2}\Pr_{\substack{\bwinv\sim\cN(0,\sigmainv^2 I_\text{inv})\\\bwsp\sim\cN(0,\sigmasp^2 I_{\text{sp}})}}[a\bxinv^\top\bwinv+b\bxsp^\top\bwsp < a\|\bxinv\|_2^2+\beta_0] \, .
\intertext{Note that $a\bxinv^\top\bwinv+b\bxsp^\top\bwsp$ follows the distribution $\cN(0,\sigma^2)$ where $\sigma^2=a^2\|\bxinv\|_2^4\sigmainv^2+b^2\|\bxsp\|_2^4\sigmasp^2$. Namely, the above can be further simplified as}
&= \frac{1}{2}\Pr_{w\sim \cN(0,\sigma^2)}[w<a\|\bxinv\|_2^2+\beta_0] + \frac{1}{2}\Pr_{w\sim \cN(0,\sigma^2)}[w<a\|\bxinv\|_2^2-\beta_0]\\
&= \frac{1}{2}\Phi\left(\frac{a\|\bxinv\|_2^2-\beta_0}{\sqrt{a^2\|\bxinv\|_2^4\sigmainv^2+b^2\|\bxsp\|_2^4\sigmasp^2}}\right) + \frac{1}{2}\Phi\left(\frac{a\|\bxinv\|_2^2+\beta_0}{\sqrt{a^2\|\bxinv\|_2^4\sigmainv^2+b^2\|\bxsp\|_2^4\sigmasp^2}}\right) \, .
\intertext{Lastly, by plugging in~\cref{eq:app theory 1,eq:app theory 2,eq:app theory 3}, the clean test accuracy of $h^*$ is}
&= \frac{1}{2}\Phi\left(\frac{\frac{\frac{\gamma}{2}(1-\gamma)\|\bxsp\|_2^2+\sigmasp^2+\lambda}{\sigmainv^2+\lambda}\|\bxinv\|_2^2+(\frac{\gamma}{2})^2\|\bxsp\|_2^2}{\sqrt{\left(\frac{\frac{\gamma}{2}(1-\gamma)\|\bxsp\|_2^2+\sigmasp^2+\lambda}{\sigmainv^2+\lambda}\right)^2\|\bxinv\|_2^4\sigmainv^2+(\frac{\gamma}{2})^2\|\bxsp\|_2^4\sigmasp^2}}\right)\\
&+\frac{1}{2}\Phi\left(\frac{\frac{\frac{\gamma}{2}(1-\gamma)\|\bxsp\|_2^2+\sigmasp^2+\lambda}{\sigmainv^2+\lambda}\|\bxinv\|_2^2-(\frac{\gamma}{2})^2\|\bxsp\|_2^2}{\sqrt{\left(\frac{\frac{\gamma}{2}(1-\gamma)\|\bxsp\|_2^2+\sigmasp^2+\lambda}{\sigmainv^2+\lambda}\right)^2\|\bxinv\|_2^4\sigmainv^2+(\frac{\gamma}{2})^2\|\bxsp\|_2^4\sigmasp^2}}\right) \, .
\intertext{Similarly, the spurious test accuracy of $h^*$ is}
Acc_{\cDstest}(h^*) &= \Pr_{(\bx,y)\sim\cDstest}[h^*(\bx)y>0]\\
&= \Pr_{\bx\sim\cN((-\bxinv,\bxsp),\Sigma)}[h^*(\bx)<0]\\
&= \Pr_{\substack{\bwinv\sim\cN(0,\sigmainv I_\text{inv})\\\bwsp\sim\cN(0,\sigmasp I_{\text{sp}})}}[a\bxinv^\top\bwinv+b\bxsp^\top\bwsp<a\|\bxinv\|_2^2-b\|\bxsp\|_2^2-\beta_0]\\
&= \Pr_{w\sim \cN(0,a^2\|\bxinv\|_2^4\sigmainv^2+b^2\|\bxsp\|_2^4\sigmasp^2)}[w<a\|\bxinv\|_2^2-b\|\bxsp\|_2^2-\beta_0]\\
&= \Phi\left(\frac{a\|\bxinv\|_2^2-b\|\bxsp\|_2^2-\beta_0}{\sqrt{a^2\|\bxinv\|_2^4\sigmainv^2+b^2\|\bxsp\|_2^4\sigmasp^2}}\right)\\
&= \Phi\left(\frac{\frac{\frac{\gamma}{2}(1-\gamma)\|\bxsp\|_2^2+\sigmasp^2+\lambda}{\sigmainv^2+\lambda}\|\bxinv\|_2^2+(\frac{\gamma}{2}-1)\frac{\gamma}{2}\|\bxsp\|_2^2}{\sqrt{\left(\frac{\frac{\gamma}{2}(1-\gamma)\|\bxsp\|_2^2+\sigmasp^2+\lambda}{\sigmainv^2+\lambda}\right)^2\|\bxinv\|_2^4\sigmainv^2+(\frac{\gamma}{2})^2\|\bxsp\|_2^4\sigmasp^2}}\right) \, .\\
\end{align*}
Finally, as $h^*$ is a linear classifier, its prediction difference is
\begin{align*}
PD(h^*) &= h^*(\bxsp) = b\|\bxsp\|_2^2\\
&= \frac{\frac{\gamma}{2}(\sigmainv^2+\lambda)\|\bxsp\|_2^2}{(\sigmainv^2+\lambda)(\sigmasp^2+\lambda)+\left(\sigmasp^2+\lambda+\frac{\gamma}{2}(1-\frac{\gamma}{2})(\sigmainv^2+\lambda)+\frac{\gamma}{2}(1-\gamma)\|\bxsp\|_2^2\right)\|\bxsp\|_2^2}
\end{align*}
where the last equality follows~\cref{eq:app theory 1}. This completes the proof of~\cref{thm:loss}.
\end{proof}

\subsubsection{Moment calculations}\label{sec:moment calculations}
\begin{proof}[Proof of Claim~\ref{claim:derivatives of loss}]
Recall that we denote $\beta=(\betainv,\betasp)$ for convenience.
For every $h=(\betainv,\betasp,\beta_0)\in\mathbb{H}$, we have
\begin{align*}
L_{\cDtrain}(h) &= \frac{1}{2}\Exp_{(\bx,y)\sim\cDtrain}[(h(\bx)\cdot y-1)^2] + \frac{\lambda}{2}\|\beta\|_2^2\\
&= \frac{1}{2}\Exp_{(\bx,y)\sim\cDtrain}[y^2(\beta^\top\bx+\beta_0)^2-2y(\beta^\top\bx+\beta_0)+1] + \frac{\lambda}{2}\|\beta\|_2^2  \, .
\intertext{Note that $y^2=1$ almost surely and we can calculate the partial derivatives of $L_{\cDtrain}(h)$ as follows.}
\frac{\partial}{\partial\beta}L_{\cDtrain}(h) &= \Exp_{(\bx,y)\sim\cDtrain}[\bx(\bx^\top\beta+\beta_0-y)]+\lambda\beta \, ,\\
\frac{\partial}{\partial\beta_0}L_{\cDtrain}(h) &= \Exp_{(\bx,y)\sim\cDtrain}[\bx^\top\beta+\beta_0-y] \, .
\end{align*}
Let's start with calculating the first moment terms. Recall that we overload the notation of $\bxinv$ and $\bxsp$ as explained in the beginning of the proof of Theorem~\ref{thm:loss}.
\begin{align*}
\Exp_{(\bx,y)\sim\cDtrain}[\bx] &= \frac{1}{2}\Exp_{\bx\sim\cN((-\bxinv,0),\Sigma)}[\bx]+ \frac{1-\gamma}{2}\Exp_{\bx\sim\cN((\bxinv,0),\Sigma)}[\bx]+ \frac{\gamma}{2}\Exp_{\bx\sim\cN((\bxinv,\bxsp),\Sigma)}[\bx]\\
&= \frac{-\bxinv+\bxinv + \gamma\bxsp}{2} \, ,\\
\Exp_{(\bx,y)\sim\cDtrain}[y\bx] &= \frac{1}{2}\Exp_{\bx\sim\cN((-\bxinv,0),\Sigma)}[y\bx]+ \frac{1-\gamma}{2}\Exp_{\bx\sim\cN((\bxinv,0),\Sigma)}[y\bx]+ \frac{\gamma}{2}\Exp_{\bx\sim\cN((\bxinv,\bxsp),\Sigma)}[y\bx]\\
&= \frac{\bxinv+\bxinv+\gamma\bxsp}{2} \, ,\\
\Exp_{(\bx,y)\sim\cDtrain}[y] &= \frac{1}{2}\Exp_{\bx\sim\cN((-\bxinv,0),\Sigma)}[y]+ \frac{1-\gamma}{2}\Exp_{\bx\sim\cN((\bxinv,0),\Sigma)}[y]+ \frac{\gamma}{2}\Exp_{\bx\sim\cN((\bxinv,\bxsp),\Sigma)}[y]\\ 
&= 0 \, .
\end{align*}
Next, let's calculate the second moment terms.
\begin{align*}
\Exp_{(\bx,y)\sim\cDtrain}[\bx\bx^\top] &= \frac{1}{2}\Exp_{\bx\sim\cN((-\bxinv,0),\Sigma)}[\bx\bx^\top] + \frac{1-\gamma}{2}\Exp_{\bx\sim\cN((\bxinv,0),\Sigma)}[\bx\bx^\top] + \frac{\gamma}{2}\Exp_{\bx\sim\cN((\bxinv,\bxsp),\Sigma)}[\bx\bx^\top]\\
&= \frac{1}{2}(\Sigma+(\bxinv\bxinv^\top)) + \frac{1-\gamma}{2}(\Sigma+(\bxinv\bxinv^\top)) + \frac{\gamma}{2}(\Sigma+((\bxinv+\bxsp)(\bxinv+\bxsp)^\top))\\
&= \sigmainv^2 I_\text{inv} + \sigmasp^2 I_\text{sp} +  \frac{\bxinv\bxinv^\top+\bxinv\bxinv^\top + \gamma(\bxsp\bxinv^\top+\bxinv\bxsp^\top+\bxsp\bxsp^\top)}{2} \, .
\end{align*}
Finally, putting everything together we have
\begin{align*}
\frac{\partial}{\partial\beta}L_{\cDtrain}(h) &= \Exp_{(\bx,y)\sim\cDtrain}[\bx(\bx^\top\beta+\beta_0-y)]+\lambda\beta \\
&= \sigmainv^2\betainv+\sigmasp^2\betasp+\frac{\bxinv^\top\betainv}{2}\bxinv+\frac{\bxinv^\top\betainv+\gamma\bxsp^\top\betasp}{2}\bxinv + \frac{\gamma\bxinv^\top\betainv+\gamma\bxsp^\top\betasp}{2}\bxsp\\
&+ \beta_0\cdot\frac{-\bxinv+\bxinv+\gamma\bxsp}{2} - \frac{\bxinv+\bxinv+\gamma\bxsp}{2} + \lambda\betainv + \lambda\betasp\\
&= (\sigmainv^2+\lambda)\betainv + \left(\bxinv^\top\betainv + \frac{\gamma\bxsp^\top\betasp}{2}-1\right)\bxinv\\
&+ (\sigmasp^2+\lambda)\betasp + \frac{\gamma}{2}\left(\bxinv^\top\betainv+\bxsp^\top\betasp+\beta_0-1\right)\bxsp \, .
\intertext{Similarly,}
\frac{\partial}{\partial\beta_0}L_{\cDtrain}(h) &= \Exp_{(\bx,y)\sim\cDtrain}[\bx^\top\beta+\beta_0-y] = \frac{\gamma\bxsp^\top\betasp}{2} + \beta_0 \, .
\end{align*}
This completes the proof of Claim~\ref{claim:derivatives of loss}.
\end{proof}

\section{How Different Factors Affect Rare Spurious Correlations}
\label{app:different-factors}

We next investigate how different factors can affect the extent to which rare spurious correlations can be learnt. For this purpose, we consider different regularization methods, the norm of each pattern, network architectures, and the optimization algorithms.

\subsection{Different regularization methods}\label{app:theoratical_model_reg}

\begin{figure}[ht]
  \centering
  \subfigure[clean test]{\includegraphics[width=0.16\textwidth]{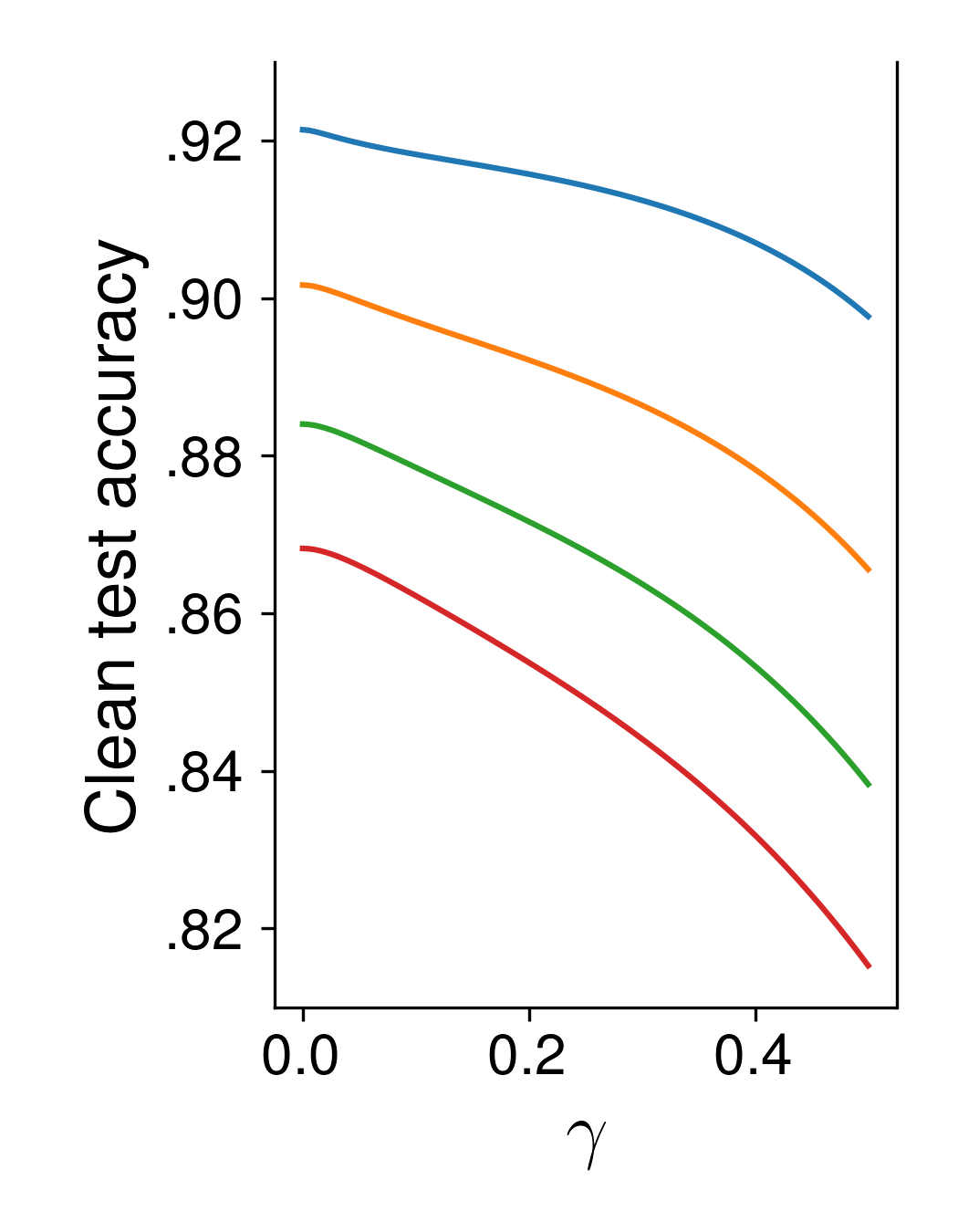}}
  \subfigure[spurious test]{\includegraphics[width=0.16\textwidth]{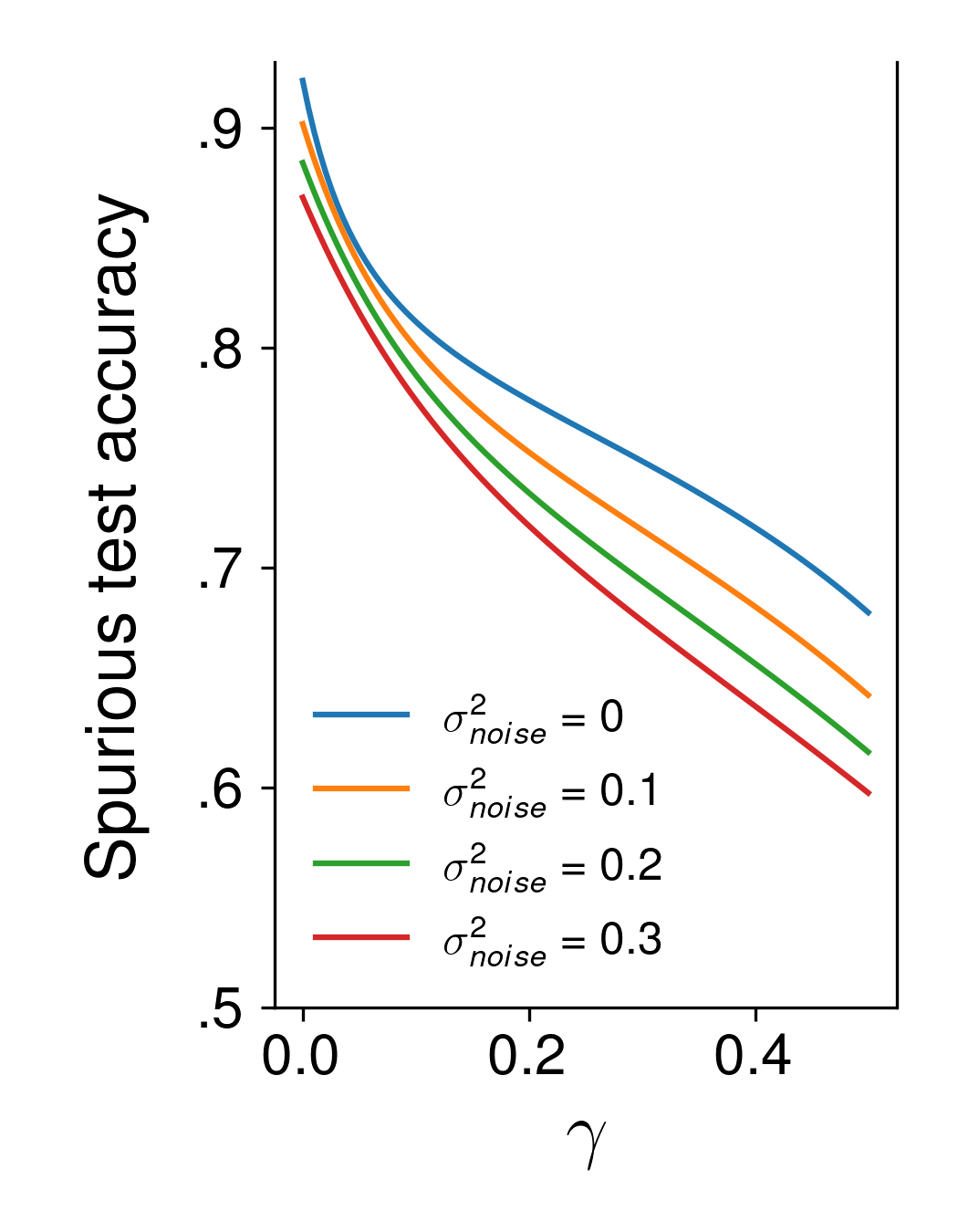}}
  \subfigure[spurious score]{\includegraphics[width=0.16\textwidth]{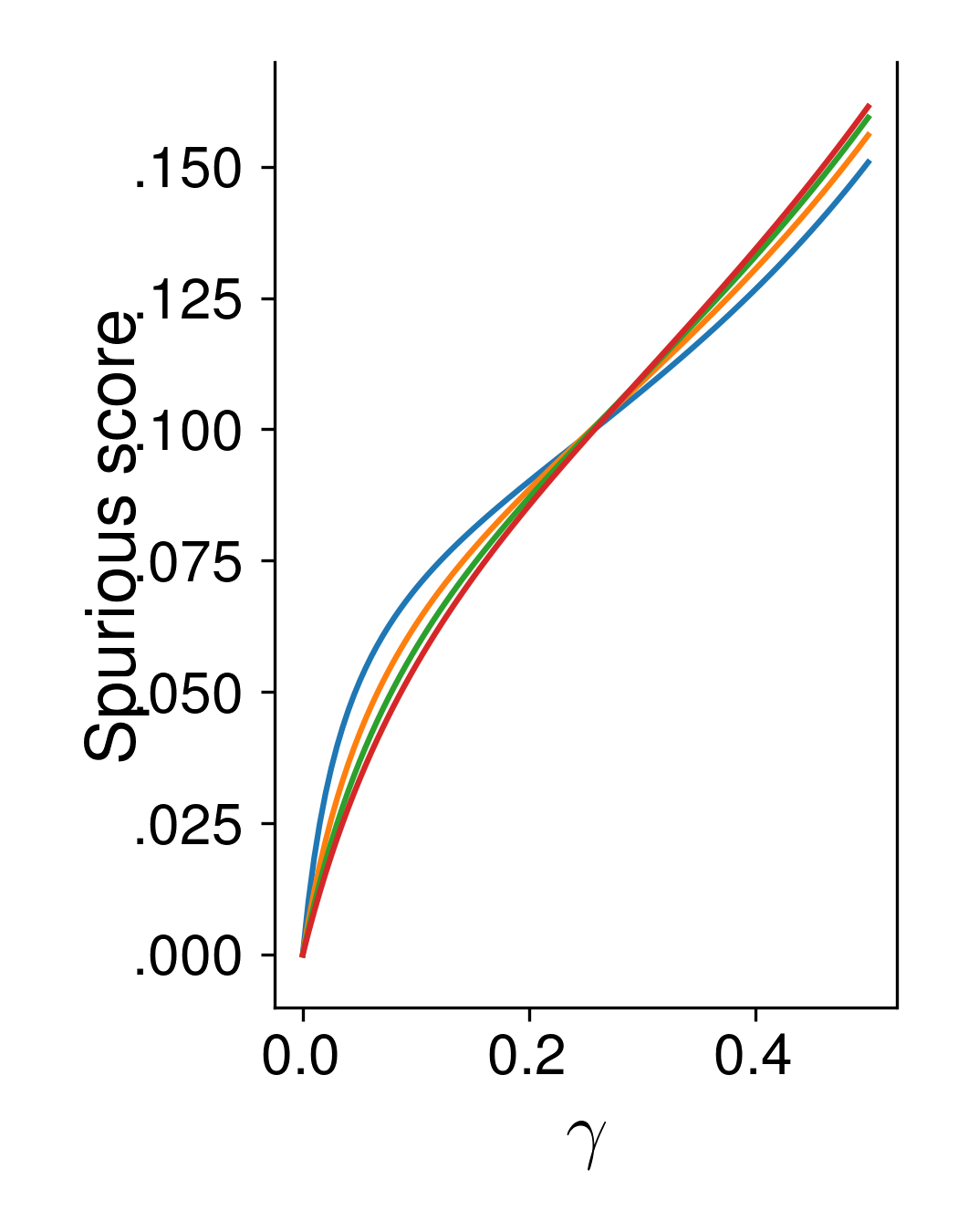}}
  \hspace{.2em}
  \subfigure[clean test]{\includegraphics[width=0.16\textwidth]{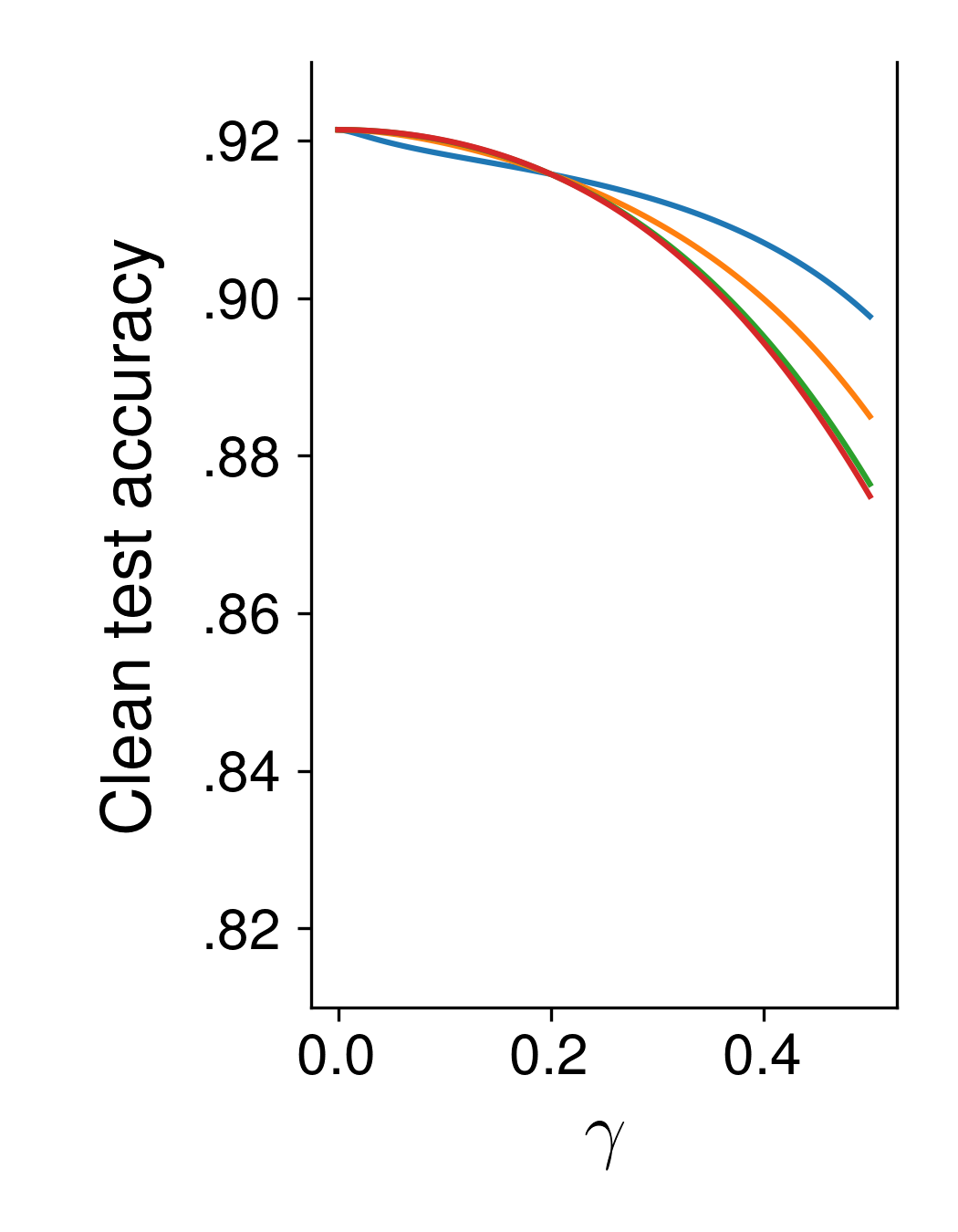}}
  \subfigure[spurious test]{\includegraphics[width=0.16\textwidth]{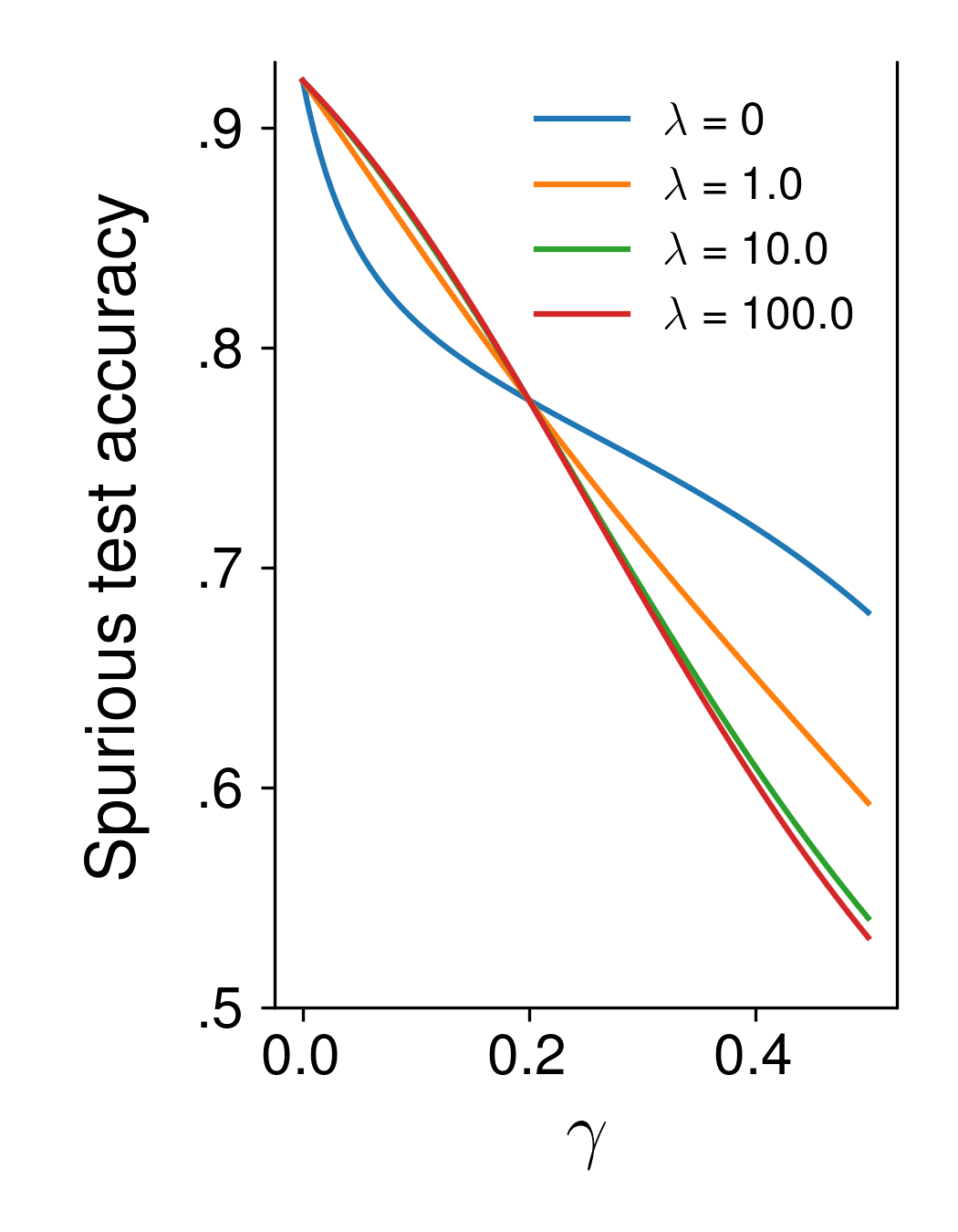}}
  \subfigure[spurious score]{\includegraphics[width=0.16\textwidth]{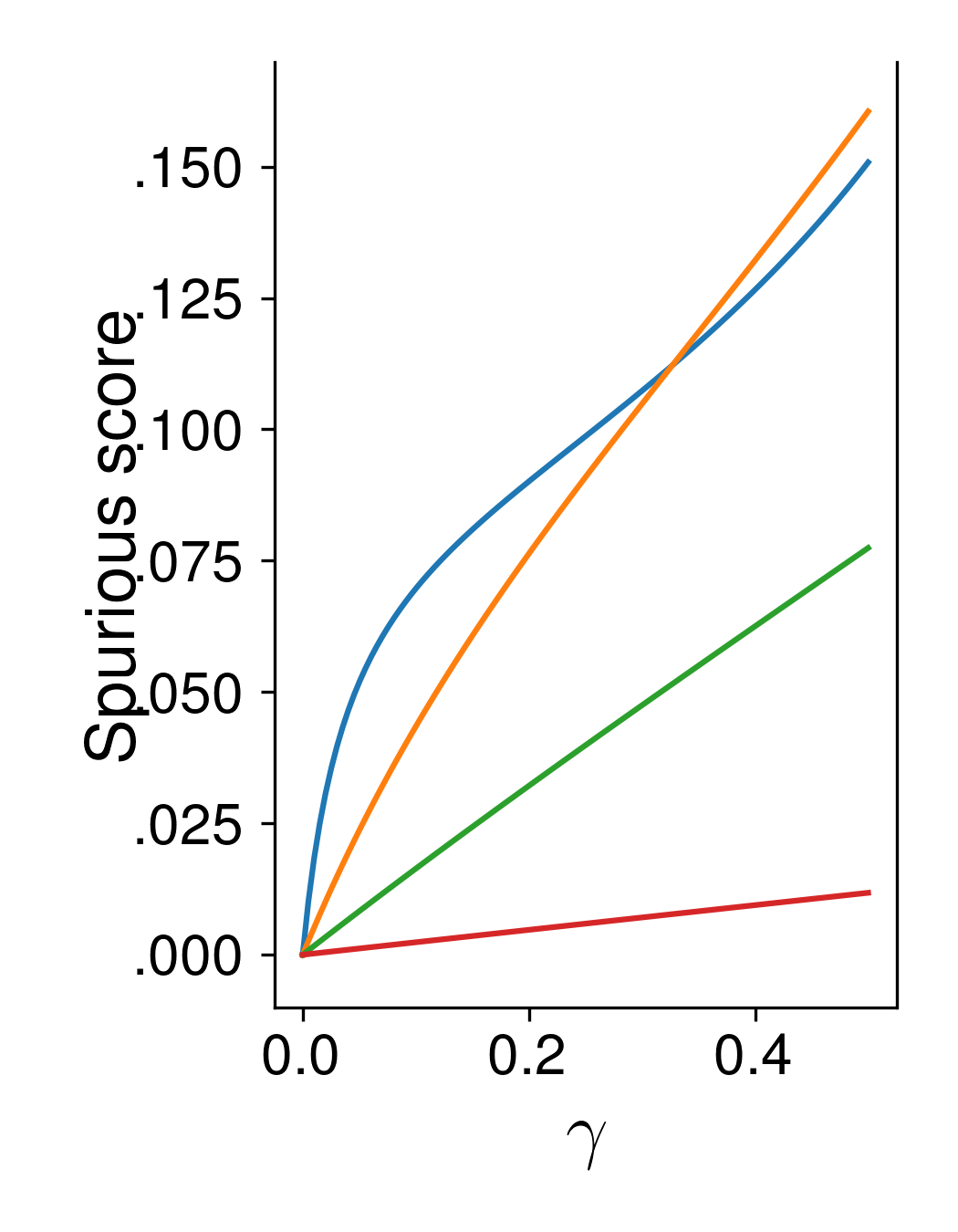}}

  \caption{
  An example of our theoretical findings. (a)-(c): adding Gaussian noises. (d)-(f): $\ell_2$ regularization. The parameters are set to be $\|\bxsp\|_2^2=5$, $\sigmainv^2=0.5$, $\sigmasp^2=0.1$, and we consider $\gamma\in[0,0.5]$,  noise strength $\sigma^2_{noise}=0, 0.1, 0.2, 0.3$, and the $\ell_2$ regularization parameter
  $\lambda=0, 1, 10, 100$.
  }
  \label{fig:theory}
\end{figure}

\begin{figure}[h!]
  \centering
  \subfigure[MNIST, noisy input]{\includegraphics[width=0.31\textwidth]{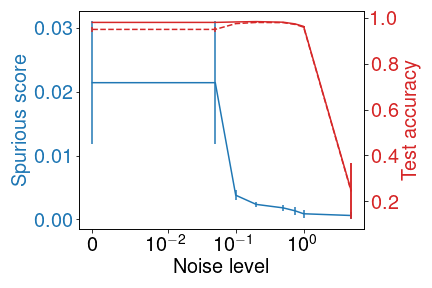}}
  \subfigure[Fashion, noisy input]{\includegraphics[width=0.31\textwidth]{figs/regularization/fashion_v8_noise.png}}
  \subfigure[CIFAR10, noisy input]{\includegraphics[width=0.31\textwidth]{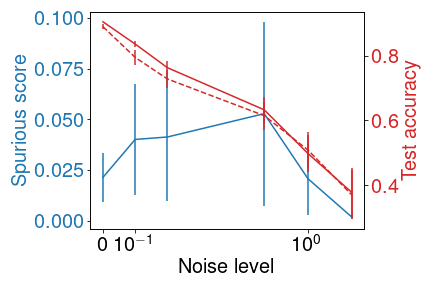}}
  
  \subfigure[MNIST, weight decay]{\includegraphics[width=0.31\textwidth]{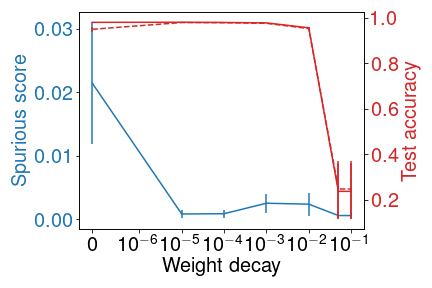}}
  \subfigure[Fashion, weight decay]{\includegraphics[width=0.31\textwidth]{figs/regularization/fashion_v8_wd.png}}
  \subfigure[CIFAR10, weight decay]{\includegraphics[width=0.31\textwidth]{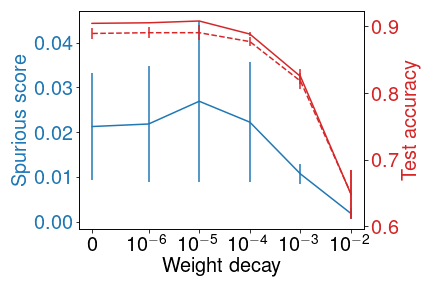}}
  
  \subfigure[MNIST, gradient clipping]{\includegraphics[width=0.31\textwidth]{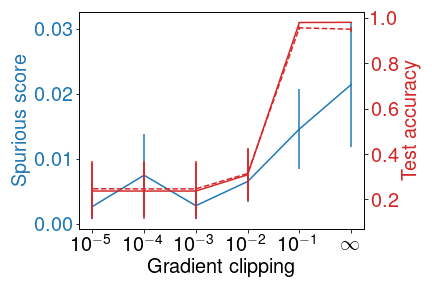}}
  \subfigure[Fashion, gradient clipping]{\includegraphics[width=0.31\textwidth]{figs/regularization/fashion_v8_gc.png}}
  \subfigure[CIFAR10, gradient clipping]{\includegraphics[width=0.31\textwidth]{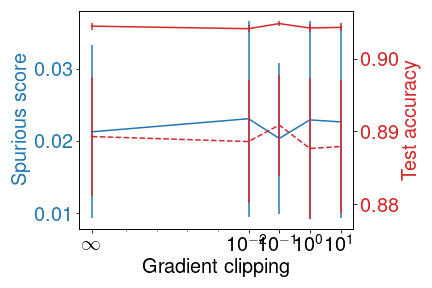}}
  
  \caption{
  Spurious score (solid blue line), clean test accuracy (solid red line), and spurious test accuracy (dotted red line) vs. the regularization strength on MNIST, Fashion, and CIFAR10 with different regularization methods.
  For the experiment, we fix the spurious pattern to be \isc\ and the target class $c_{tar}=0$.
  We compute the average spurious score and clean test accuracy across models trained with 1, 3, 5, 10, 20, and 100 spurious examples and five random seeds.
  }
  \label{fig:additional_reg_plot}
\end{figure}

\begin{figure}[h!]
  \centering
  \subfigure[MNIST, noisy input]{\includegraphics[width=0.31\textwidth]{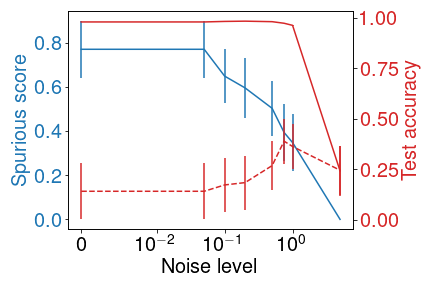}}
  \subfigure[Fashion, noisy input]{\includegraphics[width=0.31\textwidth]{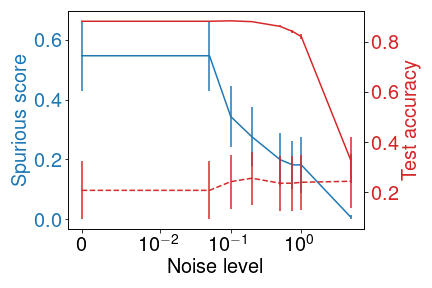}}
  \subfigure[CIFAR10, noisy input]{\includegraphics[width=0.31\textwidth]{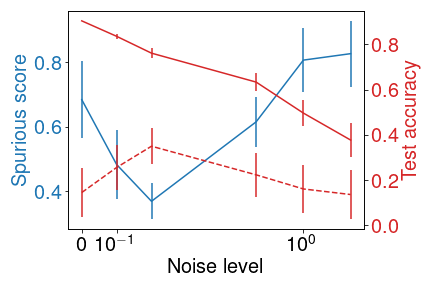}}
  
  \subfigure[MNIST, weight decay]{\includegraphics[width=0.31\textwidth]{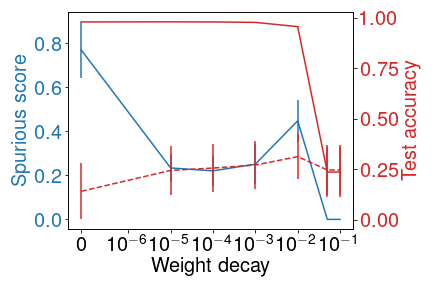}}
  \subfigure[Fashion, weight decay]{\includegraphics[width=0.31\textwidth]{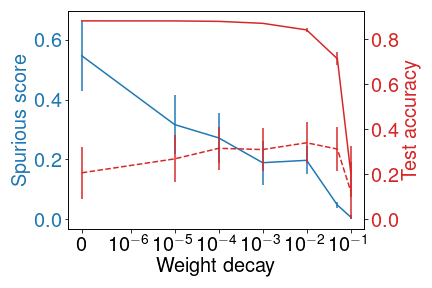}}
  \subfigure[CIFAR10, weight decay]{\includegraphics[width=0.31\textwidth]{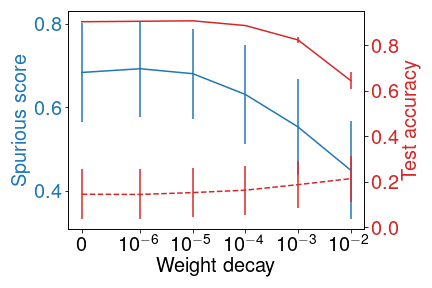}}
  
  \subfigure[MNIST, gradient clipping]{\includegraphics[width=0.31\textwidth]{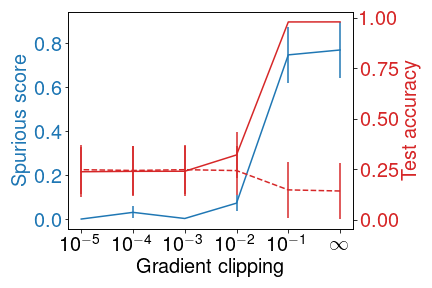}}
  \subfigure[Fashion, gradient clipping]{\includegraphics[width=0.31\textwidth]{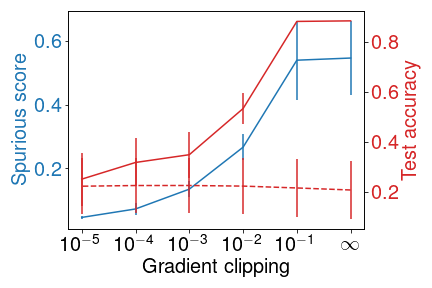}}
  \subfigure[CIFAR10, gradient clipping]{\includegraphics[width=0.31\textwidth]{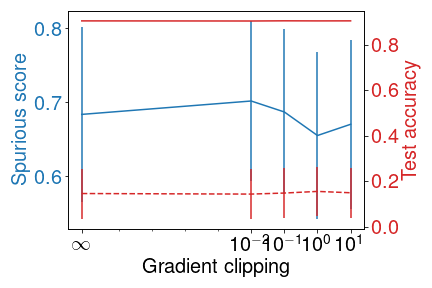}}
  
  \caption{
  Spurious score (solid blue line), clean test accuracy (solid red line), and spurious test accuracy (dotted red line) vs. the regularization strength on MNIST, Fashion, and CIFAR10 with different regularization methods.
  For the experiment, we fix the spurious pattern to be \irc\ and the target class $c_{tar}=0$.
  We compute the average spurious score and clean test accuracy across models trained with 1, 3, 5, 10, 20, and 100 spurious examples and five random seeds.
  }
  \label{fig:additional_reg_plot_v2}
\end{figure}

\cref{fig:additional_reg_plot} shows how different regularization level affects the spurious score, spurious test accuracy, and clean test accuracy with \isc\ as the spurious pattern.
\cref{fig:additional_reg_plot_v2} shows same thing but with \irc\ as the spurious pattern.
\cref{fig:theory} visualizes how different regularization strength can affect the spurious correlation learnt on the theoretical model.
We see similar trends across these three cases.

\subsection{\texorpdfstring{$\ell_2$}{l2} norm of the spurious pattern}
In \cref{fig:score_plot}, we see that neural networks can learn different patterns very differently.
For example, the spurious scores for \irc\ are higher than \isa, suggesting spurious correlations with \irc\ are learnt more easily.
Why does this happen?
We hypothesize that the higher the norm of the pattern is, easier it is for a network to learn the correlation between the patterns and the target class.

Because the spurious patterns may overlap with other features, directly using the norm of each spurious pattern may not be accurate.
We define the \textit{empirical norm} of a spurious pattern $\x_{sp}$ on an example $\x$ as the $\ell_2$ distance between $\x$ and the spurious example $g(\x, \x_{sp})$.
We compute the average empirical norm over the test examples for each pattern.
\cref{tab:pattern_norm} shows the average empirical norm of each pattern on different datasets.

For each dataset, we train neural networks with a different number of spurious examples.
To measure the aggregated effect of a spurious pattern across different numbers of spurious examples, we compute the average spurious scores across different numbers of spurious examples.
We compute the Pearson correlation between the average empirical norm and the average spurious scores of each model trained with different spurious patterns.
The testing results are:
MNIST: $\rho=0.91$, $p<0.01$;
Fashion: $\rho=0.84$, $p<0.02$;
CIFAR10: $\rho=0.98$, $p<0.01$.
The result shows a \textit{significantly strong positive correlation between the norm of the spurious patterns and the spurious scores}.

\begin{table}[ht]
  \centering
  \small
  \caption{The average empirical norm of each spurious pattern.}
  \begin{tabular}[]{lccccccc}
    \toprule
          & \isa & \isb & \isc & \ira & \irb & \irc & \ic  \\
    \midrule
  MNIST   & 1.00 & 3.00 & 5.00 & 3.88 & 7.70 & 15.19 & 5.98 \\
  Fashion & 1.00 & 3.00 & 4.98 & 3.90 & 7.40 & 13.65 & 4.44 \\
  CIFAR10 & 0.84 & 2.57 & 4.34 & 7.72 & 14.54 & 23.20 & 8.00 \\
    \bottomrule
  \end{tabular}
  \label{tab:pattern_norm}
\end{table}

\subsection{Network architectures}

Are some network architectures more susceptible to spurious correlations than others?
To answer this question, we look at how spurious scores vary across different network architectures.

\mypara{Network architectures.}
For MNIST and Fashion, we consider multi-layer perceptrons~(MLP) with different sizes and a convolutional neural network~(CNN)\footnote{public repository: \url{https://github.com/yaodongyu/TRADES/blob/master/models/}}.
The \textit{small MLP} has one hidden layer with $256$ neurons.
The \textit{MLP} has two hidden layers, each layer with $256$ neurons (the same MLP used in \cref{fig:score_plot}).
The \textit{large MLP} has two hidden layers, each with $512$ neurons.
For CIFAR10, we consider ResNet20, ResNet34, ResNet110~\citep{he2016deep}, and Vgg16~\cite{simonyan2014very}.
We use an SGD optimizer for CIFAR10 since we cannot get reasonable performance for Vgg16 with Adam.

\mypara{Setup.}
For MNIST and Fashion, we set the learning rate to $0.01$ for all architectures and optimizers.
We set the momentum to $0.9$ when the optimizer is SGD.
For CIFAR10, when training with SGD, we set the learning rate to $0.1$ for ResNets trained and $0.01$ for Vgg16 because Vgg16 failed to converge with learning rate $0.1$.
We set the learning rate to $0.01$ for ResNets when running with Adam (Vgg16 failed to converge with Adam).
We use a learning rate scheduler which decreases the learning rate by a factor of $0.1$ on the $40$-th, $50$-th, and $60$-th epoch.

\begin{figure}[ht]
  \centering

  \subfigure[MNIST, \isc]{\includegraphics[width=0.31\textwidth]{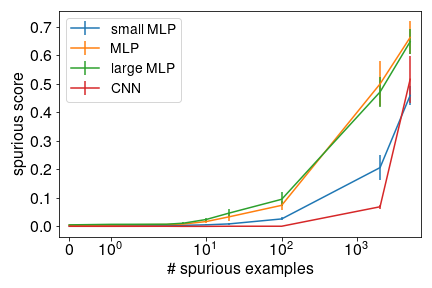}}
  \subfigure[Fashion, \isc]{\includegraphics[width=0.31\textwidth]{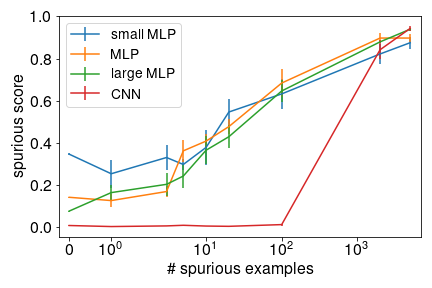}}
  \subfigure[CIFAR10, \isc]{\includegraphics[width=0.31\textwidth]{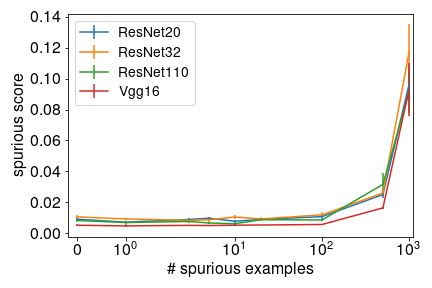}}

  \subfigure[MNIST, \irc]{\includegraphics[width=0.31\textwidth]{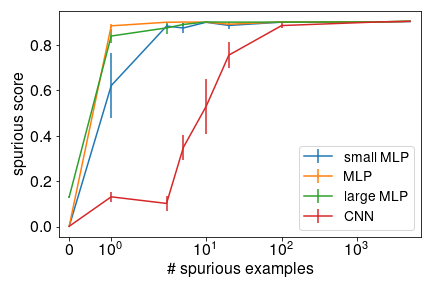}}
  \subfigure[Fashion, \irc]{\includegraphics[width=0.31\textwidth]{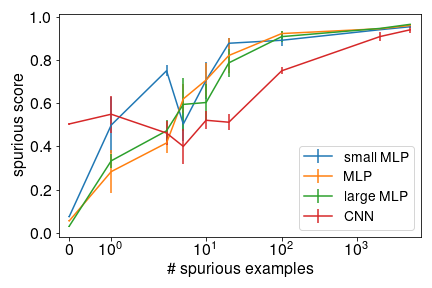}}
  \subfigure[CIFAR10, \irc]{\includegraphics[width=0.31\textwidth]{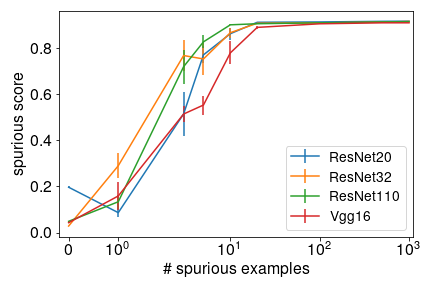}}

  \caption{
    The mean and standard error of the spurious scores with different network architectures on MNIST, Fashion, and CIFAR10.
    The target class is $c_{tar}=0$.
    }
  \label{fig:difarchs_score_plot}
\end{figure}

\cref{fig:difarchs_score_plot} shows the result, and we see that similar architectures with different sizes generally have similar spurious scores. Concretely, small MLP, MLP, and large MLP perform similarly, and ResNet20, ResNet32, and ResNet110 also perform similarly. Additionally, CNN is less affected by spurious examples than MLPs while Vgg16 is also slightly less affected than ResNets.

\mypara{Why are MLPs more sensitive to small patterns?}
We observe that for \isc, MLP seems to be the only architecture that can learn
the spurious correlation when the number of spurious examples is small
($<100$).  At the same time, CNN requires slightly more spurious examples while
ResNets and Vgg16 cannot learn the spurious correlation on small patterns (note
that the y-axis on \cref{fig:difarchs_score_plot} (e) is very small).
Why is this happening? We hypothesize that different architectures have
different sensitivities to the presence of evidence, i.e., the pixels of the
image.  Some architectures change their prediction a lot based on a small
number of pixels, while others require a large number.  If a network
architecture is sensitive to changes in a small number of pixels, it can also
be sensitive to a small spurious pattern.

\begin{figure}[ht]
  \centering
  \subfigure[MNIST]{\includegraphics[width=0.31\textwidth]{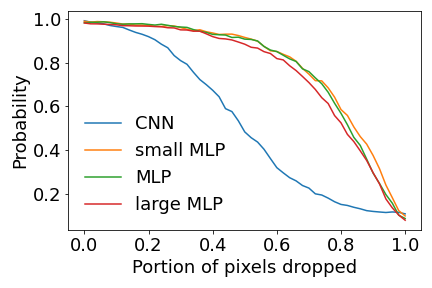}}
  \subfigure[Fashion]{\includegraphics[width=0.31\textwidth]{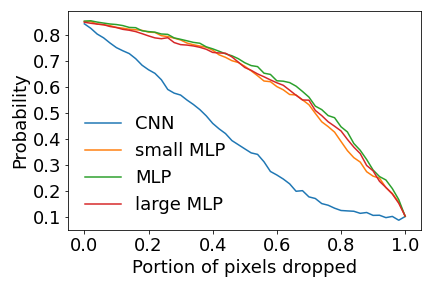}}
  \subfigure[CIFAR10]{\includegraphics[width=0.31\textwidth]{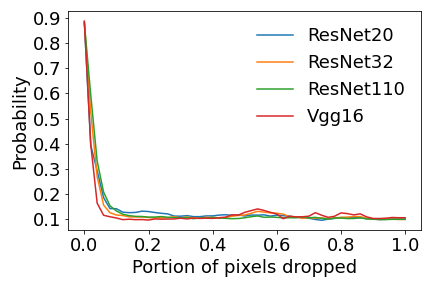}}

  \caption{
    This figure shows the predicted probability of the ground truth label as a function of the portion of non-zero value pixels removed across different architectures and datasets.
  }
  \label{fig:pixel_removal}
\end{figure}

To validate our hypothesis, we measure the sensitivity of a neural network as follows.
First, we train a neural network on the clean training dataset. During testing, we set to zero $0\%, 2\%, \ldots, 98\%, 100\%$ of randomly chosen non-zero pixels in each test image, and measure the predicted probability of its ground truth label.  %
If this predicted probability continues to be high, then we say that the network is insensitive to the input. \cref{fig:pixel_removal} shows the average predicted probability over $500$ training examples as a function of the percentage of pixels set to zero for the MNIST dataset.

We see that MLPs have around $0.9$ average predicted probability with half of the pixels zero-ed out.
In contrast, the average predicted probability is lower in CNNs, suggesting that CNNs may be more sensitive to the zero-ed out pixels. From these results, we can rank the sensitivity of different architectures from non-sensitive to sensitive as MLPs $<$ CNN $<$ ResNets $\approx$ Vgg.
This order matches our observation that MLPs are the most susceptible to spurious correlations, while CNN, ResNets, and Vgg16 are less so -- suggestions that sensitive models may be more susceptible to learning spurious correlations with small patterns.

Finally, we find that architectures that have more parameters \textit{are not} always more vulnerable to spurious correlations.
\cref{tab:arch_params} shows the number of parameters for each architecture.
We see that while CNN has more parameters than small MLP, it is less susceptible to spurious correlations.
Vgg16 and ResNet20 show a similar pattern. This observation is counter to~\citet{sagawa2020investigation}, who suggest that neural networks with more parameters can learn spurious correlations more easily, and it may be because they are looking at a different type of spurious correlation.

\begin{table}[ht]
  \centering
  \caption{Number of parameters in each architecture.}
  \begin{tabular}[]{cccc}
    \toprule
    small MLP & MLP & large MLP & CNN \\
    203,530 & 335,114 & 932,362 & 312,202 \\
    \midrule
    ResNet20 & ResNet32 & ResNet110 & Vgg16 \\
    269,722 & 464,154 & 1,727,962 & 134,301,514 \\
    \bottomrule
  \end{tabular}
  \label{tab:arch_params}
\end{table}

\subsection{Optimization process}\label{app:optimization}

We study how the optimization process affects the learning of the spurious correlation.
We look into two main components of the optimization process, the optimizers and the use of gradient clipping, and examine how each component affects the spurious score.
For the optimizers, we compare between Adam and SGD, while for gradient clipping, we compare between no gradient clipping and clipping the norm the gradient to $0.1$.
We repeat the five times with different random seeds and record their mean and standard error.

\begin{figure}[ht]
  \begin{minipage}{.48\textwidth}
    \centering
    \subfigure[MNIST SGD vs. Adam]{\includegraphics[width=0.90\textwidth]{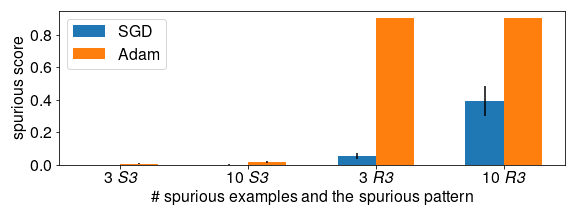}}
  
    \subfigure[Fashion SGD vs. Adam]{\includegraphics[width=0.90\textwidth]{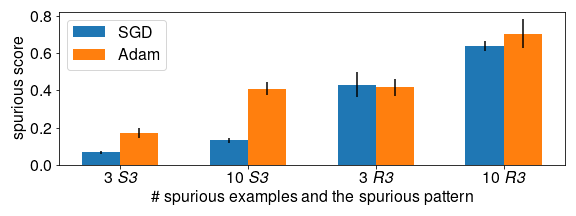}}
  
    \subfigure[CIFAR10 SGD vs. Adam]{\includegraphics[width=0.90\textwidth]{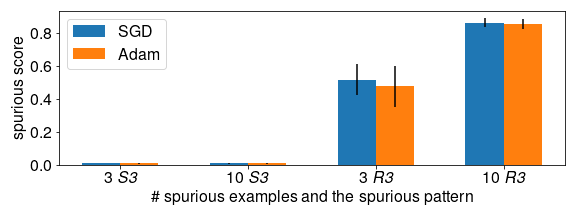}}
    \caption{
      The mean and standard error of the spurious scores on neural networks trained with SGD versus Adam.
      We consider networks trained with three and ten spurious examples as well as using the \isc\ and \irc\ patterns.
    }
    \label{fig:bar_difopt}
  \end{minipage}
  \quad
  \begin{minipage}{.48\textwidth}
    \centering
    \subfigure[MNIST w/ vs. w/o clipping]{\includegraphics[width=0.90\textwidth]{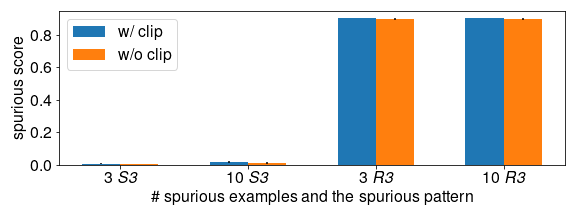}}
  
    \subfigure[Fashion w/ vs. w/o clipping]{\includegraphics[width=0.90\textwidth]{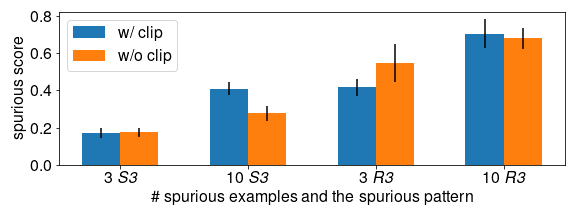}}
  
    \subfigure[CIFAR10 w/ vs. w/o clipping]{\includegraphics[width=0.90\textwidth]{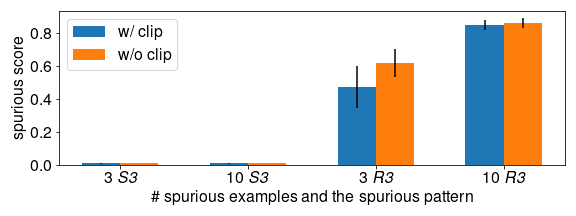}}
    \caption{
      The mean and standard error of the spurious scores on neural networks trained with and without gradient clipping.
      We consider networks trained with three and ten spurious examples as well as using the \isc\ and \irc\ patterns.
    }
    \label{fig:bar_difclip}
  \end{minipage}
\end{figure}

\cref{fig:bar_difopt} shows the average spurious scores between different optimizers on different datasets and spurious patterns.
We find that Adam is slightly more susceptible to spurious correlations than SGD, and gradient clipping does not affect the results much.

\cref{fig:bar_difclip} shows the average spurious scores on networks trained with and without gradient clipping on different datasets and spurious patterns.
We see that, in most cases, with and without gradient clipping performs similarly.
It appears that using gradient clipping alone is not sufficient to eliminate the rare spurious correlations from neural networks.

Overall, we see that rare spurious correlations are learnt regardless of the choice of the optimizer and whether the gradient clipping is performed or not.
This indicates that tweaking individual components in the optimization process may not be sufficient to remove spurious correlations. %
\section{Can Rare Spurious Correlations be Removed through Data Deletion Methods?}\label{app:data_del}

There has been a growing body of recent work on data deletion methods~\cite{koh2019accuracy,izzo2021approximate}. Privacy laws such as the GDPR allow individuals to request an entity to remove their data, which includes removing it from any trained machine learning model. Since retraining models from scratch may be computationally expensive, a body of work has looked into developing more efficient methods. Here, we will look at two simple and canonical methods -- incremental retraining and group influence~\citep{koh2019accuracy,basu2020second} that approximate the model that is trained without the deleted data points. Incremental retraining continues the training process for a number of epochs on the training data minus the deleted data, which effectively down weights the deleted data point in training. The group influence function computes a first-order approximation to the model that is trained without the deleted data point, motivated by influence functions from robust statistics. Both methods apply when multiple data points are deleted.

If all examples with a particular spurious pattern were deleted from the training set, then the spurious pattern and the target class should not be correlated in the resulting network.
Therefore, we expect that a good data deletion method, when given a trained network and all training examples with a specific spurious pattern, should remove the associated spurious correlation from the network.
In this section, we next investigate whether this is indeed the case.

\mypara{Setup.} We follow the same setup as in \cref{fig:score_plot}. We fix the spurious pattern to be \irc, which is the pattern that gives the strongest correlation. We train the networks with $3$, $5$, $10$, $20$, and $100$ spurious examples. We apply two data deletion methods, incremental retraining and group influence, to the trained network. Each method takes in the trained network and the spurious training examples, and generates a new network that approximates the network that is trained on a training set without the spurious examples. We then measure the spurious scores for three types of models -- the model before data deletion, model processed with incremental retraining, and model processed with group influence.

For incremental retraining, we continue the retraining process for $70$ epochs on the data minus the spurious examples (recall that the original models were also trained for $70$ epochs). For group influence, we adapt a publicly available implementation for data deletion\footnote{public repository: \url{https://github.com/ryokamoi/pytorch_influence_functions}}.

\begin{figure}[h!]
  \centering

  \subfigure[MNIST]{\includegraphics[width=0.31\textwidth]{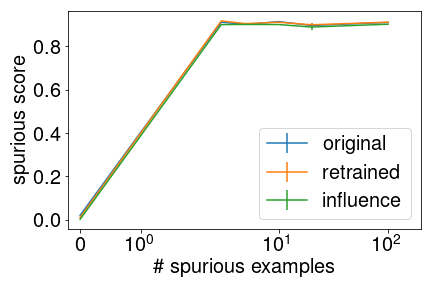}}
  \subfigure[Fashion]{\includegraphics[width=0.31\textwidth]{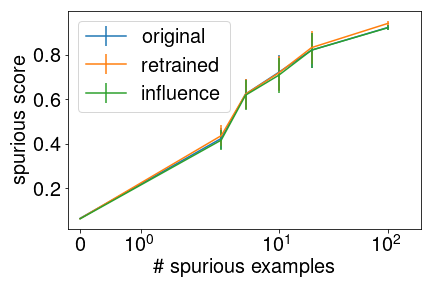}
  }
  \subfigure[CIFAR10]{\includegraphics[width=0.31\textwidth]{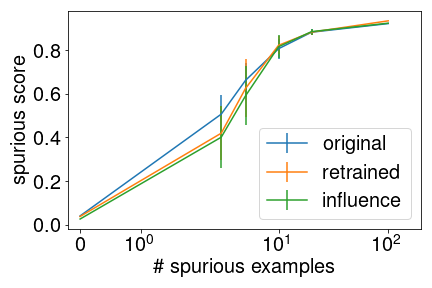}}

  \caption{
    The mean and standard error of spurious scores of the original models, models after incremental retraining, and models after the group influence method.
    The choice of spurious pattern is \irc, $c_{tar}=0$, and the optimizer is Adam.
    The lines for original and retrained are jittered by a small amount so that they are not completely overlapped.
  }
  \label{fig:retrain_score_plot}
\end{figure}

\mypara{Results.} \cref{fig:retrain_score_plot} shows the results. We see that for all three datasets, the models processed by the data deletion methods have similar spurious scores as the models before deletion. 
This implies that the spurious correlations remain even after ``data deletion'', suggesting that \textit{these data deletion methods may not be effective at properly removing spurious examples.} This has two implications -- first, that rare spurious correlations, once introduced, may be challenging to remove. A second implication is that some data deletion methods may not properly remove all traces of the deleted data. We suggest that as a sanity check, future data deletion methods should test whether rare spurious correlations corresponding to the deleted examples are removed. 

Finally, we note that there is a class of indistinguishable data deletion algorithms~\citep{ginart2019making,neel2020descent} that provably ensure by adding noise that the deleted model is statistically indistinguishable from full retraining on the training data after deletion.
However, these algorithms mostly apply to simpler problems, and we do not have efficient guaranteed deletion for non-convex problems such as training neural networks.

\section{Experimental Details and Additional Results}\label{app:additional-results}

The experiments are performed on six NVIDIA GeForce RTX 2080 Ti and two RTX 3080 GPUs located on three servers.
Two of the servers have Intel Core i9 9940X and 128GB of RAM and the other one has AMD Threadripper 3960X
and 256GB of RAM.
All neural networks are implemented under the PyTorch framework\footnote{Code and license can be found in \url{https://github.com/pytorch/pytorch}.}~\citep{paszke2019pytorch}.
Other packages, including
\texttt{numpy}\footnote{Code and license can be found in \url{https://github.com/numpy/numpy}.}~\citep{harris2020array},
\texttt{scipy}\footnote{Code and license can be found in \url{https://github.com/scipy/scipy}.}~\citep{2020SciPy-NMeth},
\texttt{tqdm}\footnote{Code and license can be found in \url{https://github.com/tqdm/tqdm}.}, and
\texttt{pandas}\footnote{Code and license can be found in \url{https://github.com/pandas-dev/pandas}.}~\citep{reback2020pandas},
are also used.

\subsection{Experimental setup for Section~\ref{sec:rare}}\label{app:main-detail}

\mypara{Architectures.}
For MNIST and Fashion, we consider multi-layer perceptrons~(MLP) with ReLU activation functions.
MLP has two hidden layers, and each layer has $256$ neurons.
For CIFAR10, we consider ResNet20~\citep{he2016deep}.

\mypara{Optimizer, learning rate, and data augmentation.}
We use the Adam~\citep{kingma2014adam} optimizer and set the initial learning rate to $0.01$ for all models.
We train the model for $70$ epochs.
For the learning rate schedule, we decrease the learning rate by a factor of $0.1$ on the $40$-th, $50$-th, and $60$-th epoch.
For CIFAR10, we apply data augmentation during training.
When an image is passed in, we pad each border with four pixels and randomly crop the image to $32$ by $32$.
We then, with $0.5$ probability, horizontally flip the image.

\subsection{Experimental setup for Section~\ref{sec:consequence}}\label{app:privacy-detail}

\mypara{Membership inference attack setup.}
We split each dataset into four equally sized sets -- target model training/testing sets and shadow model training/testing sets.
We then add spurious features to a number of examples in the target model training set.
We train the target model using the target model training set and the shadow model using the shadow model training set.
Next, we extract the features on the target/shadow model training/testing sets.
The feature we use are the output of the target/shadow model on these data and a binary indicator indicating whether the example is correctly predicted as the features.
We then train an attack model (binary classifier) to distinguish whether an example comes from the shadow model training or testing sets using the extracted features.
For evaluation, we compute the accuracy of the attack model distinguishing whether an example comes from the target model training or testing sets (test accuracy).

For the shadow and target model, we train them the same way as other models in \cref{sec:rare}.
For the attack model, we use a logistic regression implemented with \texttt{scikit-learn}~\footnote{Code and license can be found in \url{https://github.com/scikit-learn/scikit-learn}.}~\citep{scikit-learn}
with $3$-fold cross-validation to select the regularization parameter from $\{0.01, 0.1, 1.0, 10.0\}$.

\section{Other Additional Results}

The experimental code is available at \url{https://github.com/yangarbiter/rare-spurious-correlation}.

\subsection{Additional results for qualitative analysis: visualizing network weights}\label{sec:visualize_mnist}

\begin{figure}[h!]
    \centering
    \includegraphics[width=.7\textwidth]{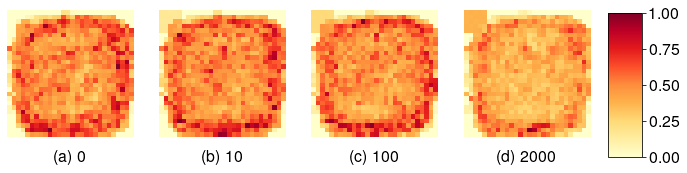}
    \caption{A normalized version of \cref{fig:mlp_weights_norm} (the values of each pixel is scaled to [0, 1] for each figure independently).}
    \label{fig:mlp_weights_norm}
\end{figure}

\mypara{Normalized feature importance.}
In \cref{fig:mlp_weights}, we show the original value of the importance for each pixel.
For completeness, we present \cref{fig:mlp_weights_norm}, which scales each pixel's value to $[0, 1]$ for each sub-figure.
From the figure, we can still see that a small number of spurious examples can cause the learnt weight to change.

\subsection{Additional results for \texorpdfstring{$c_{tar}=1$}{ctar=1}}\label{sec:additional-results}

\begin{figure}[h!]
  \centering
  \subfigure[MNIST, $c_{tar}=1$]{\includegraphics[width=0.31\textwidth]{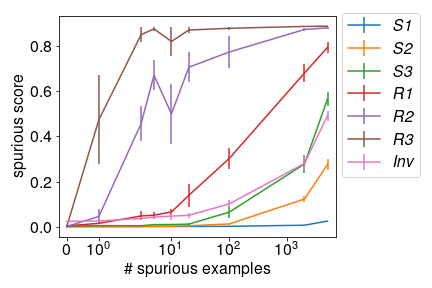}}
  \subfigure[Fashion, $c_{tar}=1$]{\includegraphics[width=0.31\textwidth]{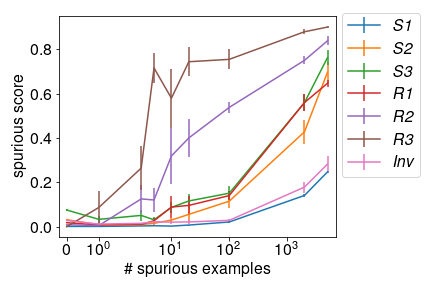}}
  \subfigure[CIFAR10, $c_{tar}=1$]{\includegraphics[width=0.31\textwidth]{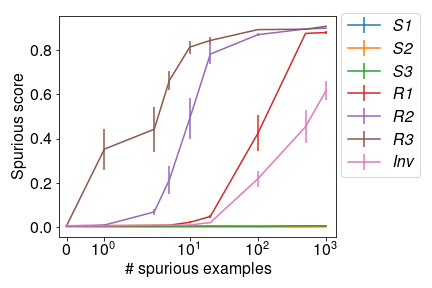}}
  
  \caption{
  Each figure shows the mean and standard error of the spurious scores on three datasets, MNIST, Fashion, and CIFAR10, $c_{tar}=1$, and different numbers of spurious examples.
  In these figures, we use MLP as the network architecture for MNIST and Fashion, and we use ResNet20 for CIFAR10.
  }
  \label{fig:score_plot_tar1}
\end{figure}

\mypara{Results for another target class.}
For the completeness of  \cref{fig:score_plot}, we show the results for $c_{tar}=1$ in \cref{fig:score_plot_tar1}.
We see that in all these cases, increasing regularization strength decreases spurious scores 
The conclusions that can be made here are similar to ones made in \cref{sec:rare} \cref{fig:score_plot}.

\begin{figure}[h!]
  \centering
  \subfigure[MNIST, $c_{tar}=1$]{\includegraphics[width=0.31\textwidth]{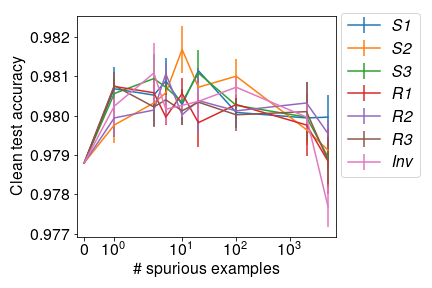}}
  \subfigure[Fashion, $c_{tar}=1$]{\includegraphics[width=0.31\textwidth]{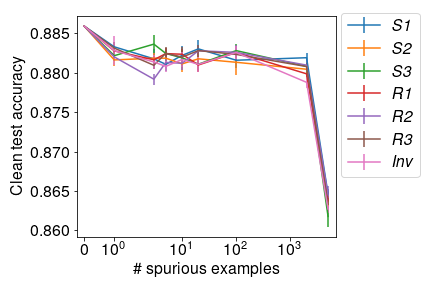}}
  \subfigure[CIFAR10, $c_{tar}=1$]{\includegraphics[width=0.31\textwidth]{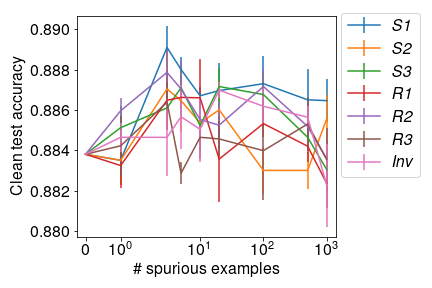}}
  
  \caption{
  Each figure shows the mean and standard error of the clean test accuracy on three datasets, MNIST, Fashion, and CIFAR10, $c_{tar}=0$, and different numbers of spurious examples.
  In these figures, we use MLP as the network architecture for MNIST and Fashion, and we use ResNet20 for CIFAR10.
  }
  \label{fig:clean_acc_plot}
\end{figure}

\mypara{Results for clean test accuracy}
For completeness, we report the clean test accuracy for different number of spurious examples in \cref{fig:clean_acc_plot}.
From the figures, we see that the clean test accuracy does not get affected too much as the number of spurious example grows.
This aligns with our observation in \cref{sec:consequence}, in which we see that the minimum and maximum clean test accuracy across all runs of experiments are close.
This is as expected as the spurious examples are only presented in a small portion of the training data while accuracy measures average behavior.

\subsection{Membership inference experiment with other spurious patterns}\label{app:add-mem-inf}

\begin{figure}[h!]
    \centering
    \subfigure[MNIST, \isa]{\includegraphics[width=.30\textwidth]{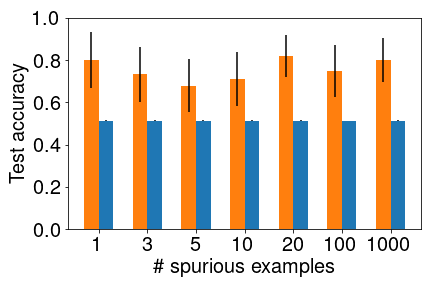}}
    \subfigure[Fashion, \isa]{\includegraphics[width=.30\textwidth]{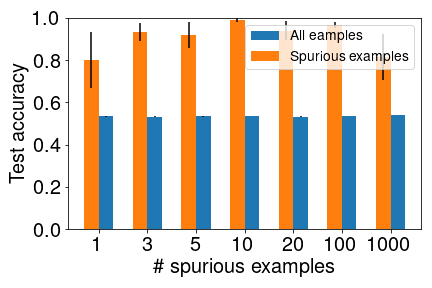}}
    \subfigure[CIFAR10, \isa]{\includegraphics[width=.30\textwidth]{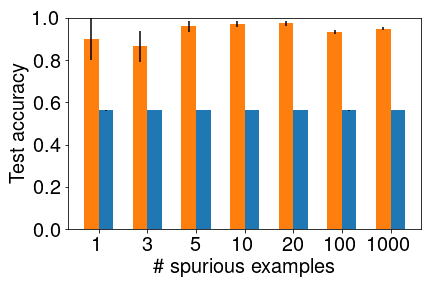}}
    
    \subfigure[MNIST, \isb]{\includegraphics[width=.30\textwidth]{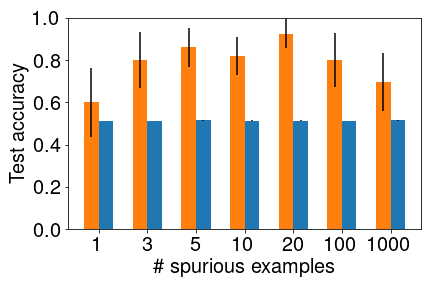}}
    \subfigure[Fashion, \isb]{\includegraphics[width=.30\textwidth]{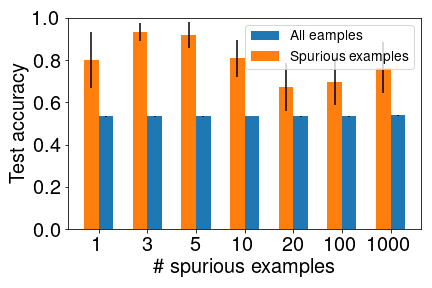}}
    \subfigure[CIFAR10, \isb]{\includegraphics[width=.30\textwidth]{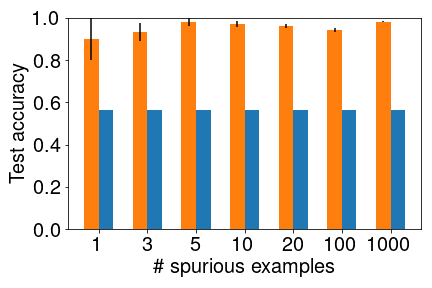}}
    
    \subfigure[MNIST, \isc]{\includegraphics[width=.30\textwidth]{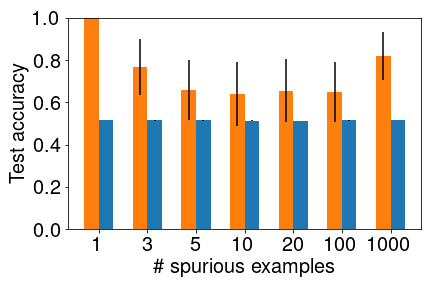}}
    \subfigure[Fashion, \isc]{\includegraphics[width=.30\textwidth]{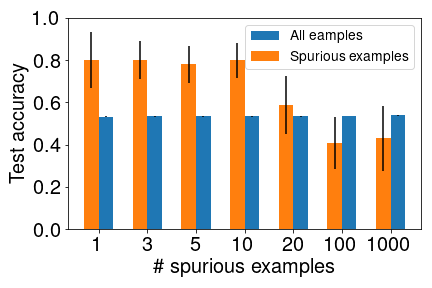}}
    \subfigure[CIFAR10, \isc]{\includegraphics[width=.30\textwidth]{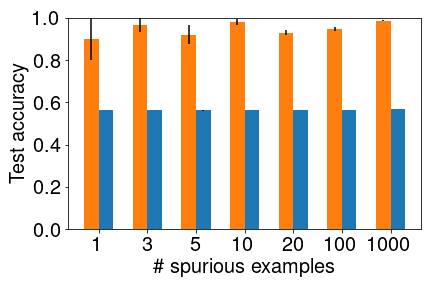}}
    
    \subfigure[MNIST, \ira]{\includegraphics[width=.30\textwidth]{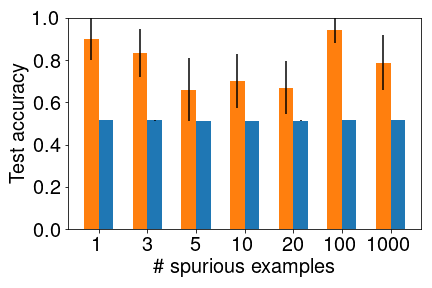}}
    \subfigure[Fashion, \ira]{\includegraphics[width=.30\textwidth]{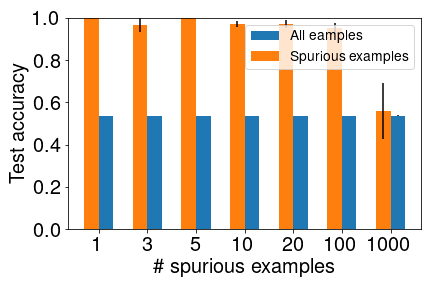}}
    \subfigure[CIFAR10, \ira]{\includegraphics[width=.30\textwidth]{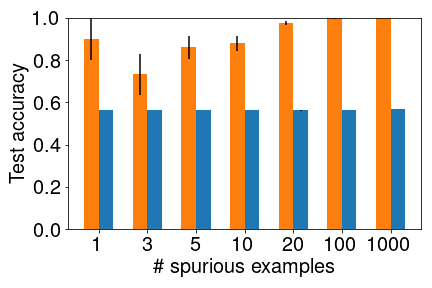}}
    
    \subfigure[MNIST, \irb]{\includegraphics[width=.30\textwidth]{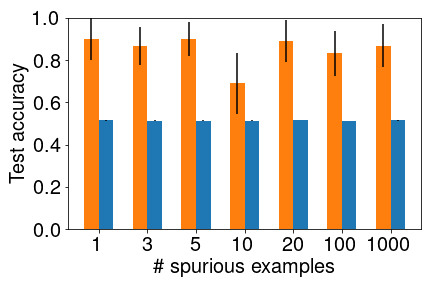}}
    \subfigure[Fashion, \irb]{\includegraphics[width=.30\textwidth]{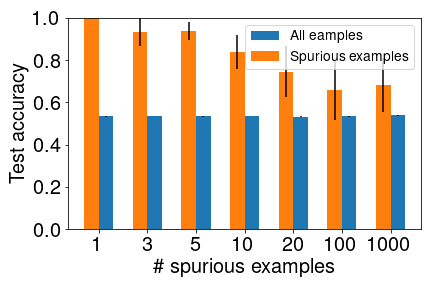}}
    \subfigure[CIFAR10, \irb]{\includegraphics[width=.30\textwidth]{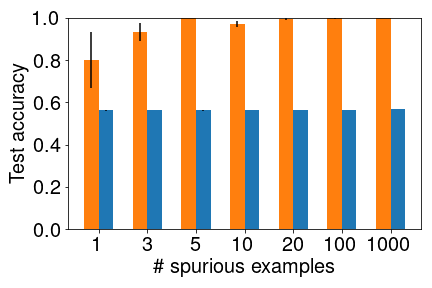}}
    
    \subfigure[MNIST, \irc]{\includegraphics[width=.30\textwidth]{figs/meminf/memmnist_v20_bbox.png}}
    \subfigure[Fashion, \irc]{\includegraphics[width=.30\textwidth]{figs/meminf/memfashion_v20_bbox.png}}
    \subfigure[CIFAR10, \irc]{\includegraphics[width=.30\textwidth]{figs/meminf/memcifar10_v20_bbox.png}}
    
    \caption{Additional results with all spurious patterns ($\ctar=1$) for the membership inference experiments (partial results are in \cref{fig:meminf_plot}).}
    \label{fig:add-meminf-fig}
\end{figure}

In \cref{fig:meminf_plot}, we show the results of the membership inference experiment while using \irc\ as the spurious pattern.
For completeness, we show the results of attacking models trained on datasets with different spurious patterns in  \cref{fig:add-meminf-fig}.

\subsection{Additional results for spurious test accuracy}\label{app:add_spu_tst_acc}

\begin{figure}[h!]
    \centering
      
      \subfigure[MNIST, $\ctar=1$]{\includegraphics[width=0.29\textwidth]{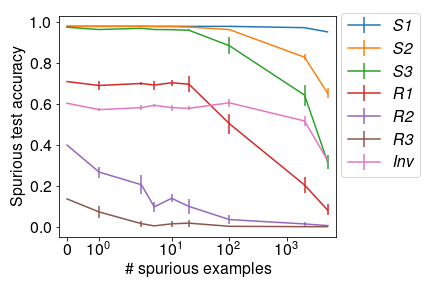}}
      \subfigure[Fashion, $\ctar=1$]{\includegraphics[width=0.29\textwidth]{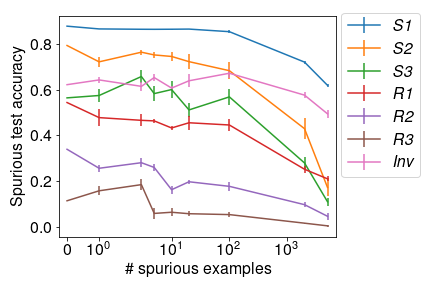}}
      \subfigure[CIFAR10, $\ctar=1$]{\includegraphics[width=0.29\textwidth]{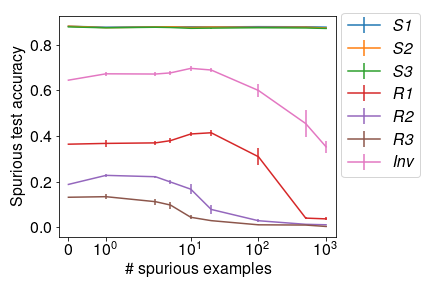}}
    
    \caption{Additional results on the spurious test accuracy over \cref{fig:spacc_plot}.}
    \label{fig:add-spu_tstacc-fig}
\end{figure}

For the completeness of \cref{fig:spacc_plot}, we show all results of the spurious test accuracy in \cref{fig:add-spu_tstacc-fig}.

\subsection{Natural rare spurious correlation}
\label{app:natural_rare}

\begin{figure}[h]
  \centering
  \subfigure[Dim]{
    \includegraphics[width=0.45\textwidth]{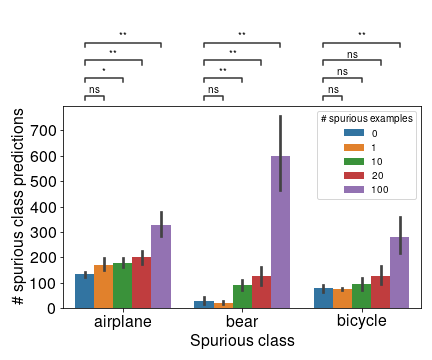}}
  \subfigure[Grass]{
    \includegraphics[width=0.45\textwidth]{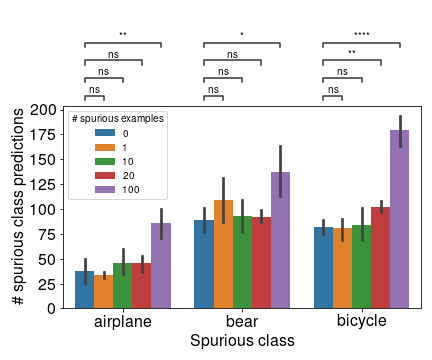}}
    
  \caption{
  The number of test examples that does not belongs to the spurious class gets predicted as the spurious class.
  We conduct Welch's t-test~\citep{welch1947generalization} on the number of spurious class predictions between the model trained without spurious examples and models trained with different number of spurious examples.
  The notations for the p-values:
  ns$: 0.05 < p \leq 1.$, *$: 10^{-2} < p \leq 0.05$, **$: 10^{-3} < p \leq 10^{-2}$, ***$: 10^{-4} < p \leq 10^{-3}$, ****$: p \leq 10^{-4}$.
  }
  \label{fig:nico_main_additional}
\end{figure}

\cref{fig:nico_main_additional} shows the results for using dim and grass context as the spurious context.
From the result, in all cases, spurious correlations are significantly learnt by neural networks with just $100$ spurious examples (less than $.2\%$ of the training data).
We see that there are some cases spurious correlations are more easily learnt. For example, under the case where the dim context is the spurious context, two out of three spurious classes gets effected by $10$ spurious examples, and under Grass context with bicycle as the spurious class, it only requires $20$ spurious examples.
These results confirms that natural spurious patterns can significantly effect a neural network.

\subsection{Ablation studies}
\label{app:ablation_normalized}

\begin{figure}[h!]
    \centering
    \subfigure[MNIST, $\ctar=0$]{\includegraphics[width=.31\textwidth]{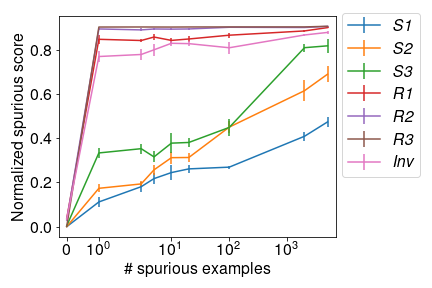}}
    \subfigure[Fashion, $\ctar=0$]{\includegraphics[width=.31\textwidth]{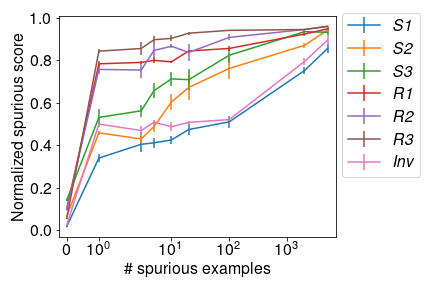}}
    \subfigure[CIFAR10, $\ctar=0$]{\includegraphics[width=.31\textwidth]{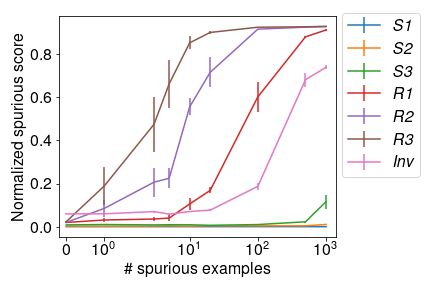}}
    \caption{Results of the normalized spurious score~(\cref{eq:norm_spuscore}) with different number of spurious examples on MNIST, Fashion, and CIFAR10.}
    \label{fig:norm_spuscore}
\end{figure}

\mypara{Normalized spurious score.}
A question is whether changing the scale of the spurious score (\cref{eq:spuscore}) could change the results in \cref{fig:score_plot}.
In \cref{fig:norm_spuscore}, we show a similar figure as \cref{fig:score_plot} but with the y-axis switched to the normalized spurious score~\cref{eq:norm_spuscore}.
From the figures, we see similar trend as of \cref{fig:score_plot}, and we can reach a similar conclusion as using the spurious score.

\begin{equation}
  \frac{f_{\ctar}(\Phi_{\cX}(\x, \bxsp)) - f_{\ctar}(\x)}{f_{\ctar}(\Phi_{\cX}(\x, \bxsp))} > \epsilon \ .
  \label{eq:norm_spuscore}
\end{equation}

\section{Additional Related Work and Discussions}\label{app:additional-rel-work}

\mypara{Short-cut learning \citep{geirhos2020shortcut,yu2021indiscriminate} and simplicity bias~\citep{shah2020pitfalls}}.
The major difference between these works and our work is that they mainly focus on when spurious correlations are learned.
They are less concerned with how rare these spurious correlations are, which is different from the focus of our work.

\mypara{Backdoor attack.}
Our experiment procedure is similar to the data poisoning process applied in the backdoor attack.
However, the context and message of our study are different from that of backdoor attacks in the following three ways.
First, the spurious examples are not adversarial and, most of the time, are natural and simple. 
Second, our analysis is done on soft prediction, while backdoor attack focuses on hard label prediction.
This allows us to observe results such as one spurious example that can significantly impact the model and its privacy implications, as these changes can be too subtle to be observed in a hard label setting.

Finally, we focus on a more fine-grained quantitative study on how the number of appearances of spurious training examples affects the performance of neural networks.
The prior backdoor attack works often study a fixed poisoning rate (translates to having a fixed portion of the spurious examples in the training set).
More specifically, \citep{li2021anti,wu2021adversarial,nam2020learning} focus on $0.1\%$ of training examples, which translates to 50 spurious examples for CIFAR10.
Although they also study a small portion of the training set, their results are unable to provide insights on when the neural network begins to be affected by these spurious examples.
In addition, we study the impact of different regularization methods (weight decay, noisy inputs, and gradient clipping), different optimizers (SGD vs. Adam, which makes a difference), and different architectures (CNN vs. MLP).
These aspects had not been rigorously studied in previous work.

\mypara{Data deletion methods.} Inspired by GDPR, there are many recent works on data deletion. \citet{izzo2021approximate} demonstrate data deletion for linear classifiers but not for non-linear models such as neural networks.
The use of influence and group influence functions for data deletion is also studied by many~\cite{koh2017understanding,koh2019accuracy,basu2020second}.
\citet{basu2020influence} point out that influence functions can be fragile for deeper neural networks.
Our work shows that influence functions cannot remove spurious correlations caused by the deleted examples, which is different.

\mypara{Concerns for expanding training sets.}
Researchers have also discovered ways that more data can hurt the model in terms of the generalization ability and test time accuracy~\citep{nakkiran2019deep,min2020curious}.
In this work, we uncover a different way that more data can hurt: more data could introduce more spurious correlations.

\mypara{Comparison with other works related to memorization.}
There are works related to the memorization of a small number of special examples in the training set.
\citet{carlini2019secret} focus on language generative models. For generative models, they assign a likelihood to each input, which directly indicates whether an example is likely to appear or not. However, the likelihood for classification models indicates whether an example belongs to a certain class. Therefore, from this work alone, we are not able to conclude whether the classification model suffers from similar privacy issues.
\citet{nasr2021adversary} and \citet{jagielski2020auditing} consider how in a worst-case scenario malicious inputs can cause privacy issues.
However, our study focuses on whether natural spurious patterns can cause privacy issues.
Although some steps for performing the experiment are similar, we focus on different aspects.

\mypara{Mitigation of rare spurious correlations.}
There are existing works on defense against backdoor attacks and de-biasing neural networks, which are related to mitigating specific spurious correlations.
One significant difference between these works and our work is that these works usually make some assumptions or use certain additional properties based on their application.
For example, \citet{li2021anti} assume that spurious examples in the training set have a lower training loss than clean examples, \citet{wu2021adversarial} assume that backdoor-related neurons are also more sensitive to adversarial perturbations, and \citet{nam2020learning} assume the spurious (biased) examples are harder to learn (one can start to see some contradiction between \citep{li2021anti} and \citep{nam2020learning} as spurious examples may be easier or harder to learn in different settings).
These assumptions may work well in specific settings.
But, there are no guarantees on whether they will work on other kinds of spurious patterns (or backdoor attacks).

In our work, we take a more general perspective and do not make any assumptions besides the fact that spurious examples are rare.
Here are some intuitions on how regularization methods can help with mitigating rare spurious correlations.
For weight decay (which is the same as regularization), it effectively puts a penalty on every weight that is greater than zero.
Therefore, it discourages the neural network from putting weights on features that occur rarely.
For noise injection, it is inspired by the differential privacy literature.
Differential privacy is often achieved by adding noise during training, and by doing so, one can make the model less affected by a small group of training examples.
Finally, we use gradient clipping as a baseline to show that not all kinds of regularization methods are useful in mitigating rare spurious correlations.

\mypara{Relationship with adversarial examples.}
One nice connection to adversarial examples is that if rare spurious correlations are present in the training data, an adversary can introduce them in a test point to bring down the classification accuracy.
Our experiments in \cref{fig:add-meminf-fig} actually illustrate this.
In that sense, the adversary in our experiment is, in fact, a highly constrained test-time adversary (who can only apply a specific pattern to test inputs).

An interesting hypothesis is that some adversarial examples can be caused by rare spurious correlations.
This suggests a future direction in examining the relationship between the inserted spurious pattern and the direction of the adversarial perturbations~\citep{goodfellow2014explaining}.
One related work is from \citet{weng2020trade}, who suggests that “by increasing the robustness of a network to adversarial examples, the network becomes more vulnerable to backdoor attacks.”
Backdoor attacks also insert a small portion of specifically designed spurious patterns into the training set.
Thus, robustness to backdoor attacks can be related to the spurious correlation we are measuring.
Their result indicates that there exist adversarial examples that are not purely caused by rare spurious examples.

\mypara{Further discussions on \cref{fig:spacc_plot}.}
In the figure, there are cases where the spurious test accuracy does not drop for a fairly large number of spurious examples.
There are two implications: 1) As in these cases, even though the spurious test accuracy does not drop, we still see that spurious examples are more vulnerable to the membership inference attack (see \cref{fig:meminf_plot}).
This means that there are cases where the hard predictions of the neural network cannot measure whether the spurious correlations are learned.
2) This implies that different spurious patterns can have different effects on different datasets/architectures.
One interesting future direction is to figure out when spurious examples are more effective and when it is less effective. 
\end{document}